\newcommand\ARXIV[2]{#1} 

\def\ieee{1}
\if 1\ieee
    \documentclass[a4paper,conference]{IEEEtran}
\else
    \documentclass[a4paper,conference]{article}
    \usepackage[margin=0.7in]{geometry}
\fi

\usepackage[caption=false,labelfont=rm,textfont=rm]{subfig}

\usepackage[pdftex,dvipdfmx]{graphicx}
\usepackage{bmpsize}

\usepackage{mystyle}

\begin{document}

\title{Deep Surface Reconstruction from Point Clouds with Visibility Information}

\if 1\ieee
    \author{\IEEEauthorblockN{Raphael Sulzer\IEEEauthorrefmark{1},
    Loïc Landrieu\IEEEauthorrefmark{1},
    Alexandre Boulch\IEEEauthorrefmark{3},
    Renaud Marlet\IEEEauthorrefmark{2}\IEEEauthorrefmark{3} and
    Bruno Vallet\IEEEauthorrefmark{1}}
    \IEEEauthorblockA{\IEEEauthorrefmark{1}LASTIG, Univ Gustave Eiffel, IGN-ENSG, F-94160 Saint-Mandé, France
    \\ Email: firstname.lastname@ign.fr}
    \IEEEauthorblockA{\IEEEauthorrefmark{2}LIGM, Ecole des Ponts, Univ Gustave Eiffel, CNRS, Marne-la-Vallée, France}
    \IEEEauthorblockA{\IEEEauthorrefmark{3}Valeo.ai, Paris, France}
    }
\else
    \date{}
    \author{Raphael Sulzer$^1$, Loïc Landrieu$^1$, Alexandre Boulch$^3$, Renaud Marlet$^{2,3}$, Bruno Vallet$^1$ 
    \\ $^1$LASTIG, Univ Gustave Eiffel, IGN-ENSG, F-94160 Saint-Mandé, France
    \\ $^2$LIGM, Ecole des Ponts, Univ Gustave Eiffel, CNRS, Marne-la-Vallée, France
    \\ $^3$Valeo.ai, Paris, France}
\fi
\maketitle

\if 0\ieee
    \thispagestyle{empty}
\fi

\begin{abstract}
Most current neural networks for reconstructing surfaces from point clouds ignore sensor poses and only operate on raw point locations. Sensor visibility, however, holds meaningful information regarding space occupancy 
and surface orientation. 
In this paper, we present two simple ways to augment raw point clouds with visibility information, so it can directly be leveraged by surface reconstruction networks with minimal adaptation. 
Our proposed modifications consistently improve the accuracy of generated surfaces as well as the generalization ability of the networks to unseen shape domains. 

\end{abstract}


%

\section{Introduction}

\global\csname @topnum\endcsname 0
\global\csname @botnum\endcsname 0

The problem of reconstructing a watertight surface from a point cloud has recently been addressed by a variety of deep learning based methods. Compared to traditional approaches, deep surface reconstruction (DSR) can learn shape priors \cite{Park2019,Mescheder2019} and leverage shape similarities \cite{point2mesh} to complete missing parts \cite{dai2020sgnn}, filter outliers, or smoothen noise in defect-laden point clouds. 
DSR methods, however, often derive priors from training datasets with few shape classes, generalizing poorly to unseen categories or datasets.
Learning more local priors improves consistency across different objects or scenes \cite{lig,dgnn} but may result in higher sensitivity to noise or other defects.
{Besides}, lack of global context complicates surface orientation.

For real world point clouds, usually acquired via active or passive methods such as LiDAR scanning or multi-view stereo (MVS), the sensor position can be known and used to relate each observed point with a line of sight. Such visibility information can then help to orient surface normals \cite{schertler2017towards}  or predict occupancy \cite{Labatut2009a,Vu2012,Jancosek2011}. 
While visibility is key for MVS, it has largely been ignored by DSR methods. In fact, sensor positions are usually not given in reconstruction benchmarks from point clouds.
\begin{figure}[t]
	\centering
\vspace*{-3mm}
\subfloat[Reconstruction using only the points position.]{
\begin{tabular}{lr}
\includegraphics[width=0.55\linewidth,trim={0cm 2cm 1cm 4cm},clip]{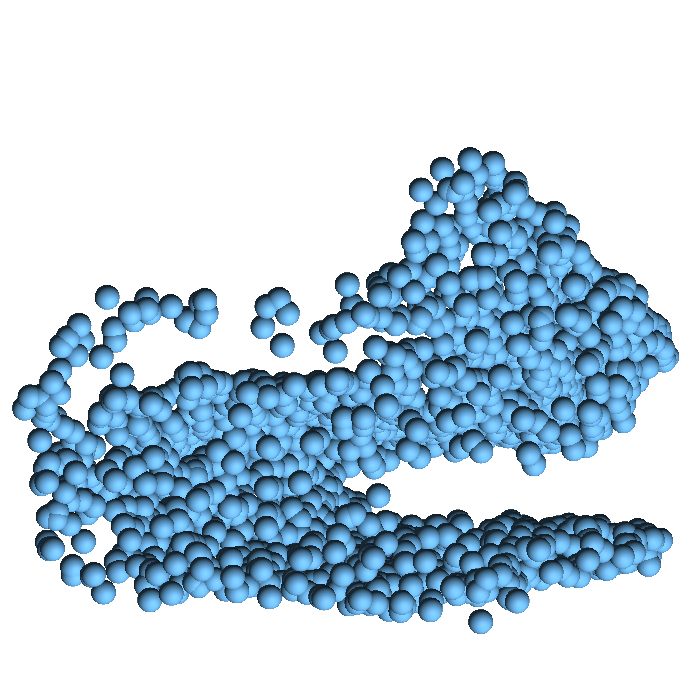}
&
\includegraphics[width=0.3\linewidth]{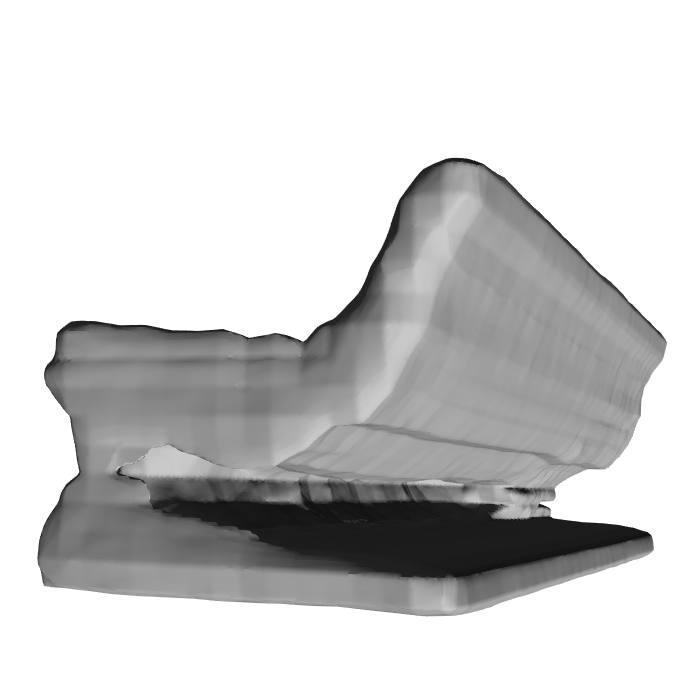}      
\end{tabular}
}
\\
\subfloat[Reconstruction with visibility augmented point cloud.]{
\begin{tabular}{lr}
\includegraphics[width=0.55\linewidth,trim={0cm 2cm 1cm 4cm},clip]{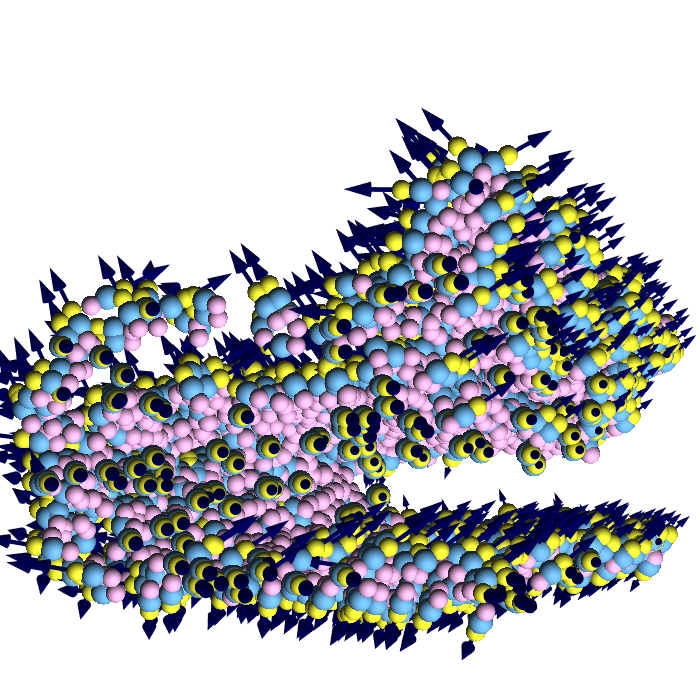}
&
\includegraphics[width=0.3\linewidth]{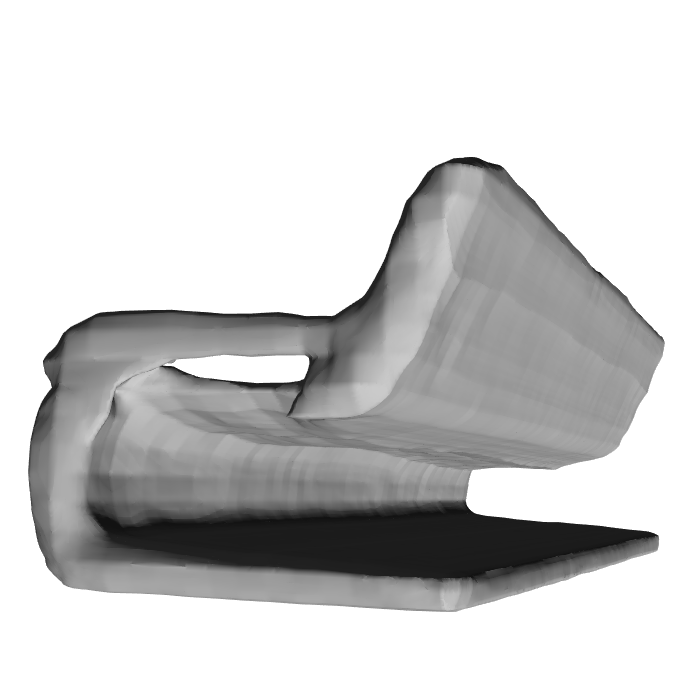}      
\end{tabular}
}
\caption{\textbf{Surface Reconstruction with Visibility Information.}
We augment each 3D point
\protect\tikz \protect\node[circle, thick, draw = none, fill = tempblue, scale = 0.7] {};
 with a sightline vector
 \protect\tikz[baseline=-.25em] \protect\draw[->, ultra thick, draw = tempdarkblue] (0,0) -- (0.3,0);
 pointing towards the sensor observing it. Additionally, two auxiliary points are placed before \protect\tikz \protect\node[circle, thick, draw = none, fill = tempyellow, scale = 0.7] {};
and after
\protect\tikz \protect\node[circle, thick, draw = none, fill = tempred, scale = 0.7] {};
the observed point along the sightline. 
This allows DSR networks, with very little modification, to reconstruct a significantly more accurate surface.
}
\vspace*{-1mm}
\label{fig:teaser}
\end{figure}
To remedy this, we consider virtual scanning rather than uniform sampling, and we show that many DSR methods can easily be adapted to benefit from visibility (cf.\ Figure~\ref{fig:teaser}). Our main contributions are as follows:
\begin{itemize}[nosep]
  \item We propose two simple ways to add visibility information to 3D point clouds, and we detail how to adapt DSR methods to utilize them, with very little changes.
  \item Using synthetic and real data, at object and scene level, we show for a wide range of state-of-the-art DSR methods that models leveraging visibility reconstruct higher-quality surfaces and are more robust to domain shifts.
  %
  %
  %
\end{itemize}
Incidentally, our benchmarks also allow us to compare a range of recent state-of-the-art DSR methods on the same ground.

\section{Related Work}
Many traditional surface reconstruction methods use visibility information \cite{Labatut2009a,Vu2012,Jancosek2011,wasure,Bodis-Szomoru2017,Jancosek2014,Zhou2019}. They are usually based on a 3D Delaunay {tetrahedralization}, which is intersected with lines of sight to attribute visibility features to Delaunay cells. While such methods can scale to billions of points \cite{caraffa2021efficiently} and are robust to moderate levels of noise and outliers, they do not incorporate learned shape priors.

In contrast, recent DSR methods have shown to produce more accurate surfaces than traditional approaches, especially for shape categories encountered during training.
Many DSR methods use an implicit surface representation, either based on occupancy \cite{Mescheder2019, Chen2019CVPR,deepLS}, or on the distance to the surface, whether it is signed \cite{Park2019, deepLS, Michalkiewicz2019ICCV, Gropp2020} or unsigned \cite{Atzmon_2020_CVPR, Chibane2020CVPR, Atzmon_2021_ICLR, Zhao2021SignAgnostic}.
To integrate local information, different forms of convolutions are used, either on regular grids \cite{dai2020sgnn, lig, Chibane2020Neural, Peng2020, tang2021sign}, directly on points \cite{Ummenhofer2021Adaptive, boulch2022poco} or via an MLP instead \cite{points2surf}. Other methods rather use an explicit surface representation such as a mesh, which is deformed \cite{point2mesh} or whose elements classified \cite{dgnn, PointTriNet}.

\begin{figure}[t]
\centering
\vspace*{-3mm}
\includegraphics[width=0.7\linewidth]{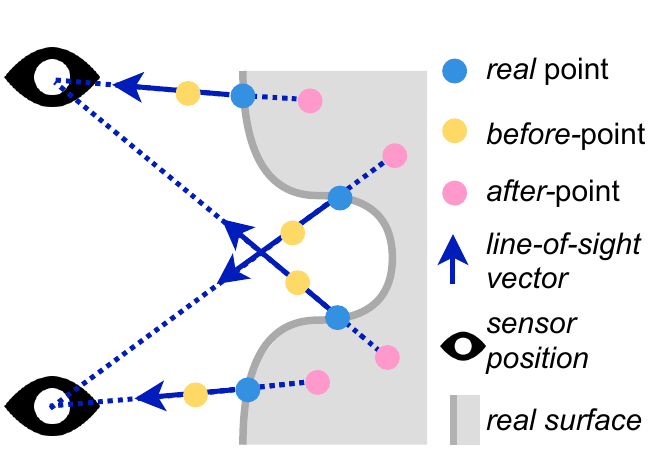}
\vspace*{-3mm}
\caption{
\textbf{Visibility-Augmented Point Cloud.}
Each observed point is associated to a sightline unit vector pointing towards its sensor. Two new points before and after each point are added. They help to disambiguate occupancy.
}
\label{fig:visibility}
\vspace*{-1mm}
\end{figure}

{A key issue is to get a sense of surface point orientation, to choose between reconstructing a thin volume (two main opposite orientations) or a thicker one (one main orientation at void-matter interface). Some methods dismiss the orientation issue by requiring oriented normals as input \cite{lig,Ummenhofer2021Adaptive, screened_poisson,  neural-splines}, albeit producing such normals is a challenging task in itself \cite{schertler2017towards,konig2009consistent,metzer2021dipole}.}
We show that oriented normals can be  advantageously replaced by visibility information.


{Only few deep-learning methods make use of visibility information, typically from multiple views with camera pose information.}
RayNet \cite{RayNet} aggregates features from pixels of different views that intersect in the same voxel, but it outputs a dense point cloud, not a watertight surface mesh.
{Neural radiance fields 
\cite{mildenhall2020nerf,UNISURF} somehow also model the free space between a point and its sensor. They, however, generally assume numerous and dense views (\ie, images), and leverage little or no shape priors.}
%
%
%
%
We argue that DGNN \cite{dgnn}, that classifies Delaunay cells with a graph neural network, currently is the only general DSR method from point clouds with visibility.  However, DGNN relies on handcrafted visibility features requiring substantial geometry processing, while, we propose to directly augment the input point clouds.
For point clouds for which visibility information is not available, Vis2Mesh~\cite{vis2mesh} shows that rendering virtual views and learning point sensor visibility can significantly improve the reconstruction quality of a traditional method.


\global\csname @topnum\endcsname 0
\global\csname @botnum\endcsname 0


\section{Method}

We consider a 3D point cloud $P$ {where} each point $p \,{\in}\, P$ has some coordinates $X_p \,{\in}\, \bR^3$ and knows the position $S_p \,{\in}\ \bR^3$ of a sensor {observing it.} Instead of only using the raw point coordinates {$(X_p)_{p\in P}$} as the input {$(I_p)_{p\in P}$} of a DSR network, we propose two simple ways to augment point cloud $P$ with visibility information, and adapt DSR methods accordingly.


\subsection{Sightline Vector (SV)} 
For each $p \in P$, we define a unit vector $\sv_p$ pointing from the observation $X_p$ to the sensor $S_p$: $\sv_p \,{=}\, {(S_p \,{-}\, X_p)}/{\Vert S_p \,{-}\, X_p\Vert}$. This contains useful information for surface orientation.
We normalize the vector as the distance to the sensor is not as relevant as the viewing angle.

\subsection{Auxiliary Points {(AP)}} To help the network predict empty and full space {immediately} in front of and behind the observed surface, we {consider} two auxiliary points to each point~$p$: a \emph{before-point} $p\beforep$ and an \emph{after-point} $p\afterp$, located along the {sightline} 
on each side of~$p$: $X_{p\beforep} = X_p + \chardist \sv_p, X_{p\afterp} = X_p - \chardist \sv_p$,
where $\chardist$ is a characteristic distance in the {point cloud}~$P$, e.g., the average distance {from a point to its nearest neighbor}.
By construction, $p\beforep$ is likely outside the scanned object {or scene (modulo sensing noise and outliers)}, and ${p\afterp}$, likely inside {(modulo object thickness too)}.

\subsection{Visibility-Augmented Point Cloud}
We use {sightline} vectors and auxiliary points to add visibility information to an input point cloud, separately or together.
\begin{itemize}[leftmargin=1em,itemindent=-1em,topsep=1pt,itemsep=3pt]
\item[]
{(SV)} To use sightline information only, 
we simply concatenate the sightline vector channelwise to the point coordinates to form the network input:
$I_p = (X_p \oplus \sv_p) \in \bR^6$.
\item[]
{(AP)} To use auxiliary points only, we add before-points $p\beforep$ and after-points $p\afterp$ to~$P$, 
with tags $\tp \in \bR^2$ concatenated to point coordinates to identify the point type, \ie, $I_q \,{=}\, (X_q \,{\oplus}\, \tp_q) \in \bR^5$ with $q \in\{p,p\beforep,p\afterp\}$, where $\tp_{p} \,{=}\, [0\,\,0]$ (observed point), $\tp_{p\beforep} \,{=}\, [1\,\,0]$ (before-point), or $\tp_{p\afterp} \,{=}\, [0\,\,1]$ (after-point).
\item[]
{(SV+AP)} When combining both kinds of visibility information, before-points $p\beforep$ and after-points $p\afterp$ are given the same sightline vector as their reference point, \ie, $\sv_{p\beforep} = \sv_{p\afterp} = \sv_p$, and we take as input $I_p = (X_p \oplus \sv_p \oplus \tp_p) \in \bR^8$.
\end{itemize}
While holding a similar kind of information, no augmentation can be reduced to the other one.  SVs alone are not enough to place APs, and APs alone, as they are not associated to their observed point in $P$, cannot determine SVs (cf. \figref{fig:visibility}).

\subsection{Modifying an Existing Architecture}
We can adapt most DSR networks to handle visibility-augmented point clouds with only few modifications:
\begin{itemize}[topsep=1pt,itemsep=3pt]
\item
We change the input size (number of channels) of the first layer of the network (generally an encoder), increasing it by 2, 3 or 5, depending on the augmentation.
%
\item We directly add auxiliary points to the point cloud, thus tripling the number of input points. For methods based on neighboring point sampling, we add auxiliary points after sampling for more efficiency.
%
\end{itemize}
\noindent The batch size may need to be adjusted to fit a larger point cloud in memory, but the rest of the network stays unchanged. Its size is mostly unaltered (\eg, +0.005\% for ConvONet \cite{Peng2020}).


\begin{figure*}[t]
	\centering\small
	\newcommand{\mywidth}{0.14\textwidth}
\begin{tabular}{cc|cccc|c}


\rotatebox{90}{\hspace{10mm}Bare}&
\includegraphics[width=\mywidth]{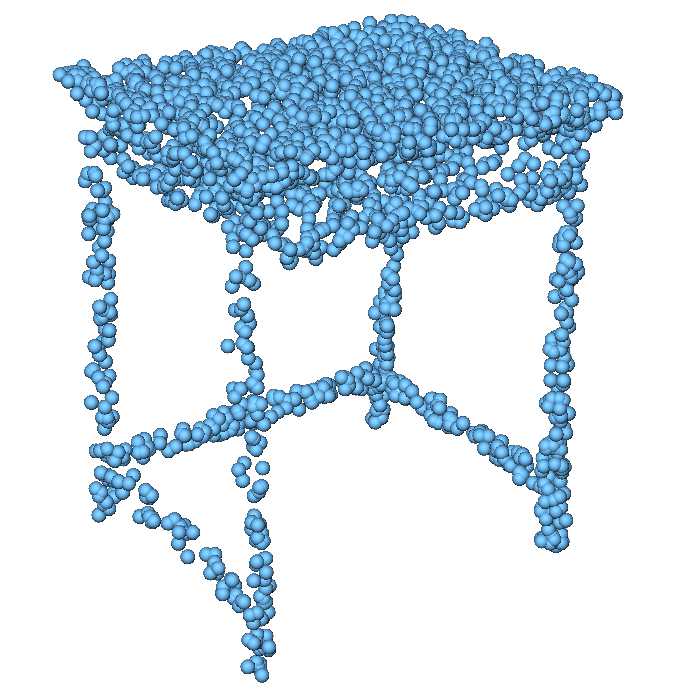}&
\includegraphics[width=\mywidth]{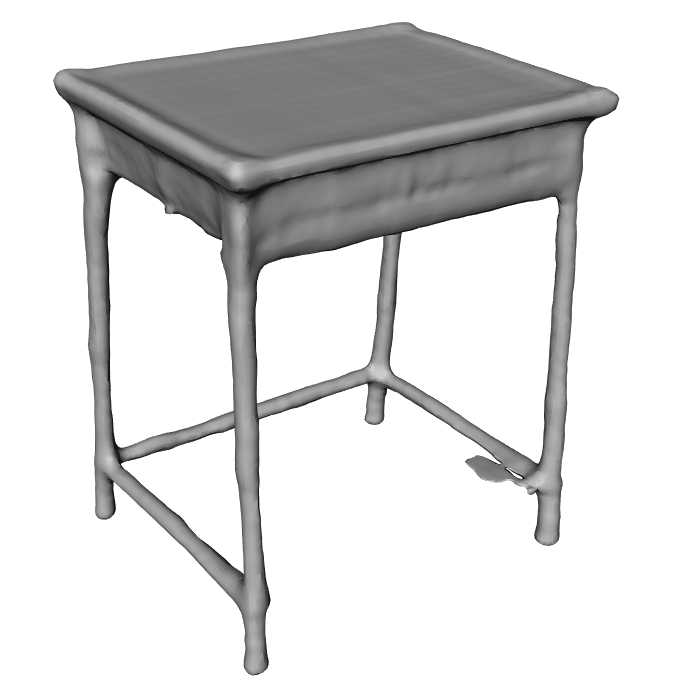}&
\includegraphics[width=\mywidth]{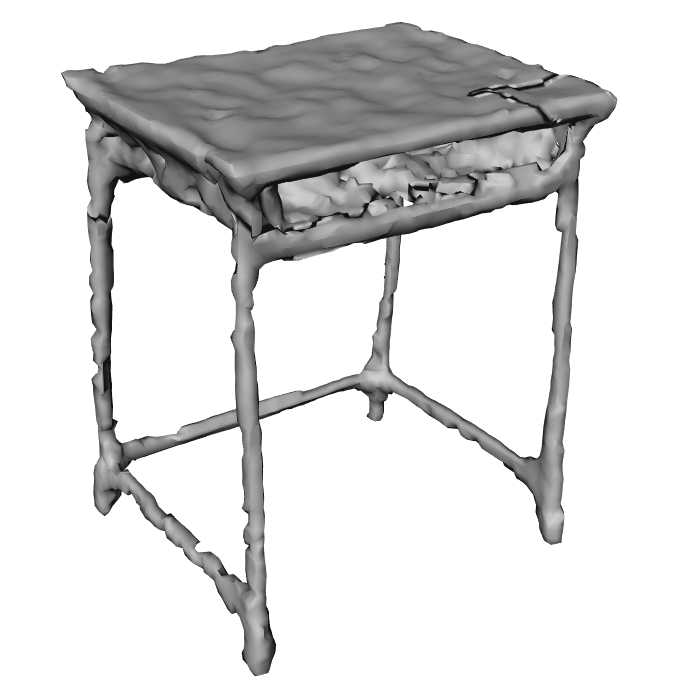}&
\includegraphics[width=\mywidth]{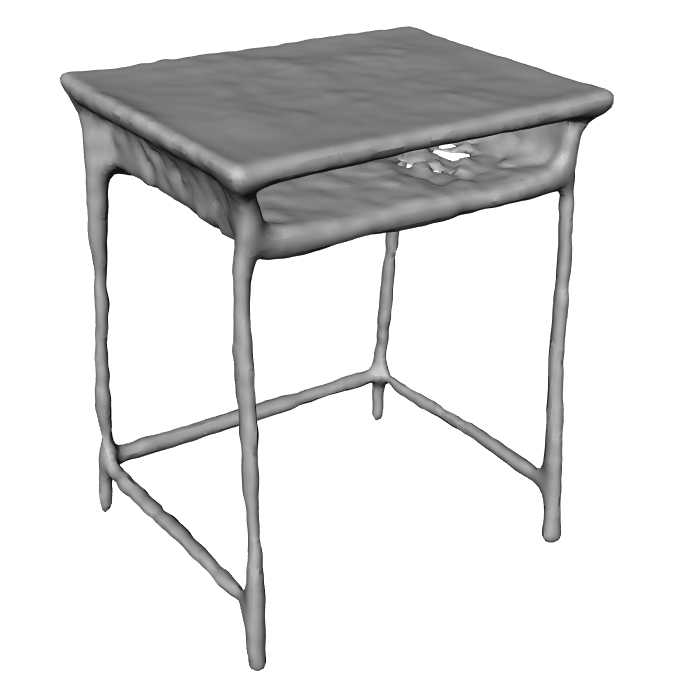}&
\includegraphics[width=\mywidth]{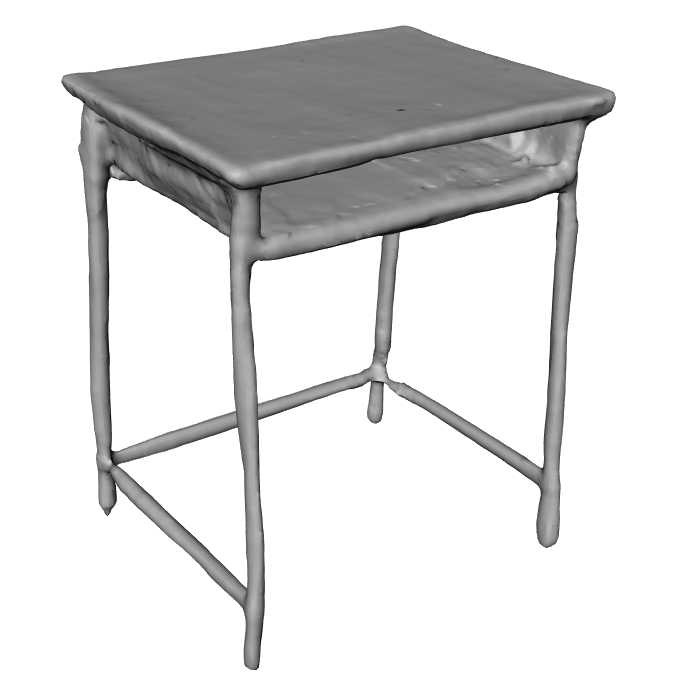}&
\multirow{2}{*}[3em]{\includegraphics[width=\mywidth]{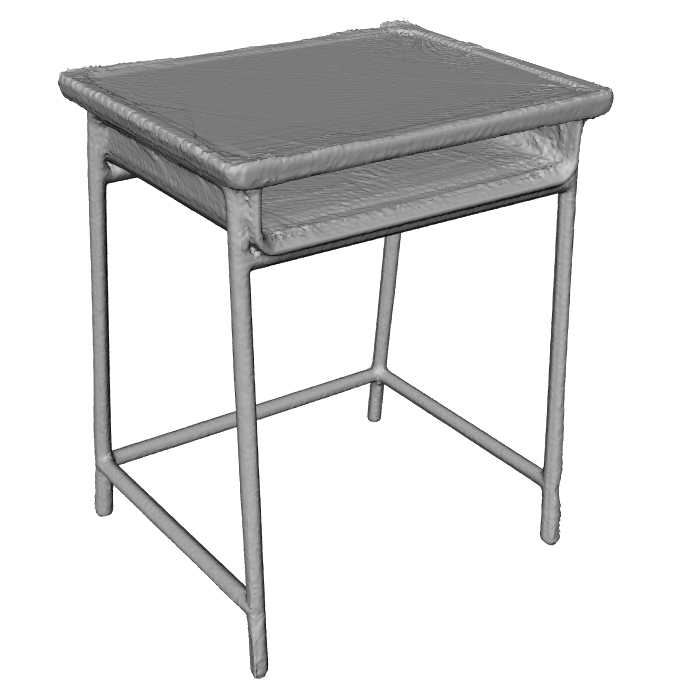}} \\
\rotatebox{90}{\hspace{5mm}Augmented}&
\includegraphics[width=\mywidth]{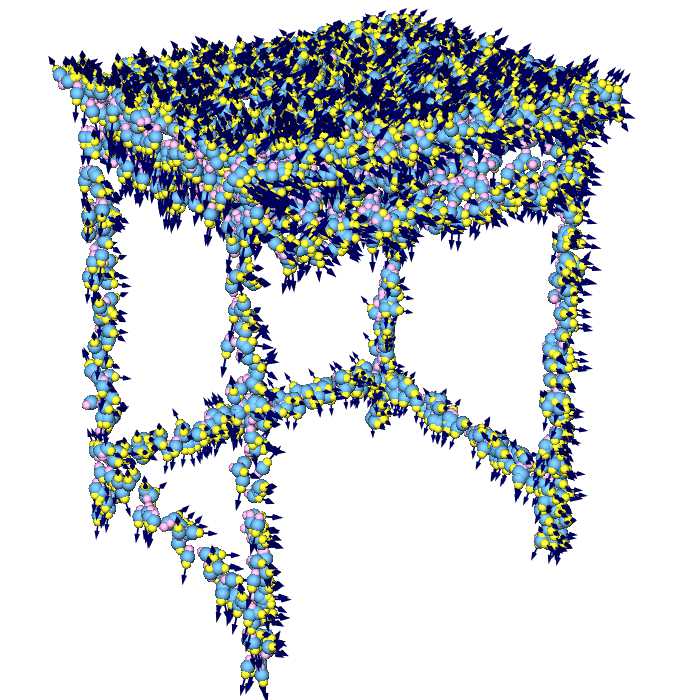}&
\includegraphics[width=\mywidth]{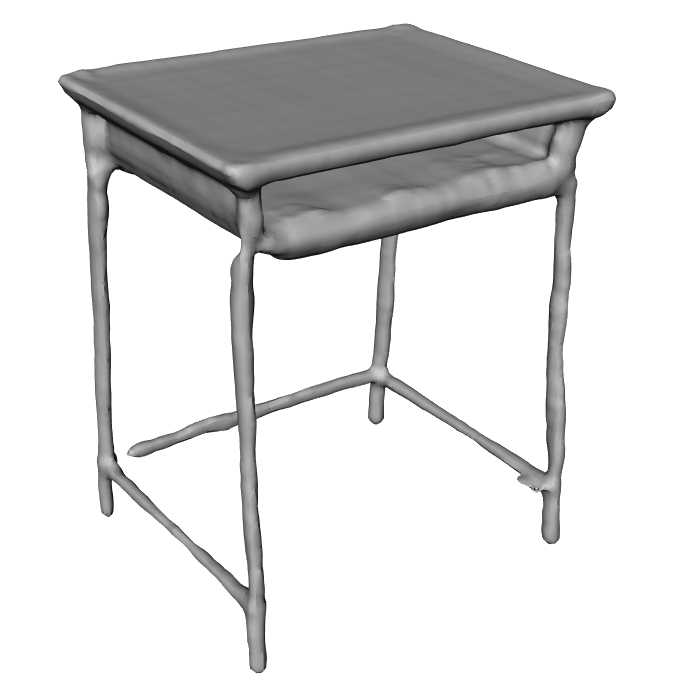}&
\includegraphics[width=\mywidth]{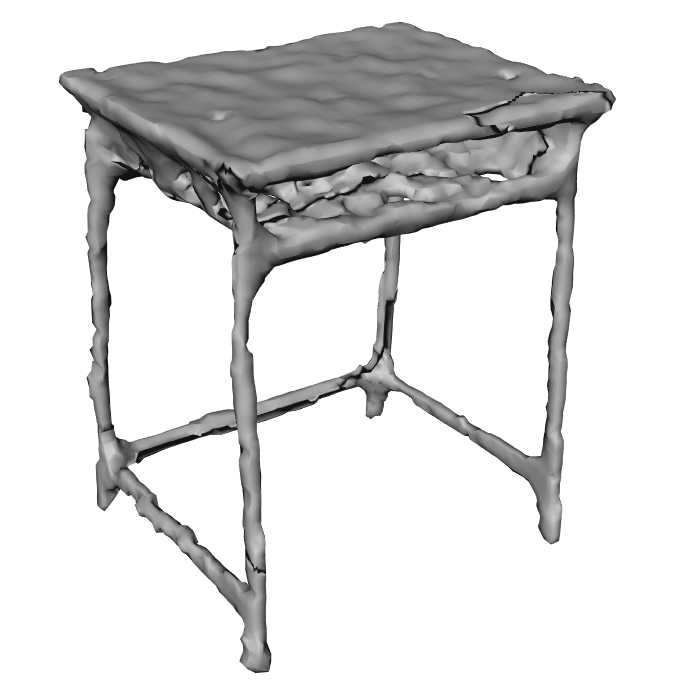}&
\includegraphics[width=\mywidth]{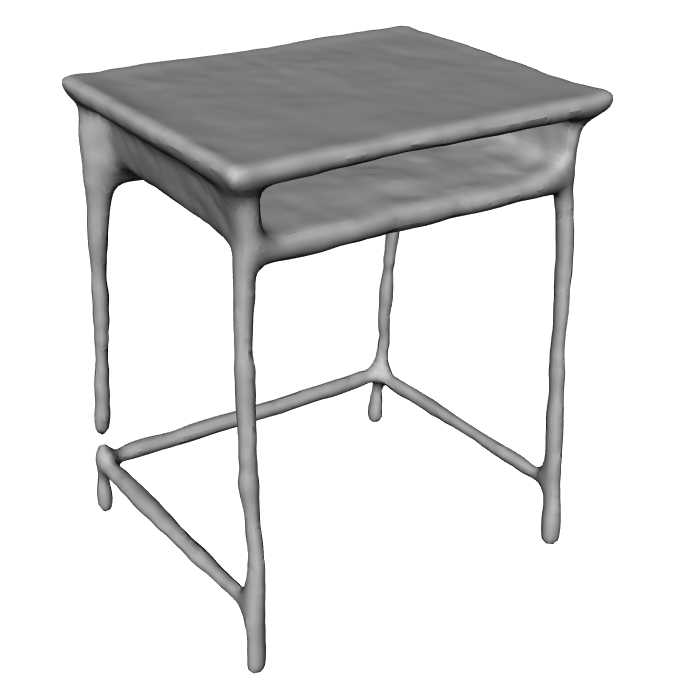}&
\includegraphics[width=\mywidth]{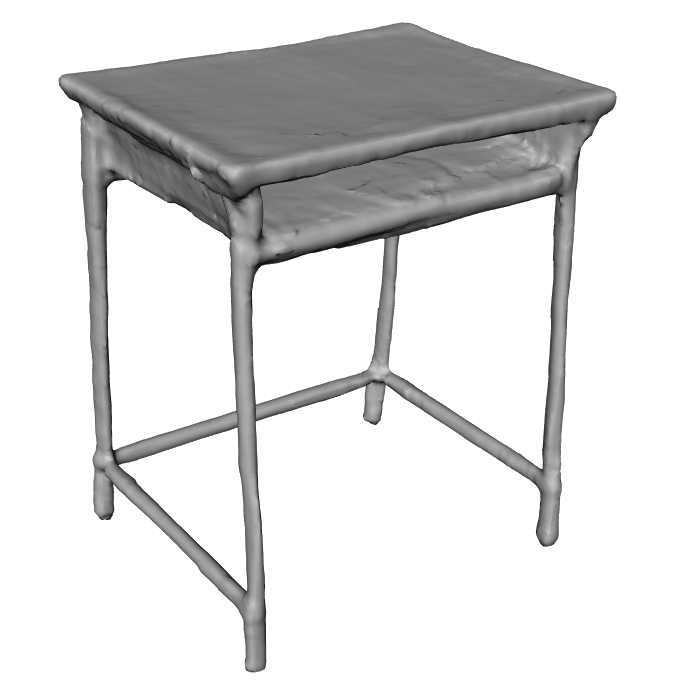}&
\\[3mm]

\rotatebox{90}{\hspace{4mm}Bare}&
\includegraphics[width=\mywidth,trim=0 130 0 200,clip]{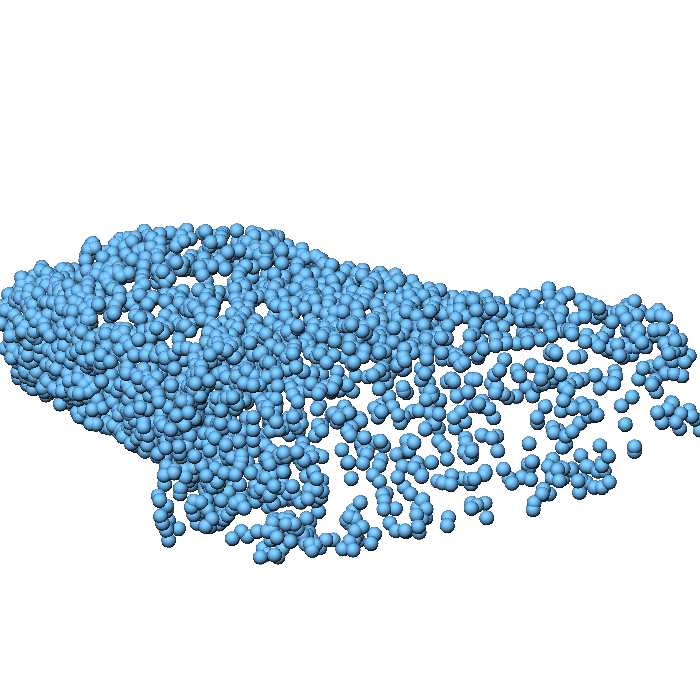}&
\includegraphics[width=\mywidth,trim=0 130 0 200,clip]{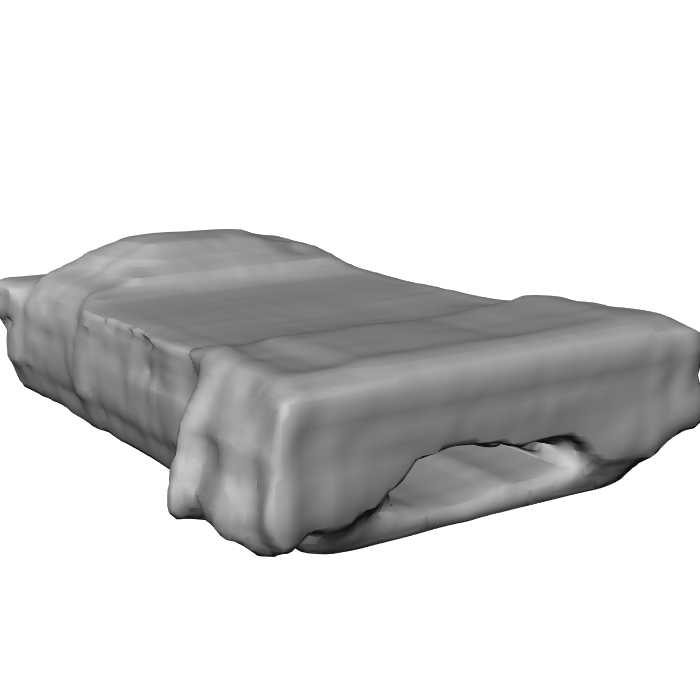}&
\includegraphics[width=\mywidth,trim=0 130 0 200,clip]{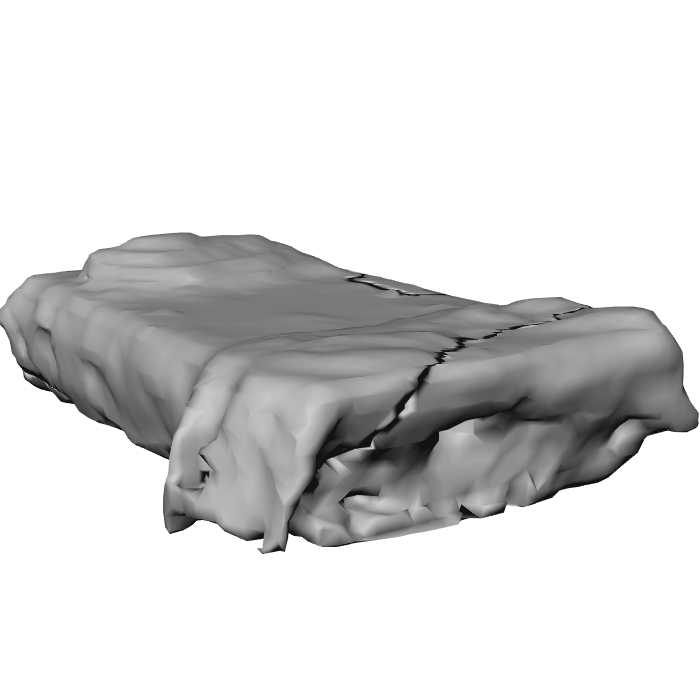}&
\includegraphics[width=\mywidth,trim=0 130 0 200,clip]{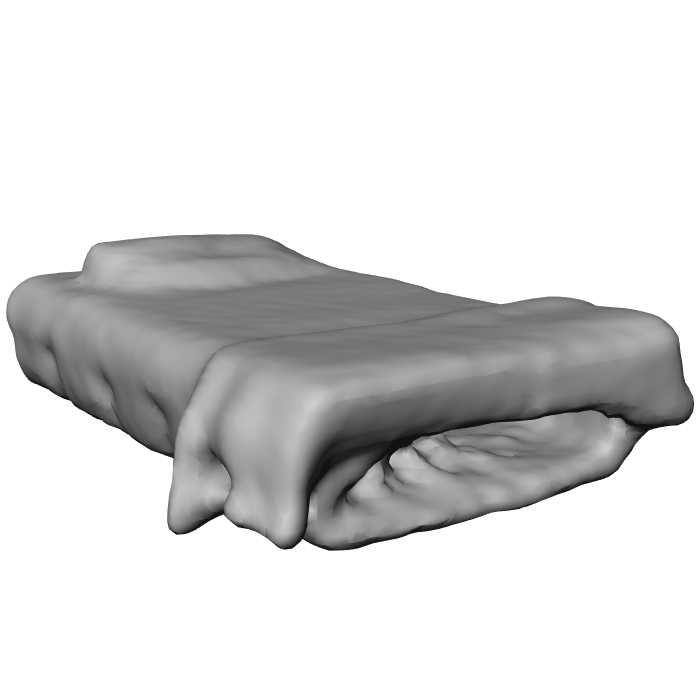}&
\includegraphics[width=\mywidth,trim=0 130 0 200,clip]{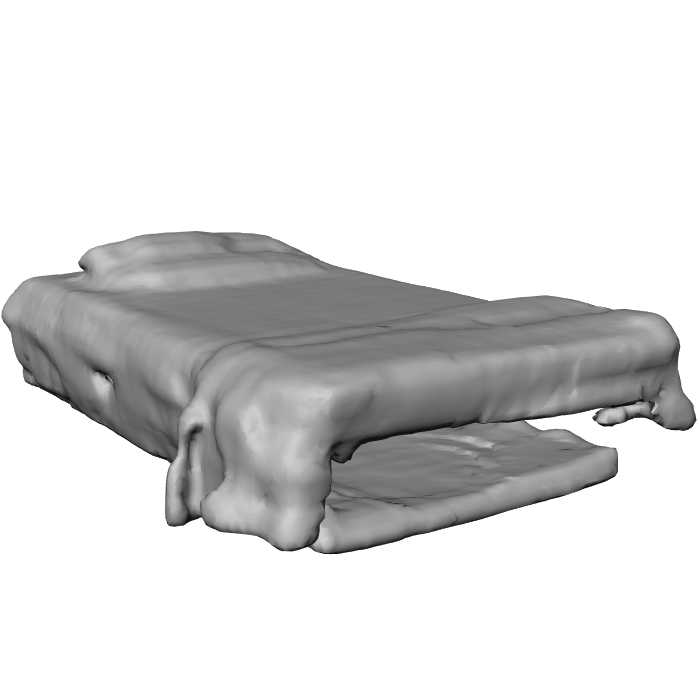}&
\multirow{2}{*}[3em]{\includegraphics[width=\mywidth]{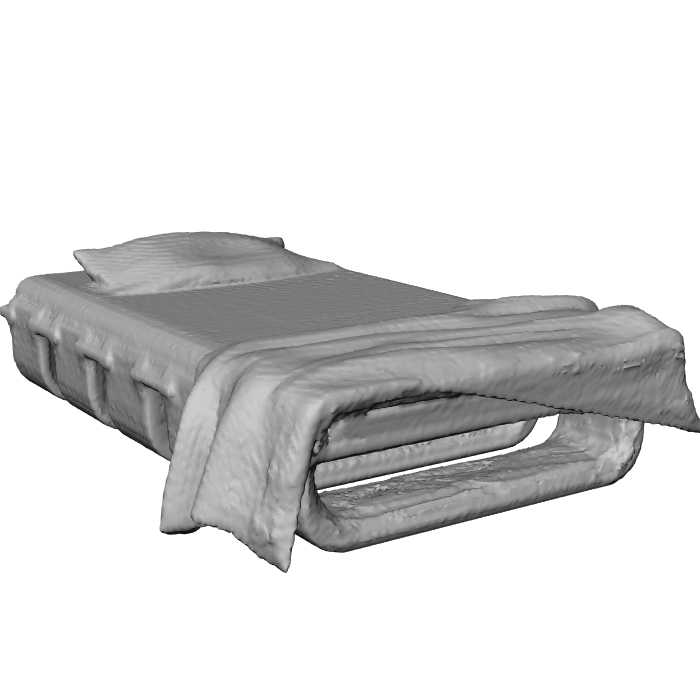}} \\[2mm]
\rotatebox{90}{\hspace{0mm}Augmented}&
\includegraphics[width=\mywidth,trim=0 130 0 200,clip]{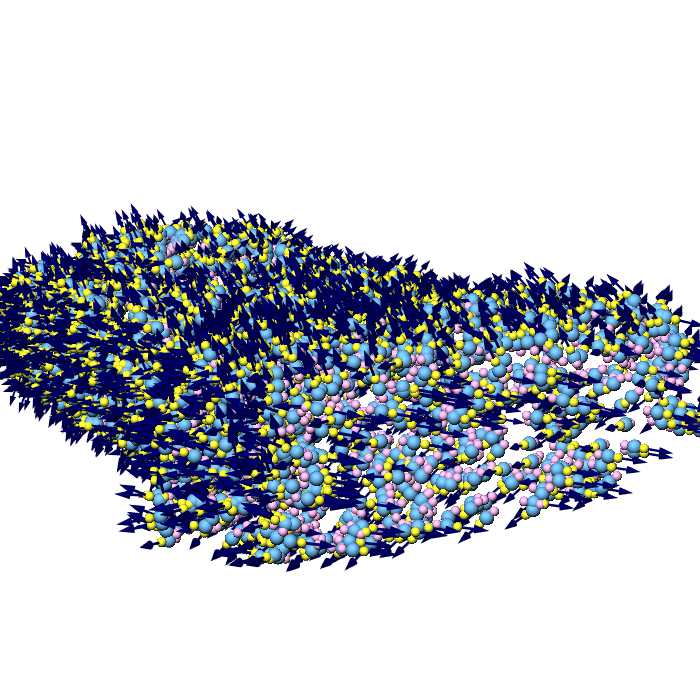}&
\includegraphics[width=\mywidth,trim=0 130 0 200,clip]{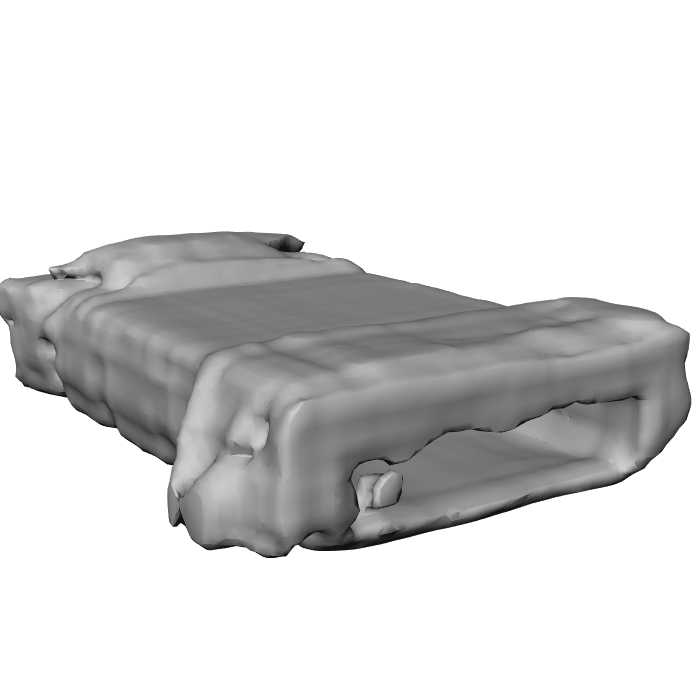}&
\includegraphics[width=\mywidth,trim=0 130 0 200,clip]{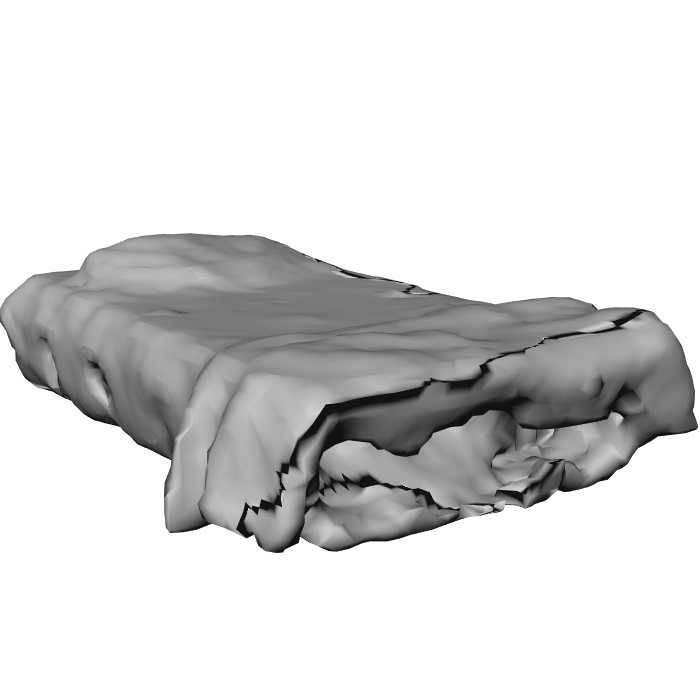}&
\includegraphics[width=\mywidth,trim=0 130 0 200,clip]{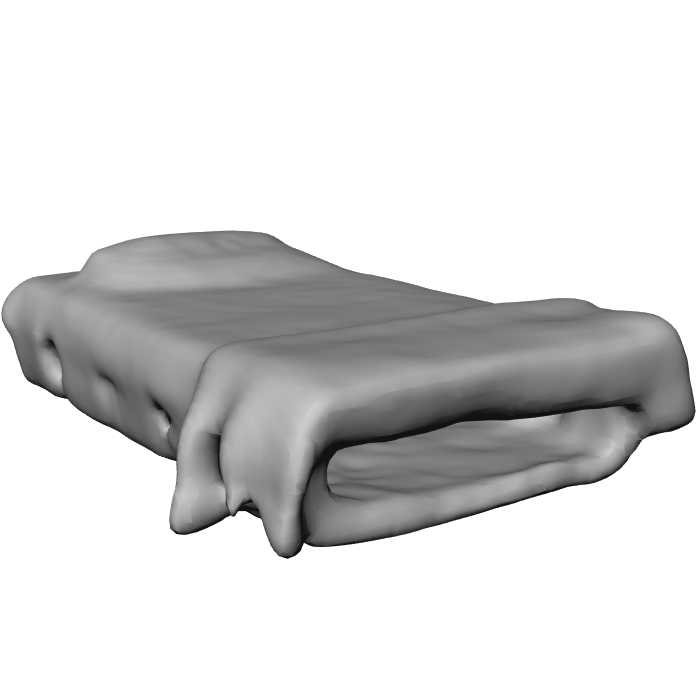}&
\includegraphics[width=\mywidth,trim=0 130 0 200,clip]{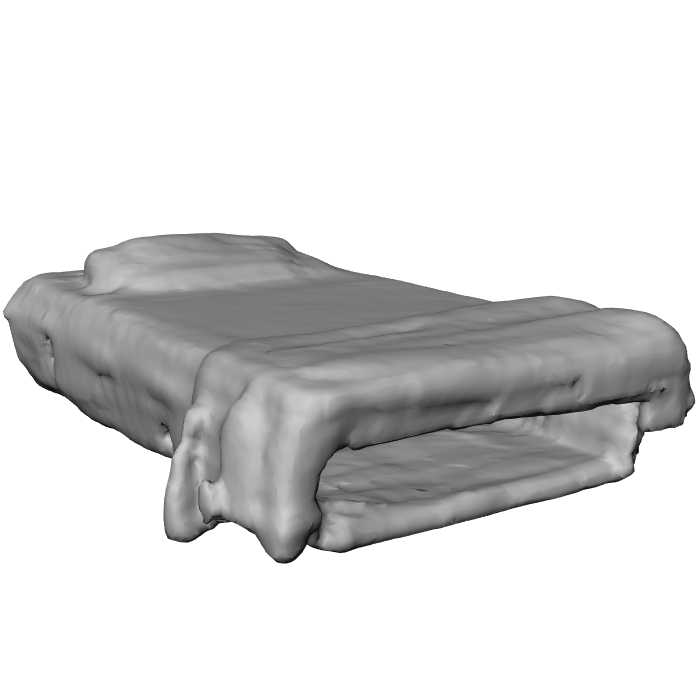}&
\\

&
Input &
ConvONet-2D~\cite{Peng2020}&
Points2Surf~\cite{points2surf}&
Shape\,As\,Points~\cite{Peng2021SAP}&
POCO~\cite{boulch2022poco}&
Ground Truth
\end{tabular}
	\caption{
		\textbf{Object-Level Reconstruction.} 
		Reconstructed shapes from the ModelNet10 test set using four different DSR methods trained on ModelNet10.
		Top rows of each object use the bare point cloud as input, and bottom rows use the point cloud augmented with visibility information.
	}
	\label{fig:modelnet}
\end{figure*}

\section{Experiments}

To assess our proposal, we first detail our simple adaptation of six different DSR baseline networks to leverage our visibility information, then compare the quality of the reconstructed surfaces and analyze the generalization ability of the networks.

\subsection{DSR Baselines}

\subsubsection{ConvONet \cite{Peng2020}}%
This method first extracts point features and projects them on three 2D grids, or one 3D grid (variant). 2D or 3D grid convolutions then create features capturing local occupancy. Last, the occupancy of a query-point is estimated after interpolating grid features.
We consider the $3\times 64^2$ 2D-plane encoder and the $64^3$ 3D-volume variant. {To adapt them, we change the input size of the point encoder's first layer.}

\subsubsection{Points2Surf \cite{points2surf}} 
This method predicts both the occupancy of a query point and its unsigned distance to the surface. It uses both a local query-point neighborhood and a global point-cloud sampling.
We use the best-performing variant (uniform global sampling, no spatial transformer). To adapt it, we increase the input size of the first layer of both the local and global encoders, and when a point is sampled, locally or globally, we add its two auxiliary points on the fly.


\subsubsection{Shape As Points \cite{Peng2021SAP}} 
For each input point, the method estimates its normal as well as $k$ point offsets that are used to correct and densify the point cloud. The resulting point cloud of size $k|P|$ is then fed to a differentiable Poisson solver \cite{screened_poisson}. 
{To adapt the method, we change the input size of the first layer of the encoder, and of the normal and offset decoders as they also input the point cloud.}
We directly add auxiliary points as input, whose normal and offsets will thus be computed too.


\subsubsection{Local Implicit Grids (LIG) \cite{lig}}
This method trains an auto-encoder from dense point cloud patches. For inference, a given sparse patch with oriented normals is first augmented, close to our idea, with 10 new points along each normal; then reconstruction uses latent vectors minimizing a decoder-based training loss, {and a post-processing removes falsely-enclosed volumes}. As training code is unavailable, we use the model pretrained on ShapeNet (without noise). For oriented normals, we use Jets \cite{jet-normal} oriented with a minimum spanning tree \cite{schertler2017towards}, as in \cite{Zhao2021SignAgnostic}. To exploit visibility, we replace normals with sightline vectors; we do not add (more) auxiliary points.

\subsubsection{POCO \cite{boulch2022poco}} 
This method extracts point features using point cloud convolution \cite{boulch2020convpoint}, then estimates the occupancy of a query point with a learning-based interpolation on nearest neighbors. {To adapt it, we increase the input size of the first layer and add auxiliary points on the fly only in the first layer.}

\subsubsection{DGNN \cite{dgnn}}
This method uses a graph neural network to estimate the occupancy of Delaunay cells in a point cloud tetrahedralization. A graph-cut-based optimization then reinforces global consistency. The method, which already uses visibility, outperforms other traditional reconstruction methods that use visibility information. As it already exploits visibility, we do not alter it, but use it as baseline for comparison.


For all methods, unless otherwise stated, training and evaluation are unchanged; we keep the value of the hyperparameters used in the original papers. When Marching cubes \cite{lorensen1987marching} are needed for surface extraction, we use a grid resolution of $128^3$.

\subsection{Datasets}
We consider {a variety of} object and scene datasets, {both synthetic and real}, to show the versatility of our approach.

\subsubsection{ModelNet10}
We use the official train/test splits of all 10 object classes of ModelNet10 \cite{wu20153d}.
We hold out 10\% of the train set for validation. We synthetically scan the models by placing $10$ {random} range scanners in two bounding spheres around the objects and shooting random rays to a sphere inscribed within {the convex hull of} the object. We sample $3\,000$ points per object and add Gaussian noise with zero mean and standard deviation $0.005$ as in \cite{Peng2020}.

\subsubsection{ShapeNet}
We study the generalizability of models trained on ModelNet10 by testing on 100 shapes per class from the ShapeNet \cite{shapenet2015} test set of Choy \etal \cite{choy20163dr2n2}   ($9$ out of $13$ classes are not represented in ModelNet10).
We use the same scanning procedure as for ModelNet10.
  
\subsubsection{Synthetic Room}
We use the train/val/test splits of Synthetic Rooms \cite{Peng2020}. For virtual scanning, we only place sensors in the upper hemispheres, and scan $10\,000$ points as in \cite{Peng2020}.

\subsubsection{SceneNet} We test on a few synthetic scenes of SceneNet \cite{handa2016scenenet} using the given virtual scans, voxel-decimated to 1\,cm.

\subsubsection{ScanNet} We test on a few real scenes of ScanNet \cite{Dai2017CVPRa} using the provided real RGB-D scans, voxel-decimated to 2\,cm\rlap.

\subsubsection{Tanks and Temples} We use the real LiDAR point cloud of the \emph{Ignatius} statue from the Tanks\,and\,Temples dataset \cite{tanksandtemples}.

\subsubsection{Middlebury} We use an MVS point cloud of the \emph{TempleRing} from Middlebury \cite{middlebury}, made with OpenMVS \cite{cernea2015openmvs}.

\smallskip

\subsection{Metrics}\!We report volumetric intersection over union (IoU), mean Chamfer distance $\times 100$ (CD) and normal consistency (NC).

\subsection{Ablation Study}

\begin{table}[]
\caption{
	\textbf{Ablation Study.} 
}
\vspace*{-3mm}
The vanilla model of ConvONet trained and tested on ModelNet10 with different ways to add visibility or normal information.
\vspace*{2mm}

\centering
\begin{tabular}{l@{~~}l@{}lc}
\toprule
Model       & SV            & AP           &       IoU  $\uparrow$ \\ \midrule
ConvONet-2D ($3\times64^2$)~\cite{Peng2020}  &                                  &       &    0.853  \\
+ sightline vectors (SV) only &        \checkmark       &                &    0.871       \\
+ auxiliairy points (AP) only&                          & \checkmark     &    0.881 
\\
+ both SV and AP  &        \checkmark       & \checkmark     &    \bf 0.886   \\
    \midrule
+ sensor position &         $S_p$       &                &    0.870        \\
+ unnormalized SV &        $S_p - X_p$           &                &    0.870       \\\midrule
+ estim.\ normals / estim.\ orientation & Jets \cite{jet-normal} / MST \cite{schertler2017towards}                     &      & 0.853     \\
+ estim.\ normals / sensor orientation & Jets \cite{jet-normal} / sensor-base\rlap{d \cite{schertler2017towards}}                    &                &    0.868     \\
+ true normals & GT normals  &                                  &    0.879      \\\bottomrule
\end{tabular}
\label{tab:ablation}
\vspace*{-1mm}
\end{table}

To validate our design, we compare in \tabref{tab:ablation} various ways to add visibility information to the vanilla model of ConvONet.

Independently, SVs and APs significantly improve performance (+$1.8$ and +$2.8$ IoU pts). A reason why APs are more profitable could be that the network is tailored for points, not points with sightline features. While SVs and APs capture a similar kind of information, they are, however, complementary: combining them is even more beneficial (+$3.5$ IoU pts). Our general interpretation is that SVs help to decide whether a locally ``thin'' point cloud is to be considered as a noisy scan of a single surface, or as a (less noisy) scan on both sides of a thin surface. They thus have an impact on local shape topology, which can bring a notable gain.
Auxiliary points convey similar information, but also contribute more directly to refine the surface position.
Replacing SVs by the sensor position or by the unnormalized point-sensor vector gives essentially 
the same performance than our unit vector.
This can be explained by the fact that our scanning procedure does not introduce significant variation in terms of distance to the sensor.
Yet, for real world acquistions, with a larger 
range of sensor distances, normalizing the SV ensures more stability.

Adding SVs outperforms estimated normals \cite{jet-normal} with estimated orientation \cite{schertler2017towards}, and even estimated normals with sensor-based orientation. While using ground-truth normals is slightly more beneficial than SVs, combining SV+AP yields the best overall performance, which highlights the richness of our visibility information.

We also experiment with adding more than two auxiliary points: (i) at distance $0.5d$ or $2d$, (ii) at the midpoint between sensor and point, or (iii) as grazing points, estimated by densely sampling the sightlines with auxiliary points and keeping the ones close to an input point. None of these strategies brought significant improvements over adding two points at distance $d$ on both sides of the real point.



\begin{table}
\caption{
	\textbf{Object-Level Reconstruction.}
}
\vspace*{-3mm}
DSR methods trained and tested on ModelNet10, with and without sightline vectors (SV) or auxiliary points (AP). $^\dagger$\,Trained on ShapeNet.
\vspace*{2mm}

\centering
\begin{tabular}{lcccccc}
\toprule
Model                               & SV          & AP    & IoU  $\uparrow$ & CD  $\downarrow$  & NC  $\uparrow$        \\ \midrule
ConvONet-2D~\cite{Peng2020}         &               &               & 0.853     & 0.618     & 0.934         \\
ConvONet-2D~\cite{Peng2020}         & \checkmark    &               & 0.871     & 0.557     & 0.936         \\
ConvONet-2D~\cite{Peng2020}         & \checkmark    & \checkmark    & \bf 0.886 & \bf 0.518 & \bf 0.943     \\ \midrule
ConvONet-3D~\cite{Peng2020}         &               &               & 0.885     & 0.493     & 0.949         \\
ConvONet-3D~\cite{Peng2020}         & \checkmark    &               & 0.911     & 0.424     & 0.956         \\
ConvONet-3D~\cite{Peng2020}         & \checkmark    & \checkmark    &\bf 0.913  & \bf0.423  & \bf0.957      \\ \midrule
Points2Surf~\cite{points2surf}      &               &               & 0.842     & 0.590     & 0.890         \\
Points2Surf~\cite{points2surf}      & \checkmark    &               & \bf 0.859 & \bf0.544  & 0.896         \\
Points2Surf~\cite{points2surf}      & \checkmark    & \checkmark    & 0.856     & 0.548     & \bf0.897      \\ \midrule
Shape As Points~\cite{Peng2021SAP}  &               &               & 0.903     & 0.438     & 0.948         \\
Shape As Points~\cite{Peng2021SAP}  & \checkmark    &               & 0.907     & 0.430     & 0.950         \\
Shape As Points~\cite{Peng2021SAP}  & \checkmark    & \checkmark    & \bf 0.913 & \bf0.414  & \bf0.953         \\ \midrule
POCO~\cite{boulch2022poco}          &               &               & 0.907     & 0.422     & 0.945         \\
POCO~\cite{boulch2022poco}          & \checkmark    &               & 0.915     & 0.408     & \bf 0.950     \\ 
POCO~\cite{boulch2022poco}          & \checkmark    & \checkmark    & \bf 0.917 & \bf 0.406 & \bf 0.950     \\ \midrule
\rowcolor{MyLightGray}
$^\dagger$\,LIG~\cite{lig}          &               &               & --        & 0.974     & 0.849         \\
\rowcolor{MyLightGray}
$^\dagger$\,LIG~\cite{lig}          & \checkmark    &               & --        & \bf0.880  & \bf0.882      \\ \midrule
DGNN~\cite{dgnn}                    & \checkmark    &               & 0.866     & 0.543     & 0.884         \\
\bottomrule
\end{tabular}
\vspace*{-2mm}
\label{tab:modelnet}
\end{table}

\subsection{Object-Level Reconstruction}

%

\tabref{tab:modelnet} reports the performance on ModelNet10 of various models, with and without sightline vectors or auxiliary points.

ConvONet (both planar and volumetric) gains about +$3$~IoU pts with visibility information. The resulting surface is more accurate, especially in concave parts, as illustrated in \figref{fig:modelnet}. 

Points2Surf improves with sightline vectors, but auxiliary points do not improve further: the sensor vectors are enough to resolve ambiguities for the occupancy estimation, but distance estimation does not further benefit from auxiliary points. 

Shape\,As\,Points benefits from sightline vectors, although not as much as other methods, probably because the model also estimates normals which provide a similar information as visibility. Still, adding auxiliary points further gains +$0.6$ IoU\,pts, yielding more complete and smoother surfaces.

%

POCO similarly benefits +$1$\,IoU\,pts from sightline vectors but not much from the further addition of auxiliary points. While sightline vectors help for surface orientation, POCO is already accurate enough for APs to bring little refinement.

%
{LIG produces poor results, likely because the only available model is trained on ShapeNet, with uniform sampling, little or no noise, and because oriented normals are only estimated. We cannot report IoU because LIG's post-processing creates holes in some objects. Yet, replacing the estimated normals by sightline vectors improves the predicted surface.}

DGNN, which already exploits visibility and outperforms ConvONet-2D and Points2Surf, is outdistanced on this dataset 
by methods that use our augmented point clouds.

Running time is mostly unaffected when adding sightline vectors. The effect of auxiliary points depends on the method. For ConvONet and POCO, it is negligible ($<$\,+2\%);
running time is $\times$\,2.25 for Points2Surf and $\times$\,1.25 for Shape As Points.


\begin{table}[t]
\setlength{\tabcolsep}{2pt}
\caption{
	\textbf{Scene-Level Reconstruction.} 
}
\vspace*{-3mm}
ConvONet trained and tested in sliding-window mode on Synthetic Rooms.
\vspace*{2mm}

\centering
\begin{tabular}{lccccc}
\toprule
Model                       & SV            & AP            & IoU  $\uparrow$   & CD  $\downarrow$  & NC   $\uparrow$       \\ \midrule
ConvONet-3D~\cite{Peng2020} &               &               & 0.805             & 0.598             & 0.906                 \\
ConvONet-3D~\cite{Peng2020} & \checkmark    & \checkmark    & \bf 0.832         & \bf0.569          & \bf0.911                 \\ \bottomrule
\end{tabular}
\label{tab:synthetic_room}
\vspace*{-2mm}
\end{table}

\begin{table}[t]
\caption{
	\textbf{Out-of-Domain Object-Level Reconstruction.}
}
\label{tab:generalize}
\vspace*{-3mm}
DSR methods trained on ModelNet10 and tested on ShapeNet, with and without sightline vectors (SV) or auxiliary points (AP). $^\dagger$\,Trained on ShapeNet.
\vspace*{2mm}

\centering
\begin{tabular}{lccccc}
\toprule
Model                                           & SV            & AP            & IoU $\uparrow$      & CD $\downarrow$    & NC $\uparrow$    \\ \midrule
\rowcolor{MyLightGray}
$^\dagger$\,ConvONet-2D~\cite{Peng2020}         &               &               & 0.852     & 0.560     & 0.929     \\ \midrule
ConvONet-2D~\cite{Peng2020}                     &               &               & 0.685     & 0.979     & 0.878     \\
ConvONet-2D~\cite{Peng2020}                     & \checkmark    &               & 0.667     & 1.042     & 0.833     \\ 
ConvONet-2D~\cite{Peng2020}                     & \checkmark    & \checkmark    & \bf0.750  & \bf0.891  & \bf0.878  \\ \midrule
ConvONet-3D~\cite{Peng2020}                     &               &               & 0.628     & 0.972     & 0.885     \\
ConvONet-3D~\cite{Peng2020}                     & \checkmark    &               & 0.759     & 0.724     & 0.905     \\ 
ConvONet-3D~\cite{Peng2020}                     & \checkmark    & \checkmark    & \bf0.854  & \bf0.554  & \bf 0.925 \\ \midrule
Points2Surf~\cite{points2surf}                  &               &               & 0.807     & 0.561     & 0.876     \\
Points2Surf~\cite{points2surf}                  & \checkmark    &               & \bf0.836  & \bf0.516  & 0.886     \\ 
Points2Surf~\cite{points2surf}                  & \checkmark    & \checkmark    & 0.833     & 0.522     & \bf0.887  \\ \midrule
\rowcolor{MyLightGray}
$^\dagger$\,Shape As Points~\cite{Peng2021SAP}  &               &               & 0.838     & 0.577     & 0.923     \\ \midrule
Shape As Points~\cite{Peng2021SAP}              &               &               & 0.494     & 0.997     & 0.859     \\
Shape As Points~\cite{Peng2021SAP}              & \checkmark    &               & 0.749     & 0.843     & 0.881     \\ 
Shape As Points~\cite{Peng2021SAP}              & \checkmark    & \checkmark    & \bf0.821  & \bf0.617  & \bf0.919  \\ \midrule
POCO~\cite{boulch2022poco}                      &               &               & 0.391     & 1.119     & 0.839     \\
POCO~\cite{boulch2022poco}                      & \checkmark    &               & \bf0.832  & \bf0.618  & \bf0.901  \\ 
POCO~\cite{boulch2022poco}                      & \checkmark    & \checkmark    & 0.815     & 0.635     & 0.887     \\ \midrule
DGNN~\cite{dgnn}                                & \checkmark    &               & 0.844     & 0.549     & 0.854     \\

\bottomrule
\end{tabular}
\vspace*{-1mm}
\label{tab:shapenet}
\end{table}

\subsection{Scene-Level Reconstruction}

To study the impact of visibility information 
at scene level, we train and test ConvONet on Synthetic Rooms, in sliding-window mode \cite{Peng2020}.
We report quantitative results in \tabref{tab:synthetic_room} and qualitative results in the supplementary material. 
The model gains almost +$3$ IoU\,pts with visibility information, showing that benefits scale to scenes, not just objects.



\subsection{Generalization to New Domains}

\begin{figure}[!t]
\setlength{\tabcolsep}{2pt}
\small
\begin{tabular}{ccc}
    &SceneNet & ScanNet\\
    \rotatebox{90}{\qquad\quad Input}
   &
    \includegraphics[width=0.48\linewidth,trim={0cm 0cm 3.5cm 3.5cm},clip]{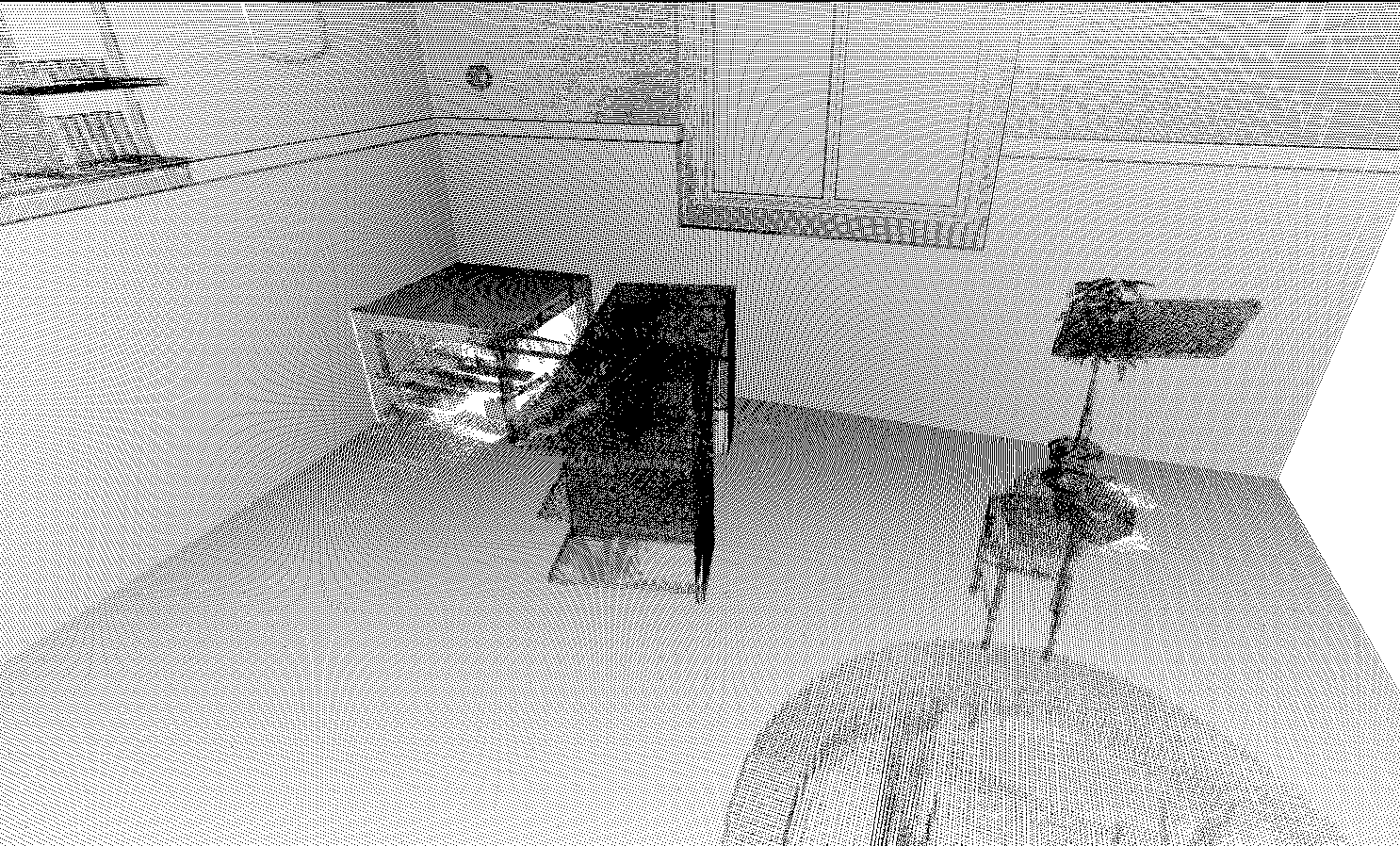}
    &
    \includegraphics[width=0.45\linewidth,trim={0cm 0cm 0.0cm 0.0cm},clip]{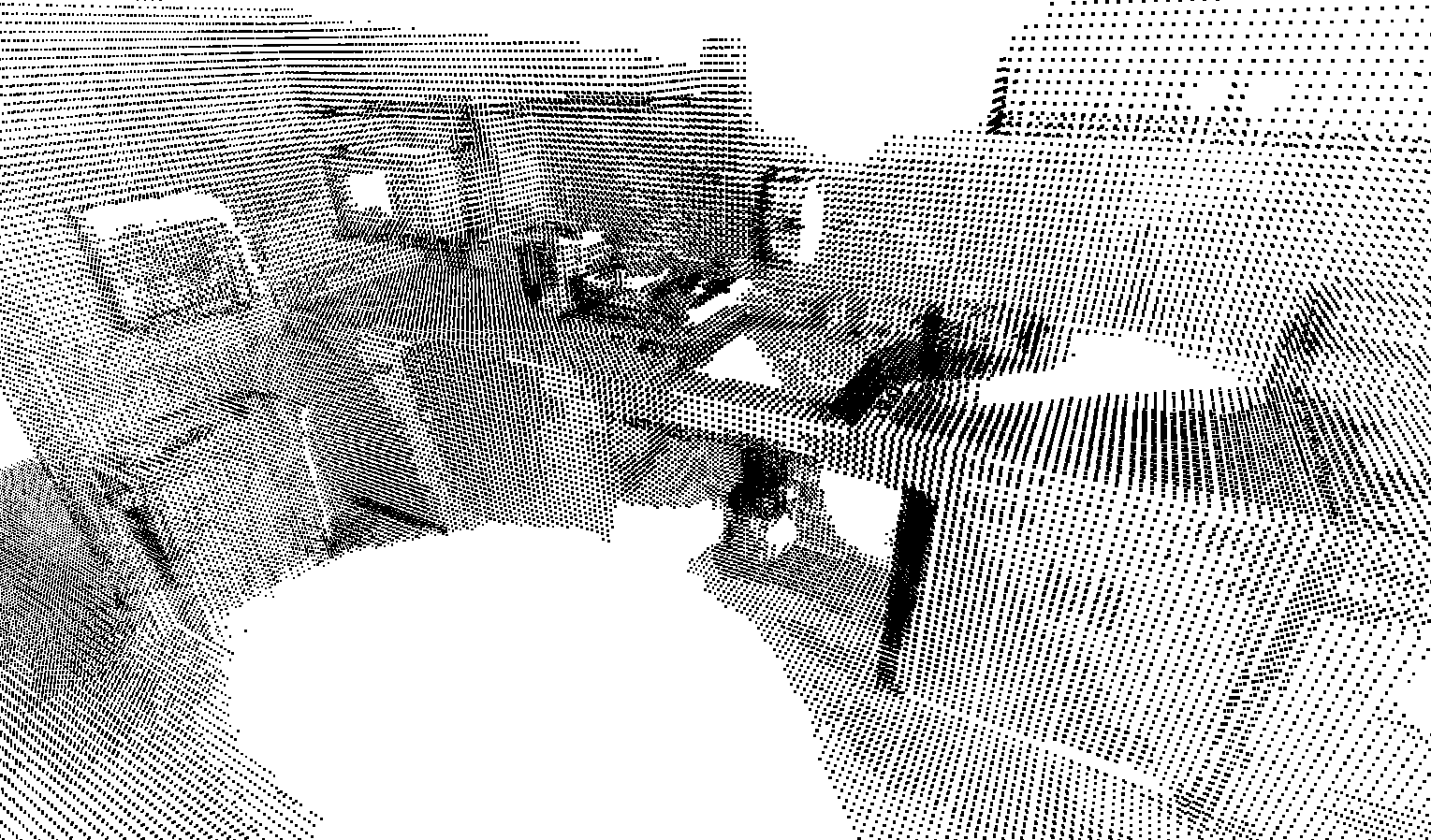}
    \\
    \rotatebox{90}{\quad\qquad Bare}
   &
    \includegraphics[width=0.48\linewidth,trim={0cm 0cm 3.5cm 3.5cm},clip]{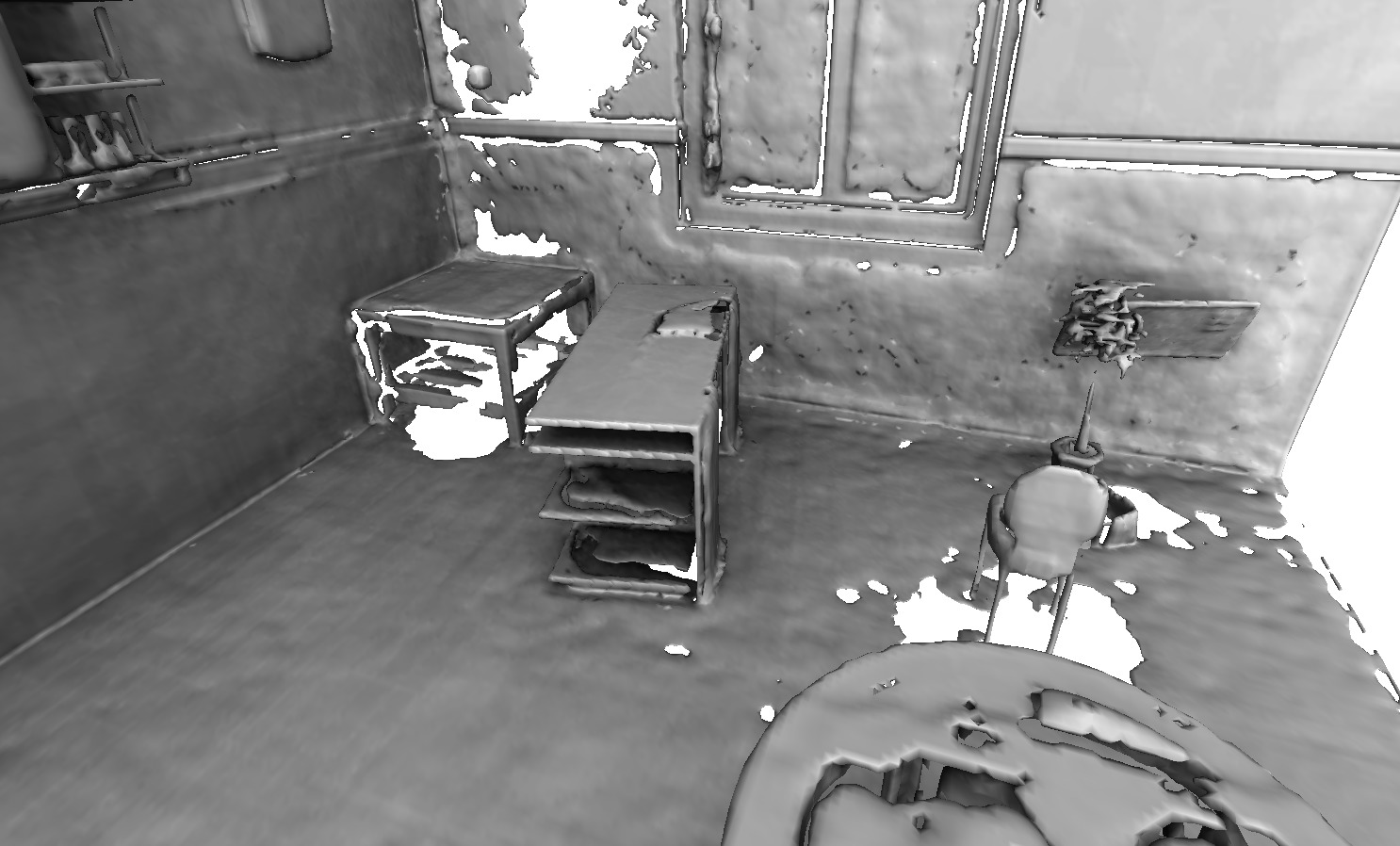}
    &
    \includegraphics[width=0.45\linewidth,trim={0cm 0cm 0.0cm 0.0cm},clip]{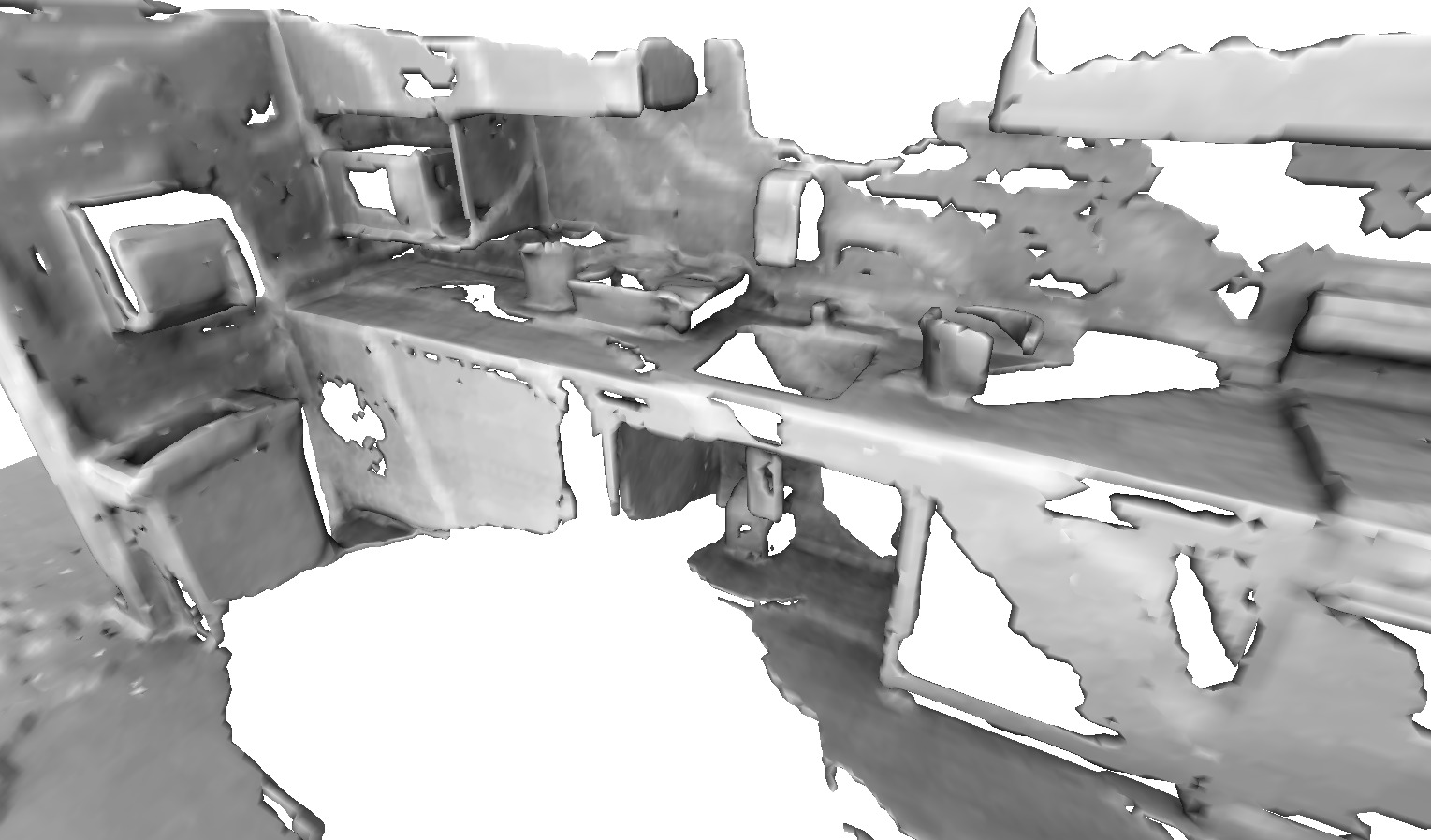}
    \\
    \rotatebox{90}{\quad\; Augmented}
   &
    \includegraphics[width=0.48\linewidth,trim={0cm 0cm 3.5cm 3.5cm},clip]{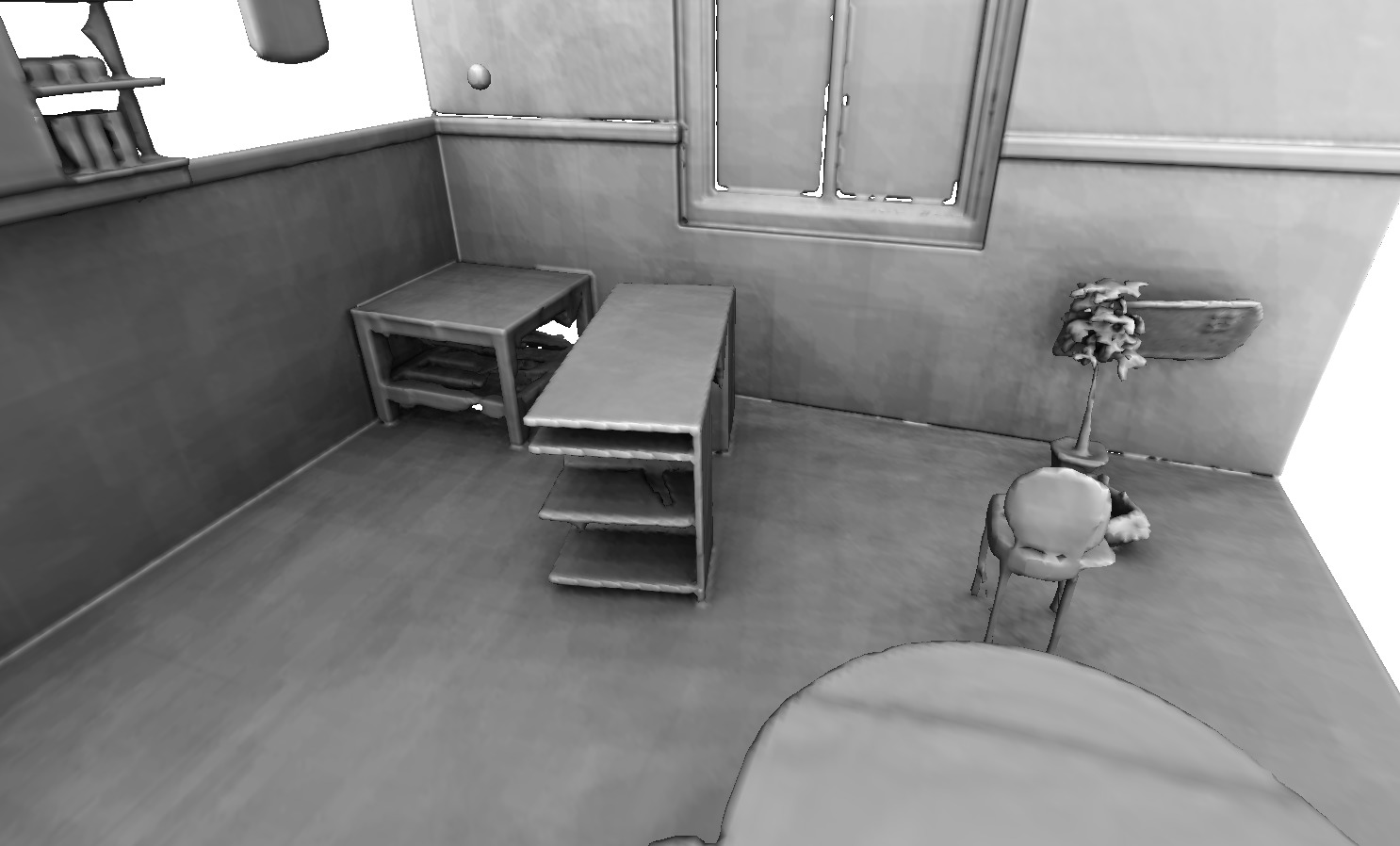}
    &
    \includegraphics[width=0.45\linewidth,trim={0cm 0cm 0.0cm 0.0cm},clip]{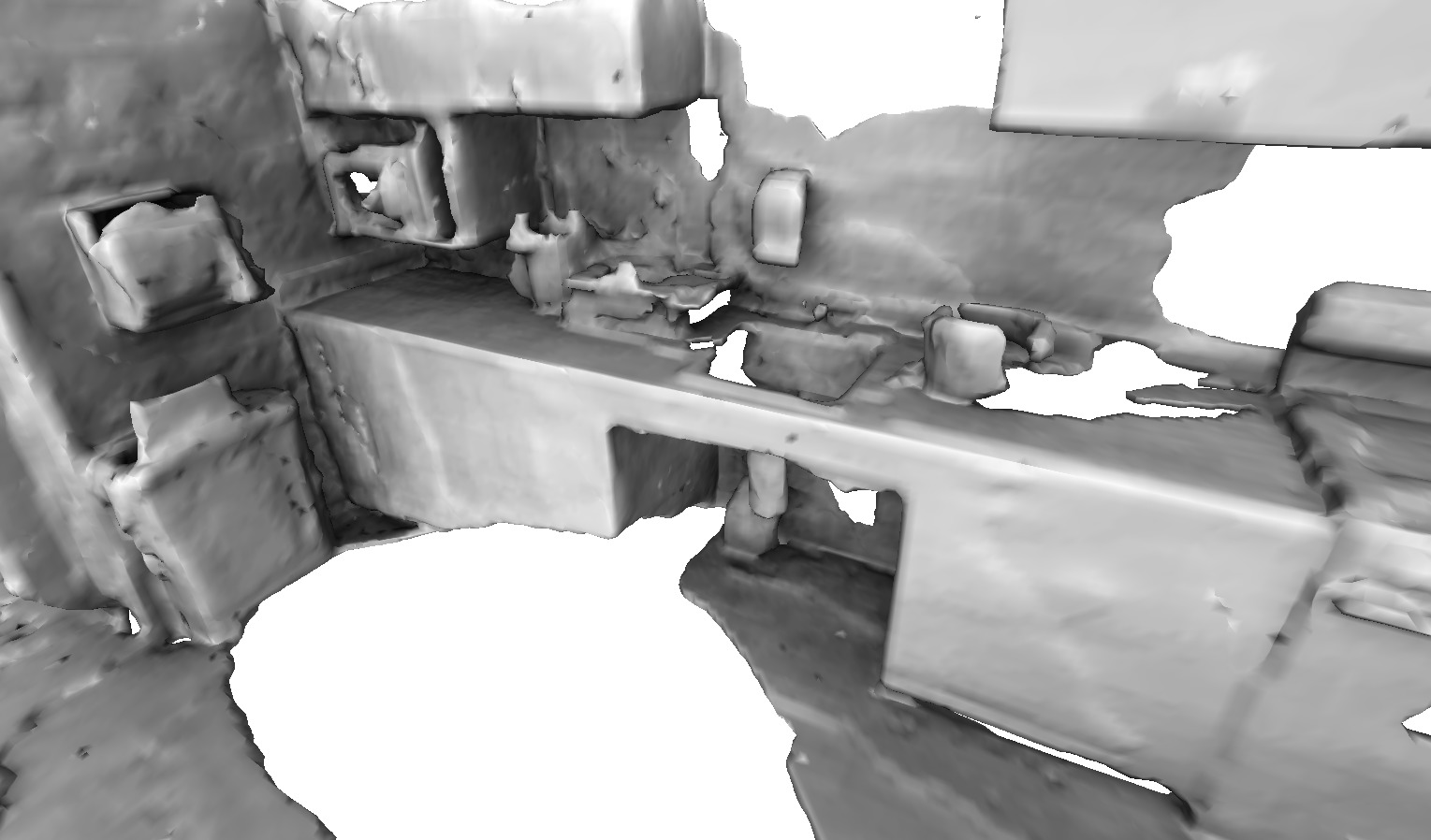}
    \end{tabular}
    \vspace*{-2mm}
\caption{{\bf Out-of-Domain Scene-Level Reconstruction.}
{POCO trained on ModelNet10, with and without visibility information, is run on scenes from SceneNet (synthetic RGB-D scan) and ScanNet (real RGB-D scan).}
}
\vspace*{-1mm}
\label{fig:pocoscenescannet}
\end{figure}

{To evaluate the impact of added visibility on the generalization capability of DSR methods, we train on ModelNet10 and test on ShapeNet (\tabref{tab:generalize}).}

We observe that ConvONet, Shape As Points and POCO trained with visibility information generalize much better on the new objects and classes, with a gain up to +44\,IoU\,pts. For comparison, we also show the scores of official models trained on ShapeNet, although trained on uniformly sampled points rather than virtual scans, which explains the drop of performance compared to the numbers in the papers \cite{Peng2020,Peng2021SAP}.
Points2Surf also improves by up to +3\,IoU\,pts with added sightline vectors, but not further with APs. 
\begin{figure*}[]
	\centering
	\newcommand{\mywidth}{0.14\textwidth}
	\small
\begin{tabular}{cc|cccc|c}



\rotatebox{90}{\hspace{12mm}Bare}&
\includegraphics[width=\mywidth]{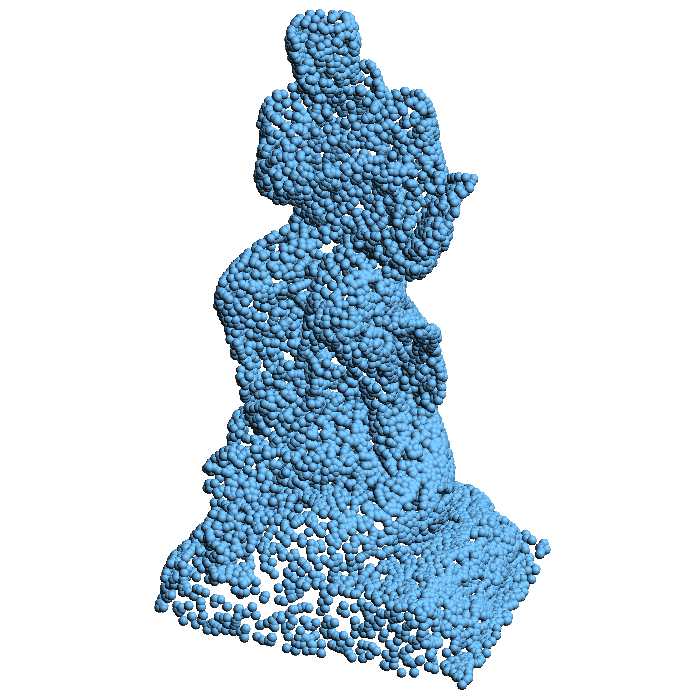}&
\includegraphics[width=\mywidth]{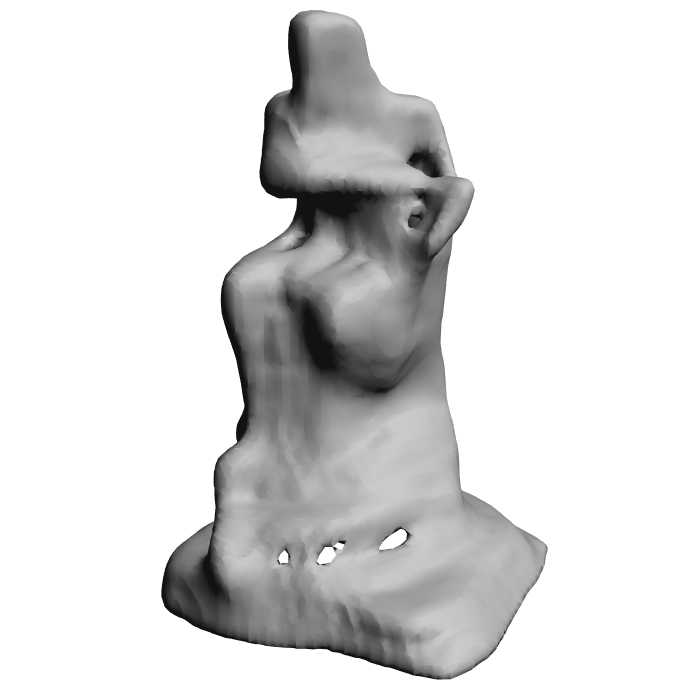}&
\includegraphics[width=\mywidth]{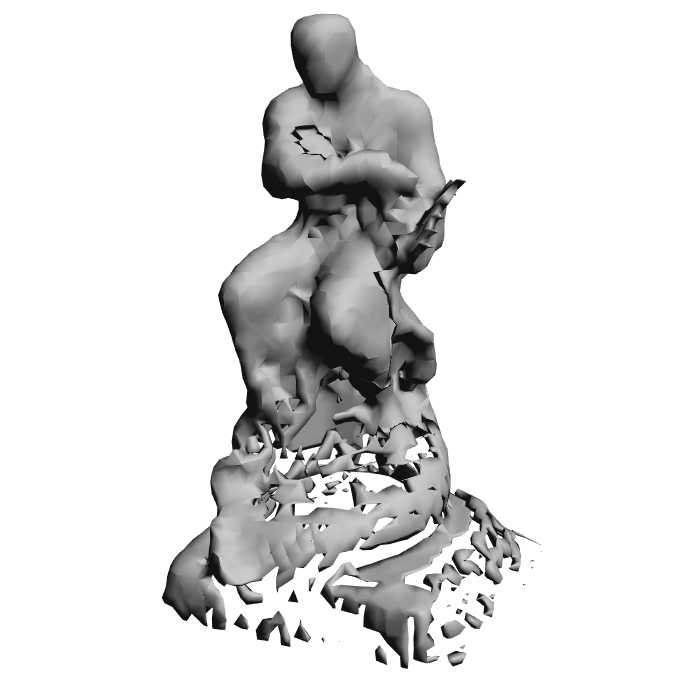}&
\includegraphics[width=\mywidth]{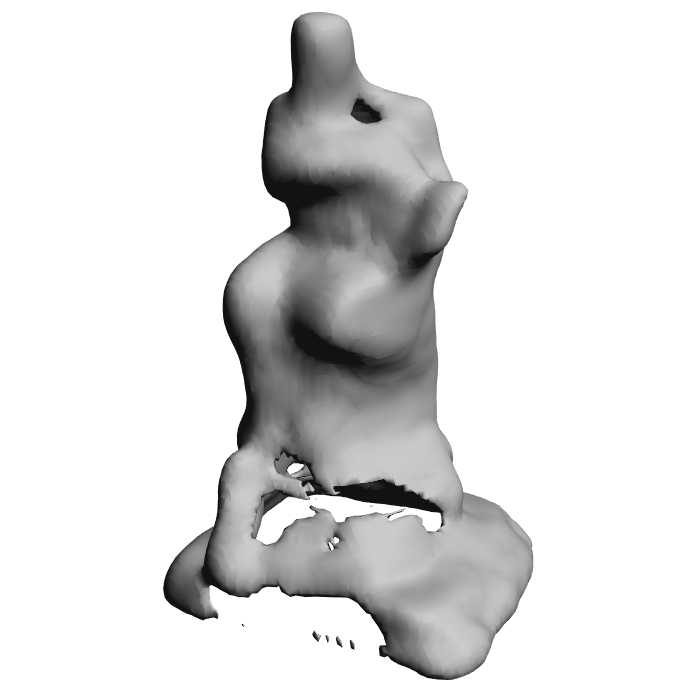}&
\includegraphics[width=\mywidth]{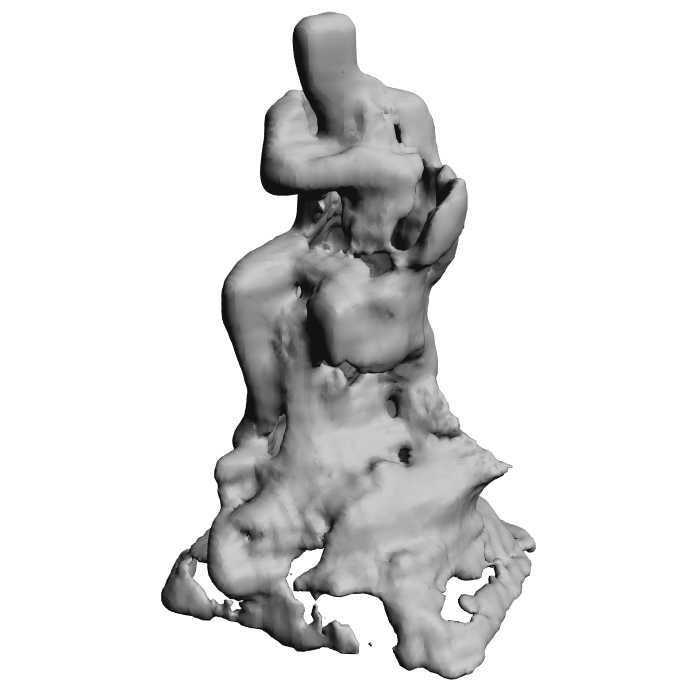}&
\multirow{2}{*}[3em]{\includegraphics[width=\mywidth]{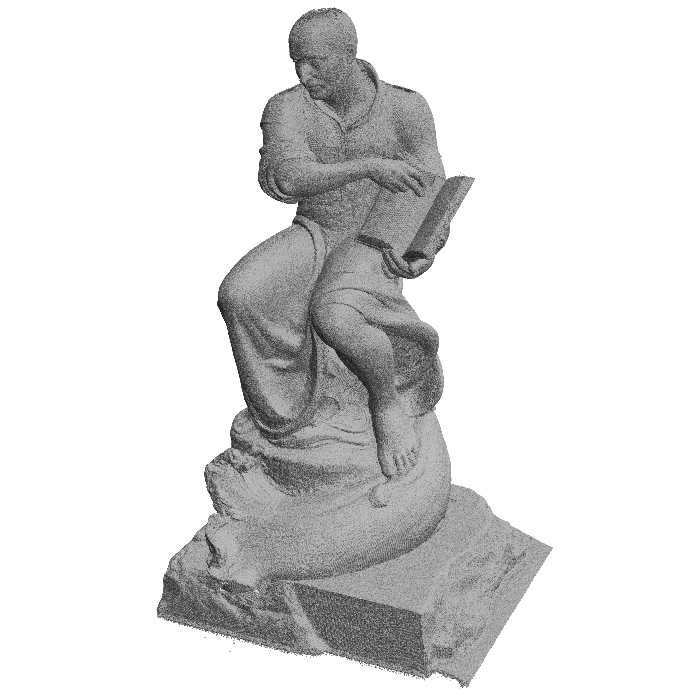}} \\

\rotatebox{90}{\hspace{7mm}Augmented}&
\includegraphics[width=\mywidth]{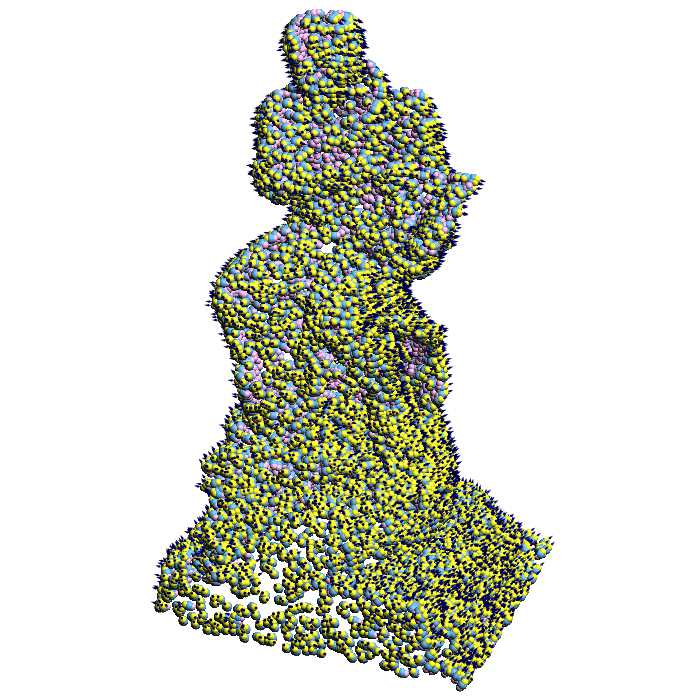}&
\includegraphics[width=\mywidth]{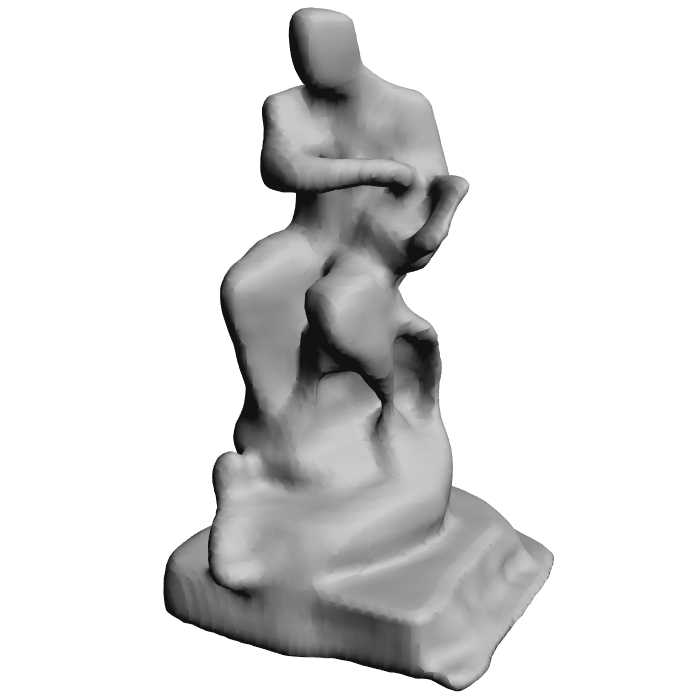}&
\includegraphics[width=\mywidth]{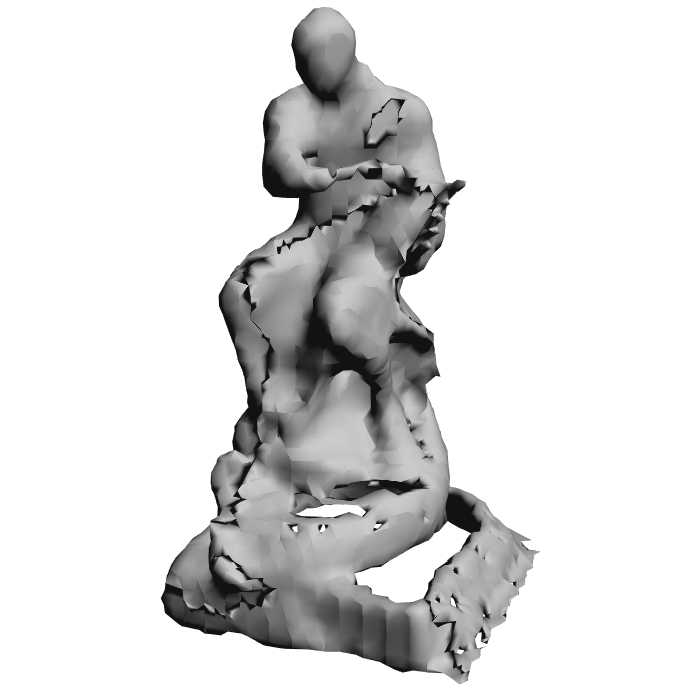}&
\includegraphics[width=\mywidth]{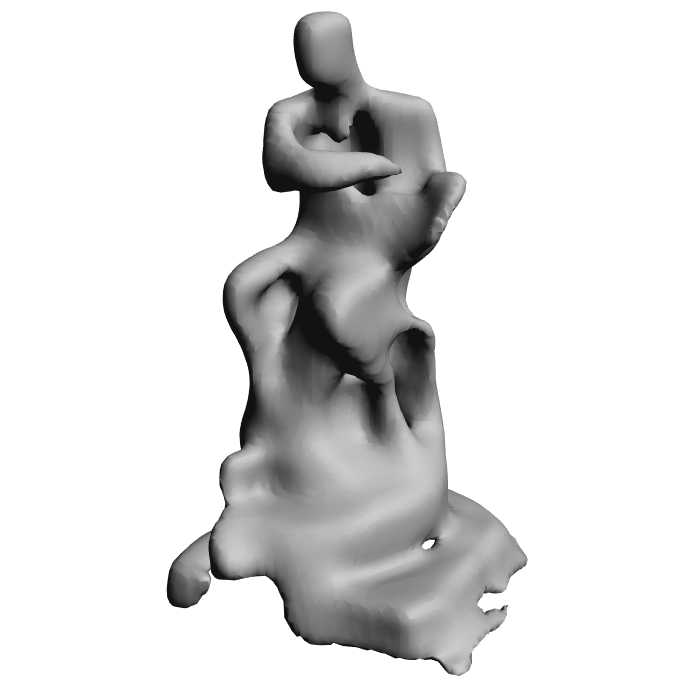}&
\includegraphics[width=\mywidth]{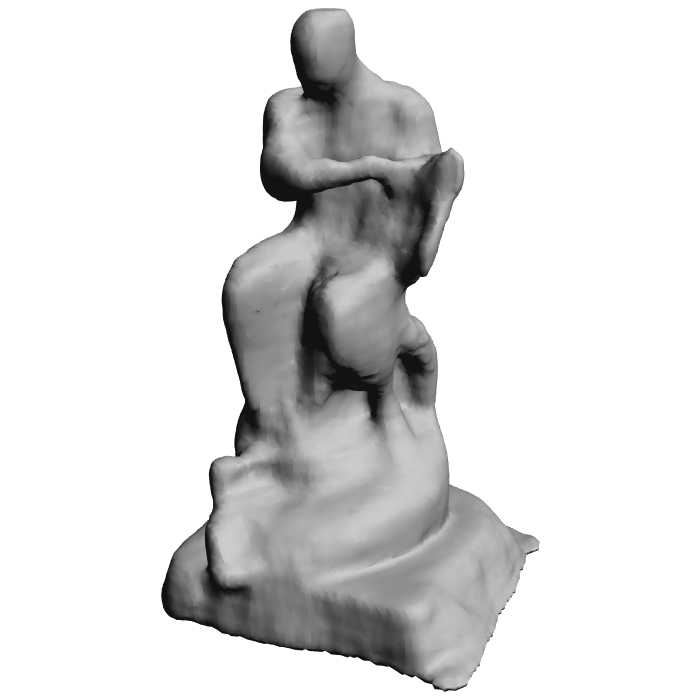}&
\\[2mm]

\rotatebox{90}{\hspace{12mm}Bare}&
\includegraphics[width=\mywidth]{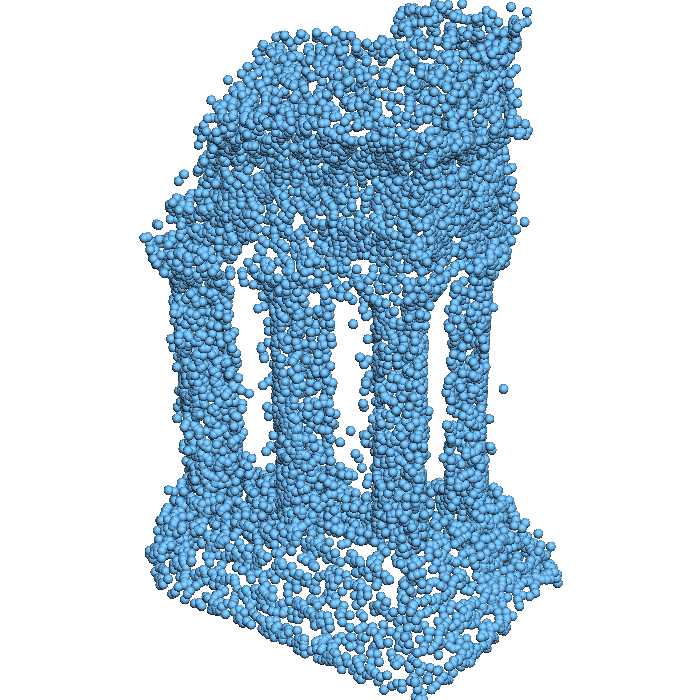}&
\includegraphics[width=\mywidth]{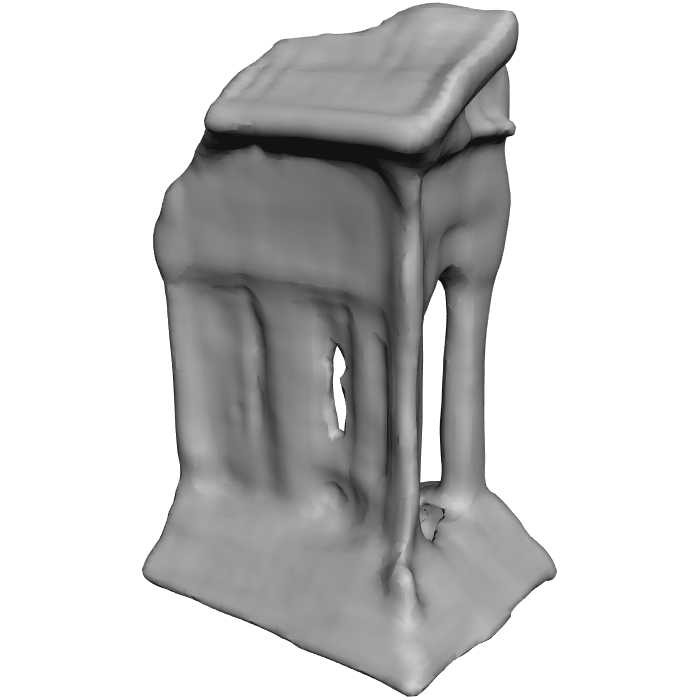}&
\includegraphics[width=\mywidth]{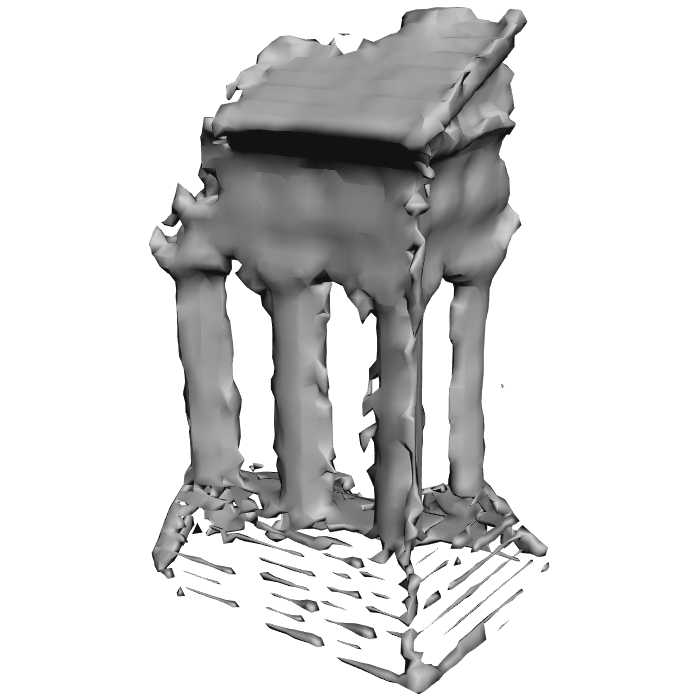}&
\includegraphics[width=\mywidth]{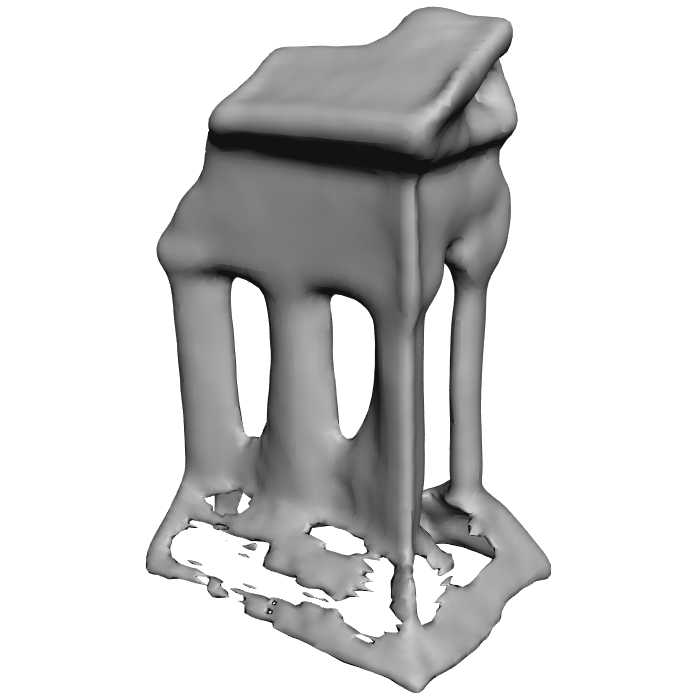}&
\includegraphics[width=\mywidth]{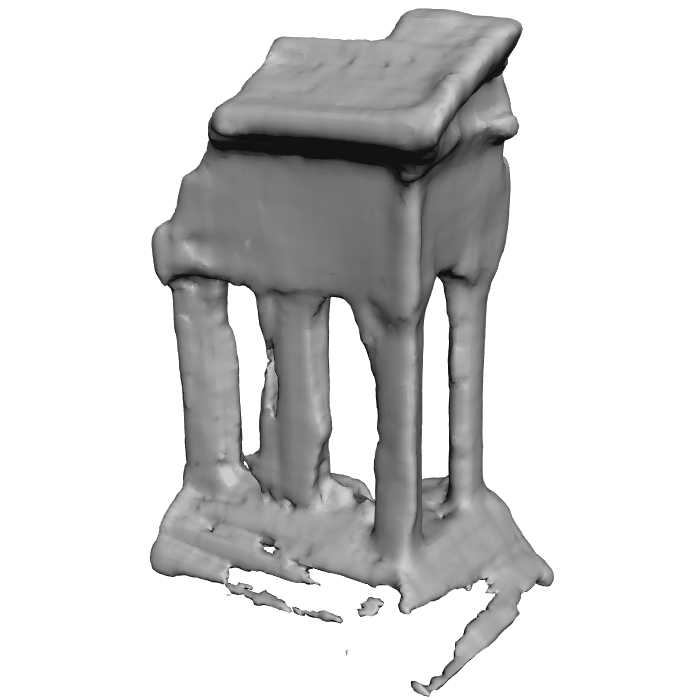}&
\multirow{2}{*}[3em]{\includegraphics[width=\mywidth]{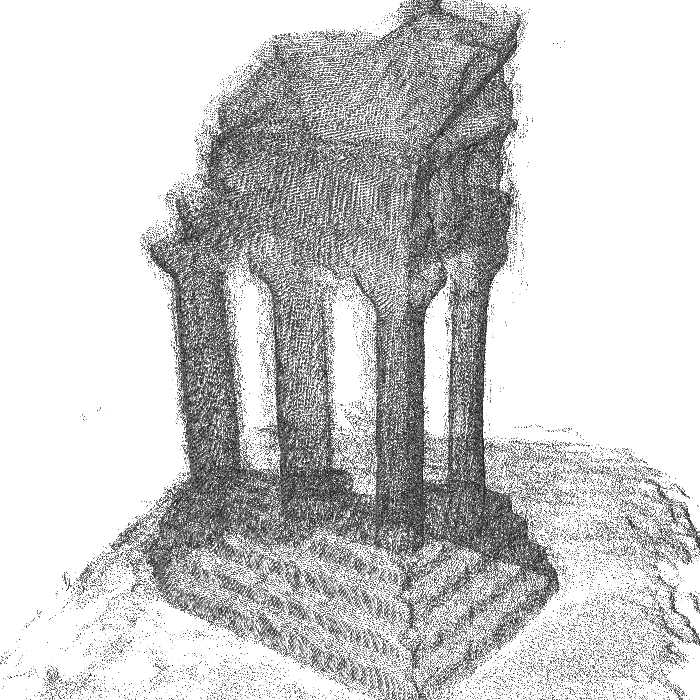}} \\

\rotatebox{90}{\hspace{7mm}Augmented}&
\includegraphics[width=\mywidth]{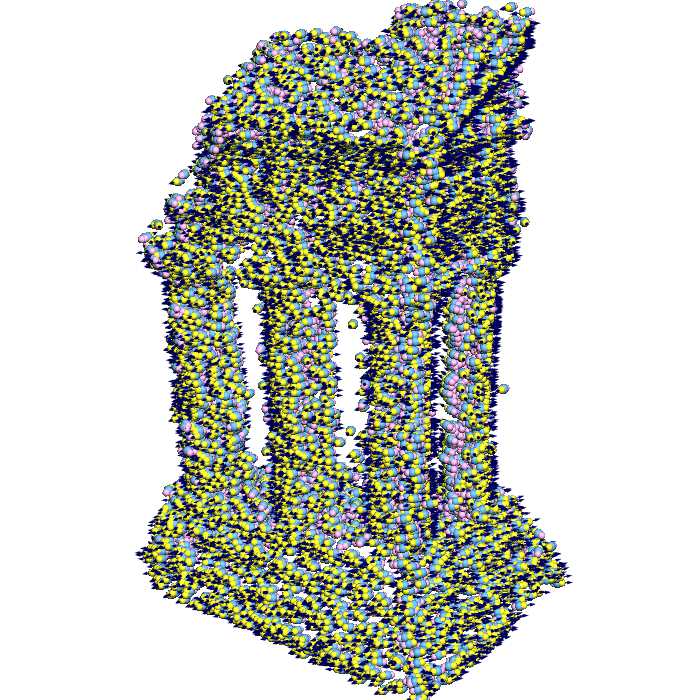}&
\includegraphics[width=\mywidth]{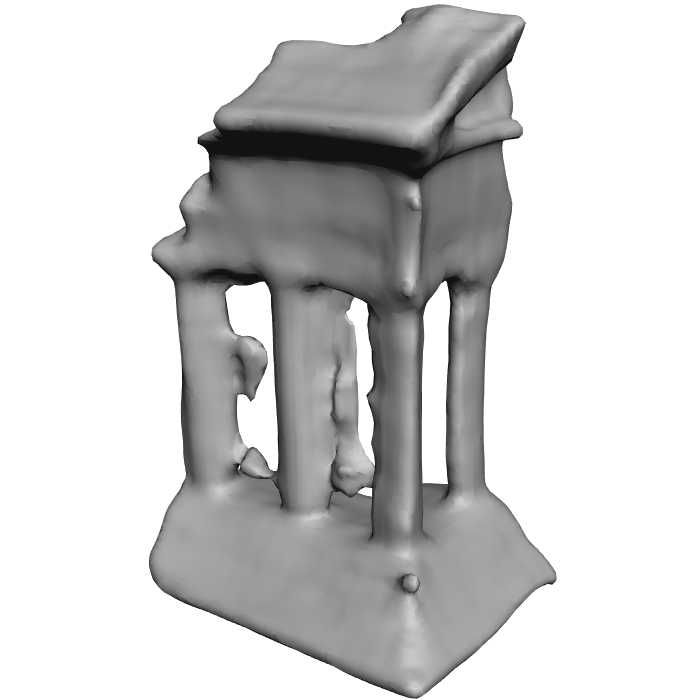}&
\includegraphics[width=\mywidth]{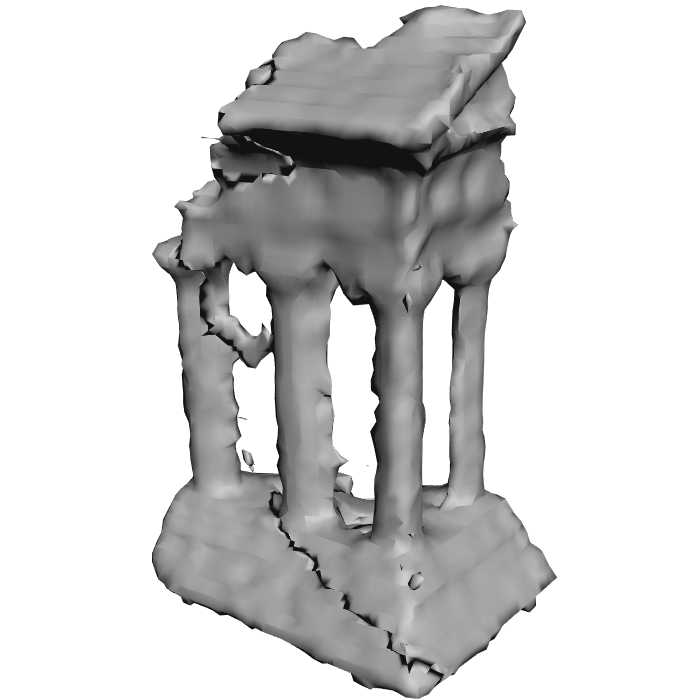}&
\includegraphics[width=\mywidth]{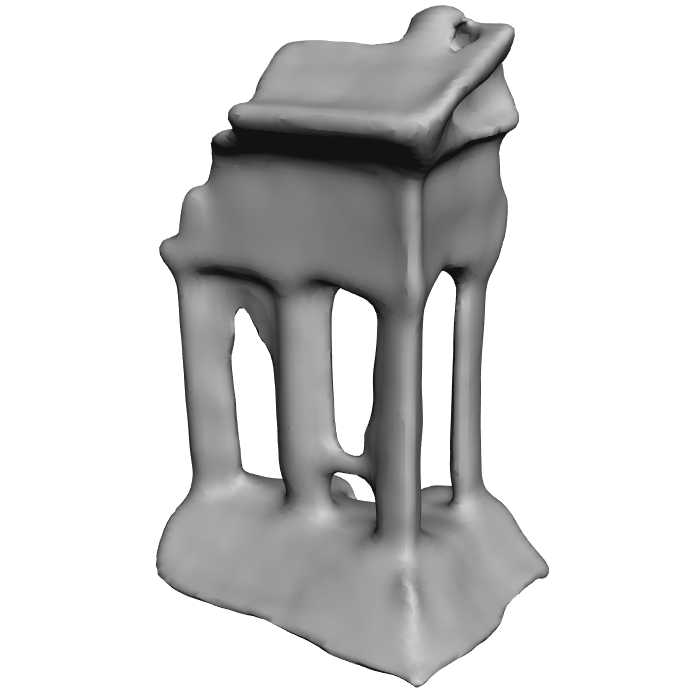}&
\includegraphics[width=\mywidth]{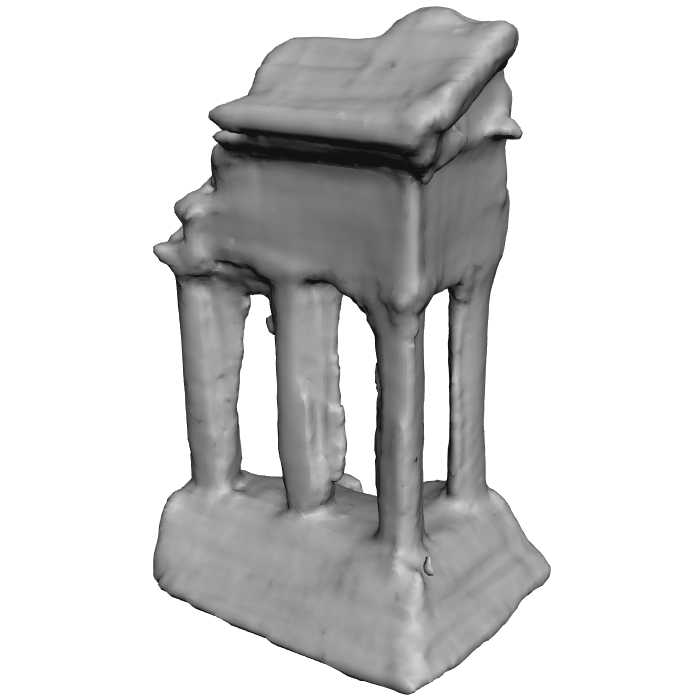}&
\\

&
Input &
ConvONet-3D~\cite{Peng2020} &
Points2Surf~\cite{points2surf} &
Shape\,As\,Points~\cite{Peng2021SAP} &
POCO~\cite{boulch2022poco}&
HD Scan
\end{tabular}
	\caption{
		\textbf{Out-of-Domain Object-Level Reconstruction.}
		Reconstructed shapes from a LiDAR point cloud (top, \emph{Ignatius} from Tanks\,And\,Temples) 
		and a MVS point cloud (bottom, \emph{TempleRing} from Middlebury) using four different DSR methods trained on ModelNet10. 
		Top rows of each object use the bare point cloud as input, and bottom rows use the point cloud augmented with visibility information.
		\emph {HD Scan} is a high-density point cloud.
	}
	\label{fig:real}
\end{figure*}

\begin{figure}
\small
\setlength{\tabcolsep}{2pt}
\begin{tabular}{ccc}

& ConvONet-3D~\cite{Peng2020} & POCO~\cite{boulch2022poco} \\
\rotatebox{90}{\qquad\qquad\quad Bare}&
\includegraphics[width=0.43\linewidth]{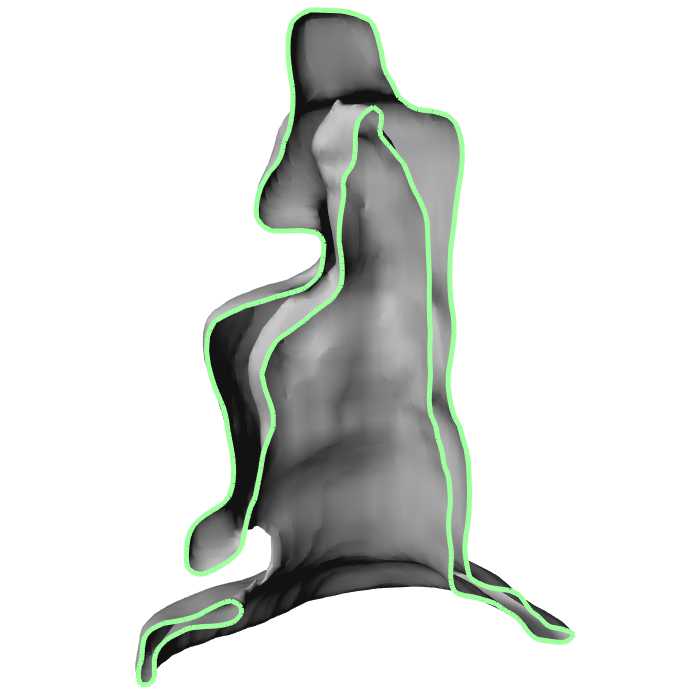}&
\includegraphics[width=0.43\linewidth]{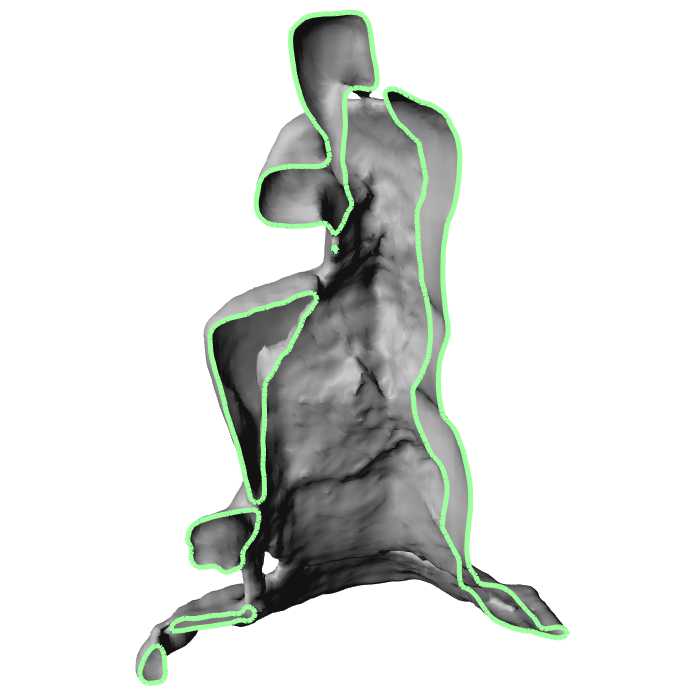}
\\
\rotatebox{90}{\qquad\quad Augmented}&
\includegraphics[width=0.43\linewidth]{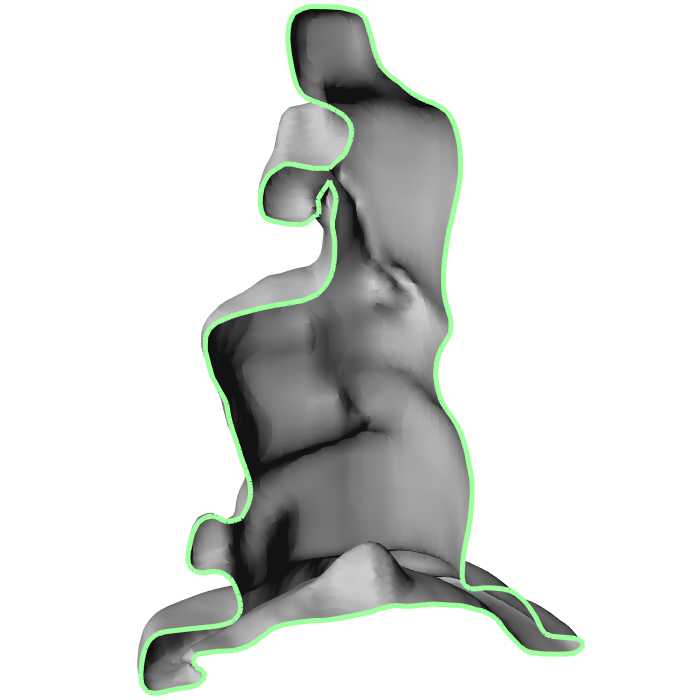}&
\includegraphics[width=0.43\linewidth]{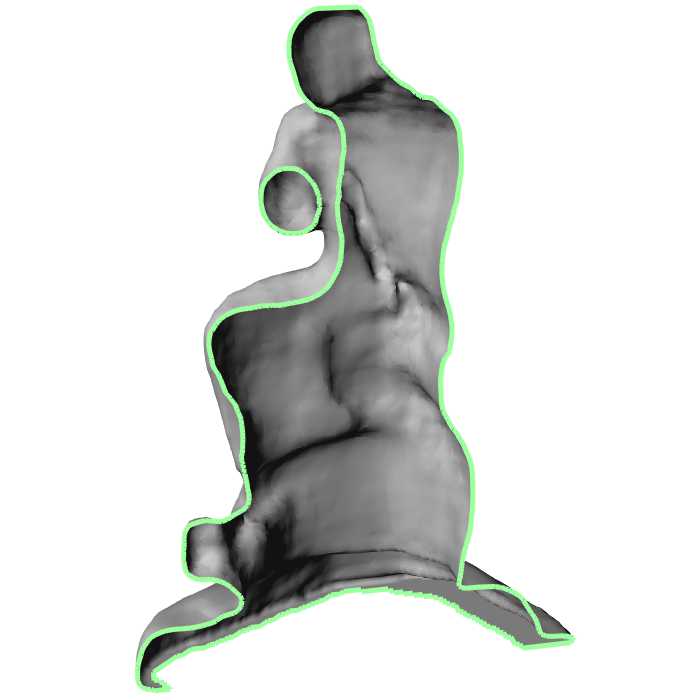}

\end{tabular}

\caption{{\bf Cut of Out-of-Domain Object Reconstruction.} A cut (along the green curve) of the reconstructed surface of \emph{Ignatius}. The reconstructions from the bare point cloud (top row) include empty space enclosed inside the object with backfaces, leading to a poor volumetric IoU. The reconstructions from the point cloud augmented with visibility information (bottom row) include only one surface, close to the input points.
}
\label{fig:cut}
\end{figure}

The increased generalization 
capability of the models is also validated when reconstructing surfaces from real-world scans obtained with LiDAR or MVS. In Figures \ref{fig:pocoscenescannet} and \ref{fig:real}, we show that networks using visibility information can reconstruct a more accurate and more complete surface. The reason for the largely improved volumetric IoU when using visibility information is illustrated in \figref{fig:cut}. For out-of-domain reconstructions, the baseline methods often predict hollow shapes, i.e., empty space enclosed inside an object. This leads to backfaces behind the real surface and a poor volumetric IoU. On the contrary, our models, trained on visibility-augmented point clouds, learn to distinguish between empty and full space more reliably and do not produce such artifacts.


\section{Limitations and Perspectives}

The position of auxiliary points depends on parameter~$d$, which is the average distance, across the whole scene, from a point to its nearest neighbor. To better handle point density variations, it could be set locally rather than globally.
Besides, as this positioning is also sensitive to sampling noise, $d$ could also be directly adjusted after noise estimation.


Our current approach only associates each point with a single sensor, while MVS points typically have several. A more efficient and versatile approach than simply duplicating sightlines is still an open issue.


Last, we resort to virtual scans because current 3D reconstruction benchmarks do not provide sensor positions. While we show that using our augmented point clouds allows common architectures to successfully generalize from virtual to real scenes, our training set may fail to replicate some challenging configurations encountered using actual sensors.

\section{Conclusion}
The sensor poses are often ignored in point cloud processing, even though available with most acquisition technologies. We present two straightforward ways to exploit sensor positions to augment point clouds with visibility information. 
Our experiments show that various deep surface reconstruction methods can be adapted with minimal effort to exploit these visibility-augmented point clouds, resulting in improved accuracy and completeness of reconstructed surfaces, as well as a substantial increase in generalization capability. 



\smallskip
\paragraph*{Acknowledgments} 
This work was partially funded by the ANR-17-CE23-0003 BIOM grant.
\FloatBarrier

\newpage

\bibliographystyle{IEEEtran} 
\bibliography{bibliography}
\balance

\ARXIV{\newpage \section*{\Large Supplementary Material} \global\csname @topnum\endcsname 0
\global\csname @botnum\endcsname 0

In this supplementary document, we first provide additional information about the datasets that we use (\secref{sec:datasets}), 
formal definitions of the evaluation metrics (\secref{sec:metrics}) and additional quantitative and qualitative results (\secref{sec:add_results}). Our code, data and pretrained models can be found online: \url{https://github.com/raphaelsulzer/dsrv-data} .

\section{Datasets}
\label{sec:datasets}

\subsection{Scanning procedure}
In \figref{fig:scan}, we represent a visualization of our scanning procedure. Viewpoints and ray target points 
are uniformly sampled on the surface of the spheres.

\begin{figure}[ht!]
    \centering
    \includegraphics[width=.6\linewidth]{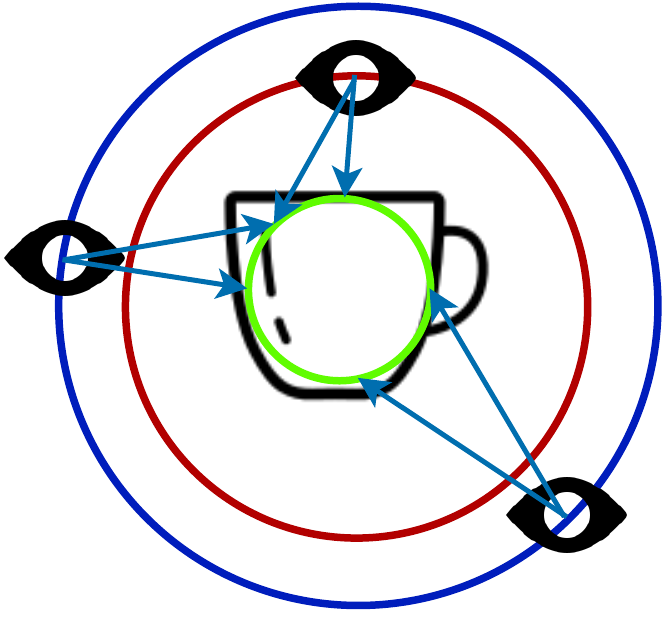}
    \vspace*{-2mm}
    \caption{\textbf{Scanning Procedure.} We randomly place sensors on two spheres (\textcolor{red}{red} and \textcolor{blue}{blue}) around the object, and consider rays aiming at uniformly sampled points on a sphere inscribed in the convex hull of the object (\textcolor{green}{green}).}
    \label{fig:scan}
\end{figure}

\subsection{ModelNet}

We use the official ModelNet10 dataset and make the models  watertight using ManifoldPlus \cite{huang2020manifoldplus}. We scan the watertight models using the procedure described above.

\subsection{ShapeNet}
We use the watertight models provided\footnote{\url{https://s3.eu-central-1.amazonaws.com/avg-projects/occupancy_networks/data/watertight.zip}} by the authors of Occupancy Networks \cite{Mescheder2019} and scan the models using the procedure described above. We apply a transformation to the models (and scans) to match their orientation to the orientation of the ModelNet10 objects (except for networks marked with $\dagger$ in Table~IV, which were trained with the original orientation).

\subsection{Synthetic Rooms Dataset}
We use the watertight scenes provided\footnote{\url{https://s3.eu-central-1.amazonaws.com/avg-projects/convolutional_occupancy_networks/data/room_watertight_mesh.zip}} by the authors of ConvONet \cite{Peng2020}. We scan the scenes using the procedure described above, limiting the sensors to lie in the upper hemisphere only.

\subsection{Tanks And Temples, Middlebury and DTU}
For the reconstructions in Figure~5, 6 and \figref{fig:real_suppmat}, we downsample the input point clouds to 10,000 points.

\begin{figure*}[t!]
	\centering
	\newcommand{\mywidth}{0.14\textwidth}
	\small
\begin{tabular}{cc|cccc|c}
\rotatebox{90}{\hspace{12mm}Bare}&
\includegraphics[width=\mywidth]{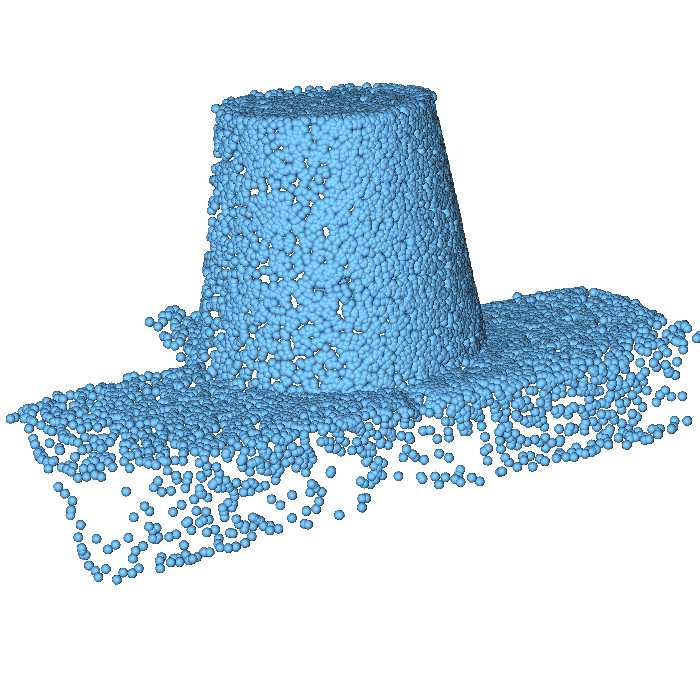}&
\includegraphics[width=\mywidth]{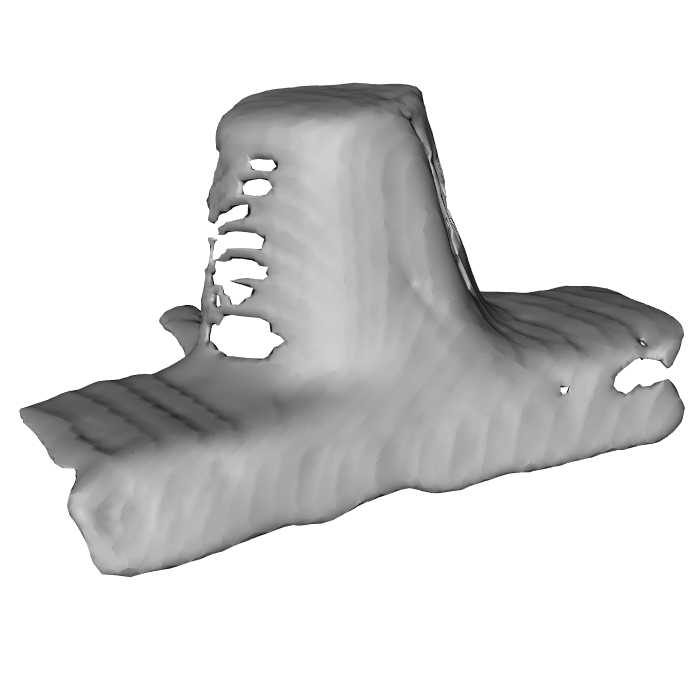}&
\includegraphics[width=\mywidth]{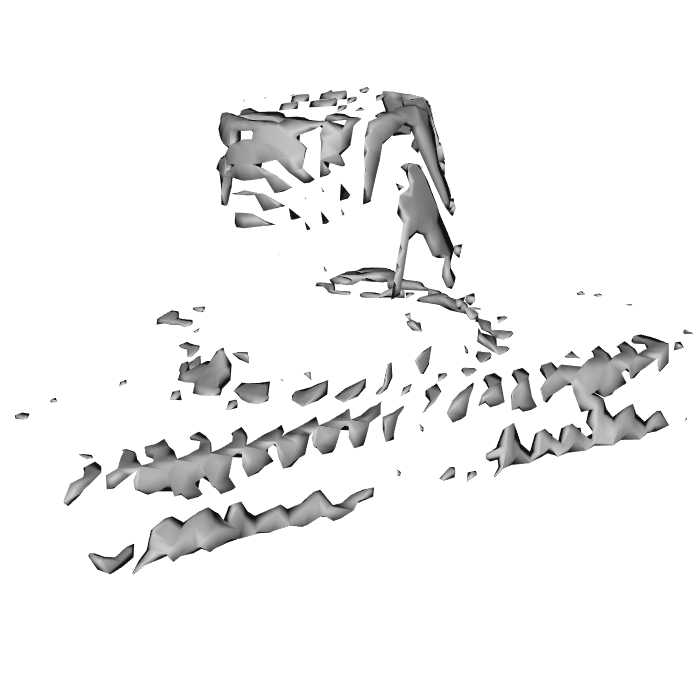}&
\includegraphics[width=\mywidth]{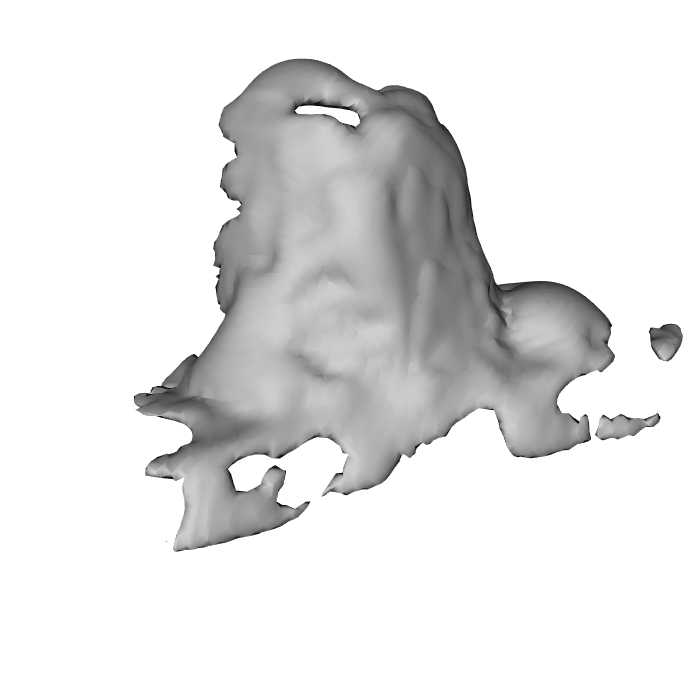}&
\includegraphics[width=\mywidth]{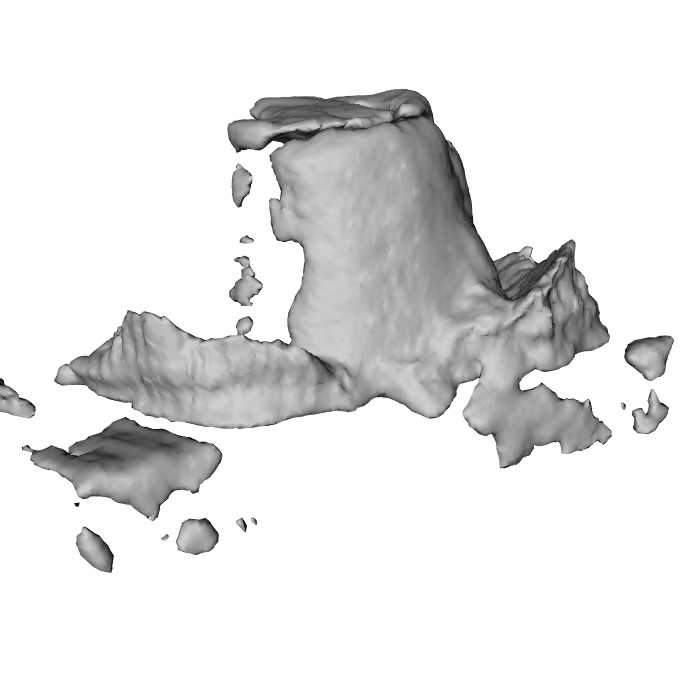}&
\multirow{2}{*}[3em]{\includegraphics[width=\mywidth]{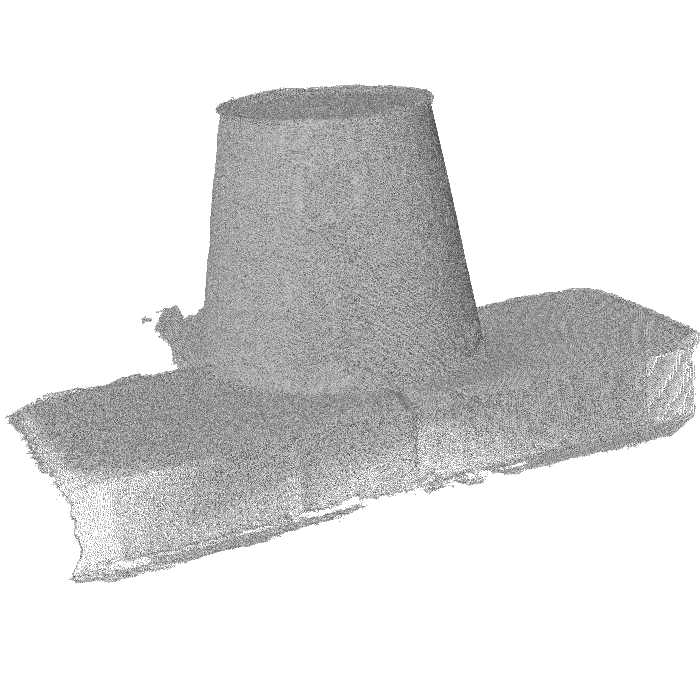}} 
\\
\rotatebox{90}{\hspace{7mm}Augmented}&
\includegraphics[width=\mywidth]{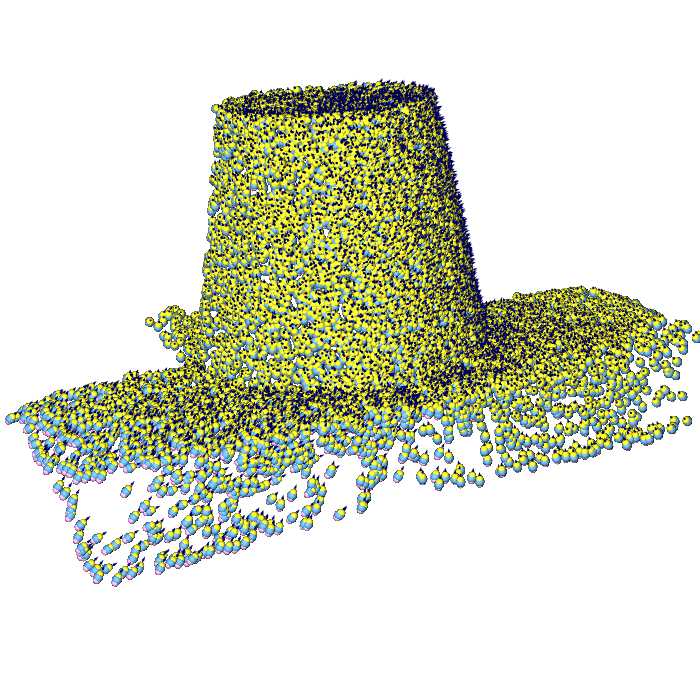}&
\includegraphics[width=\mywidth]{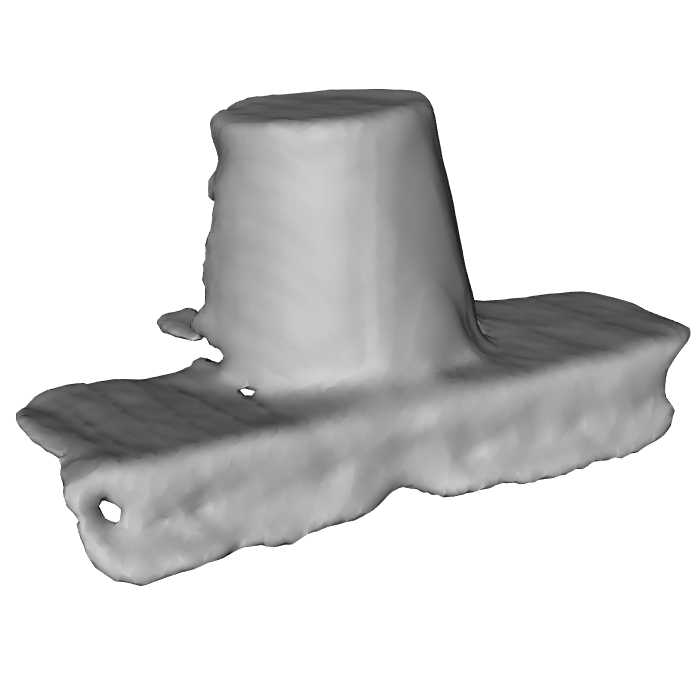}&
\includegraphics[width=\mywidth]{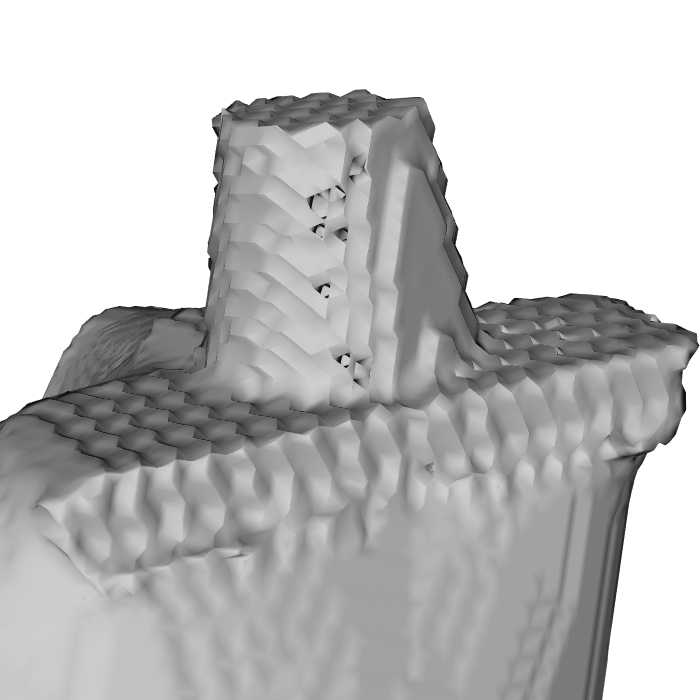}&
\includegraphics[width=\mywidth]{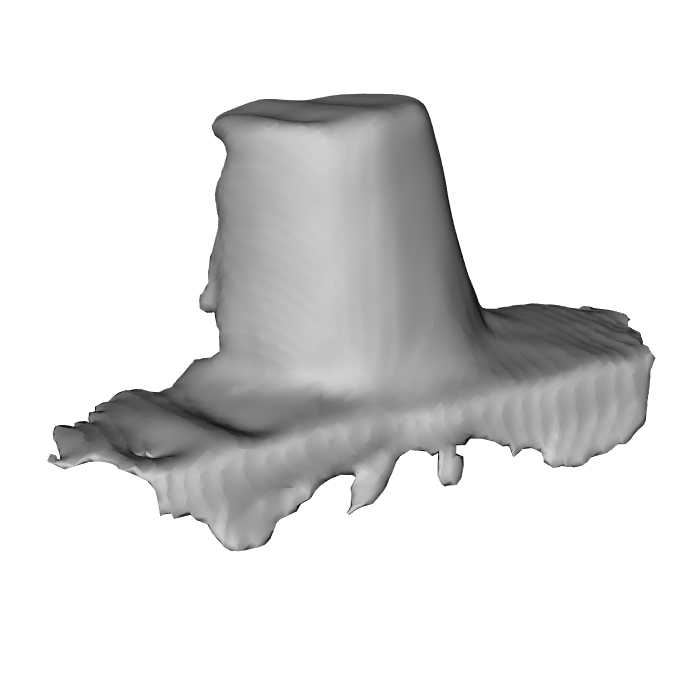}&
\includegraphics[width=\mywidth]{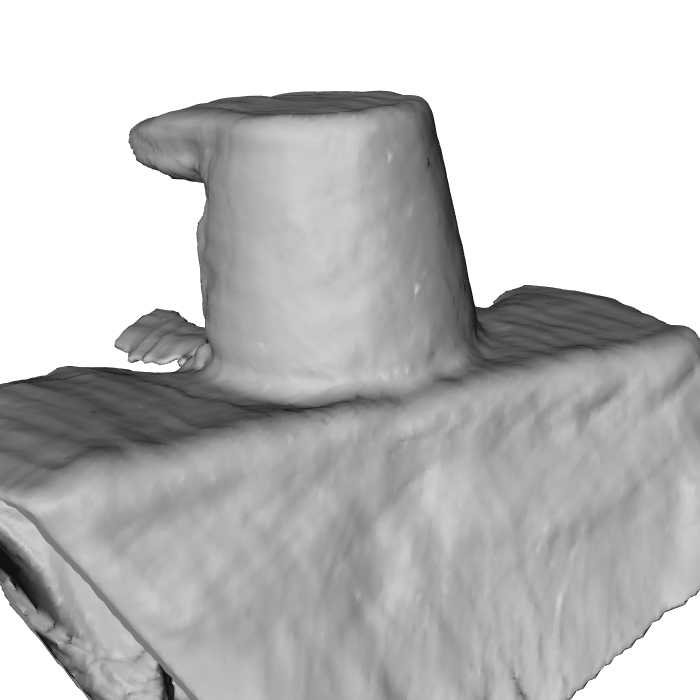}&
\\
&
Input &
ConvONet-3D~\cite{Peng2020} &
Points2Surf~\cite{points2surf} &
Shape\,As\,Points~\cite{Peng2021SAP} &
POCO~\cite{boulch2022poco}&
HD Scan
\end{tabular}
	\caption{
		\textbf{Out-of-Domain Object-Level Reconstruction.}
		Reconstructed shape from a MVS point cloud of \emph{scan1} from DTU, using four different DSR methods trained on ModelNet10.
		The top row uses the bare point cloud as input, and the bottom row uses the point cloud augmented with visibility information.
		\emph{HD Scan} is a high-density point cloud.
	}
	\label{fig:real_suppmat}
\end{figure*}

\section{Metrics}
\label{sec:metrics}
We evaluate the quality of reconstructions 
with the volumetric IoU (IoU), symmetric Chamfer distance (CD) and normal consistency (NC). 

Let $\mathcal{M_G}$ be the ground truth mesh and $\mathcal{M_P}$ be the reconstructed mesh.
The volumetric IoU is defined as:
\begin{align*}
    \text{IoU}\left(\mathcal{M_G},\mathcal{M_P}\right) =
    &\frac{\vol(\mathcal{M_G} \cap \mathcal{M_P})}{\vol(\mathcal{M_G} \cup \mathcal{M_P})},
\end{align*}
We approximate volumetric IoU by sampling $100,000$~points in the union of the bounding boxes of $\mathcal{M_G}$ and $\mathcal{M_P}$.

To compute the Chamfer distance and normal consistency, we sample a set of points $S_G$ on the ground-truth mesh and a set of points $S_P$ on the reconstructed mesh with $\vert S_G \vert = \vert S_P \vert = 100,000$.
We approximate the  two-sided Chamfer distance between $\mathcal{M_G}$ and $\mathcal{M_P}$ as follows:

\begin{align*}\nonumber
    \text{CD}(\mathcal{M_G},\mathcal{M_P}) =
    &\frac{1}{2\vert S_{{G}} \vert} \sum_{x \in S_G} \min_{y \in S_P} {\vert \vert x - y \vert \vert }_2 \\
    + &\frac{1}{2\vert S_{{P}} \vert}\sum_{y \in S_P} \min_{x \in S_G} {\vert \vert y - x \vert \vert }_2
\end{align*}

Let $n(x)$ be the unit normal associated to a point $x$ taken on a mesh, and $\langle\cdot{,}\cdot\rangle$ the Euclidean scalar product in $\bR^3$.
Normal consistency is defined as:
\begin{align*}\nonumber
    \text{NC}(\mathcal{M_G},\mathcal{M_P}) =
    &\frac{1}{2\vert S_{{G}} \vert} \sum_{x \in S_G} \left\langle n(x),n\left(\argmin_{y \in S_P} {\vert \vert x - y \vert \vert }_2\right) \right\rangle  \\
    + &\frac{1}{2\vert S_{{P}} \vert}\sum_{y \in S_P} \left\langle n(y),n\left(\argmin_{x \in S_G} {\vert \vert y - x \vert \vert }_2\right) \right \rangle
\end{align*}

\section{Additional Results}
\label{sec:add_results}

\subsection{Runtimes}

In \tabref{tab:timing}, we report detailed runtimes for the tested methods, with and without visibility information. 

Adding sightline vectors does not significantly increase the runtime for any of the tested methods. The effect of auxiliary points depends on the method. For ConvONet, most of the processing time is spent computing grid features. The encoding of 3d points is performed by a small PointNet network \cite{qi2017pointnet}, whose runtime is only a small fraction of the total time. As a consequence, adding APs does not incur significant changes in computation time.
In contrast, Points2Surf uses a large point encoding network and is $2.2$ times slower with APs. Shape As Points is $1.3$ times slower due to the fact that we decode $3$ times as many points as the baseline method. POCO is also essentially unaffected by the addition of auxiliary points as they only impact the first (small) layer of the point-convolution backbone.
\begin{table}
\caption{
	\textbf{Runtimes for Object-Level Reconstruction.}
}
\vspace*{-3mm}
 	Average times (in seconds) for reconstructing one object from a point cloud of 3\,000 points with and without sightline vectors (SV) or auxiliary points (AP). MC is marching cubes. Times are averaged over the ModelNet10 test set.
\vspace*{2mm}

\centering
\resizebox{\columnwidth}{!}{
\begin{tabular}{@{}lccccccc@{}}
\toprule
Model                               & SV            & AP            & Encoding  & Decoding & MC        & Total         \\ \midrule
ConvONet-2D~\cite{Peng2020}         &               &               & 0.016     & 0.25     & 0.17     & 0.44         \\
ConvONet-2D~\cite{Peng2020}         & \checkmark    &               & 0.016     & 0.27     & 0.17     & 0.47         \\
ConvONet-2D~\cite{Peng2020}         & \checkmark    & \checkmark    & 0.016     & 0.26     & 0.17     & 0.45         \\ \midrule
Points2Surf~\cite{points2surf}      &               &               &\multicolumn{2}{c}{69.06} &11.51  & 80.57        \\
Points2Surf~\cite{points2surf}      & \checkmark    &               &\multicolumn{2}{c}{71.92} &11.35  & 83.27        \\
Points2Surf~\cite{points2surf}      & \checkmark    & \checkmark    &\multicolumn{2}{c}{173.2}&11.41  & 184.7      \\ \midrule
Shape As Points~\cite{Peng2021SAP}  &               &               & 0.022     & 0.017     & 0.047     & 0.088         \\
Shape As Points~\cite{Peng2021SAP}  & \checkmark    &               & 0.023     & 0.017     & 0.046     & 0.086         \\
Shape As Points~\cite{Peng2021SAP}  & \checkmark    & \checkmark    & 0.024     & 0.041     & 0.047     & 0.114         \\ \midrule
POCO~\cite{boulch2022poco}          &               &               & 0.088     & 13.72     & 0.33      & 15.74         \\
POCO~\cite{boulch2022poco}          & \checkmark    &               & 0.091     & 13.68     & 0.33      & 15.66         \\ 
POCO~\cite{boulch2022poco}          & \checkmark    & \checkmark    & 0.093     & 13.70     & 0.33      & 15.67         \\ 
\bottomrule
\end{tabular}
}
\vspace*{-2mm}
\label{tab:timing}
\end{table}




\subsection{DTU Dataset}
In \figref{fig:real_suppmat}, we show the reconstruction of an MVS point cloud, generated with OpenMVS, of \emph{scan1} from the DTU dataset \cite{dtu}. The point cloud represents an open scene, while all methods were trained  on the closed ModelNet10 objects. The methods using our augmented point clouds with visibility information cope much better with this domain shift.

\subsection{ShapeNet}
\begin{figure*}[]
	\centering\small
	\newcommand{\mywidth}{0.13\textwidth}
\begin{tabular}{cc|cccc|c}

\rotatebox{90}{\hspace{10mm}Bare}&
\includegraphics[width=\mywidth]{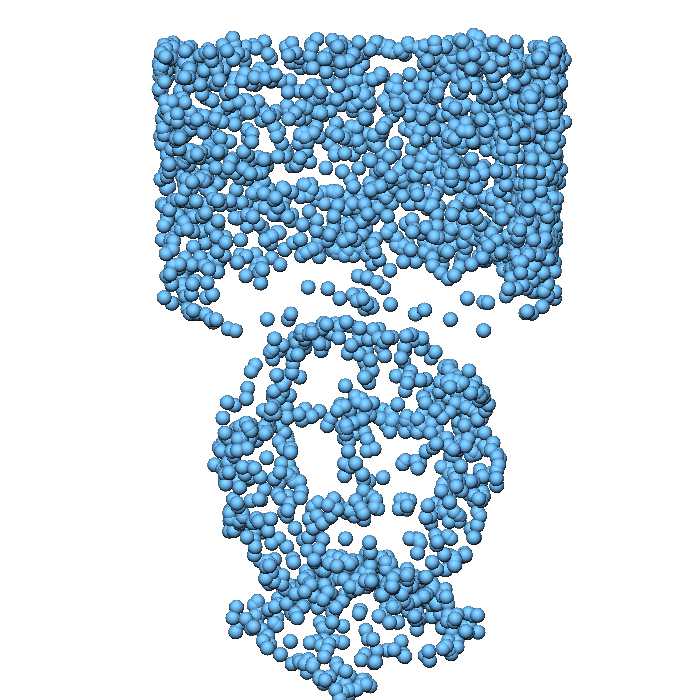}&
\includegraphics[width=\mywidth]{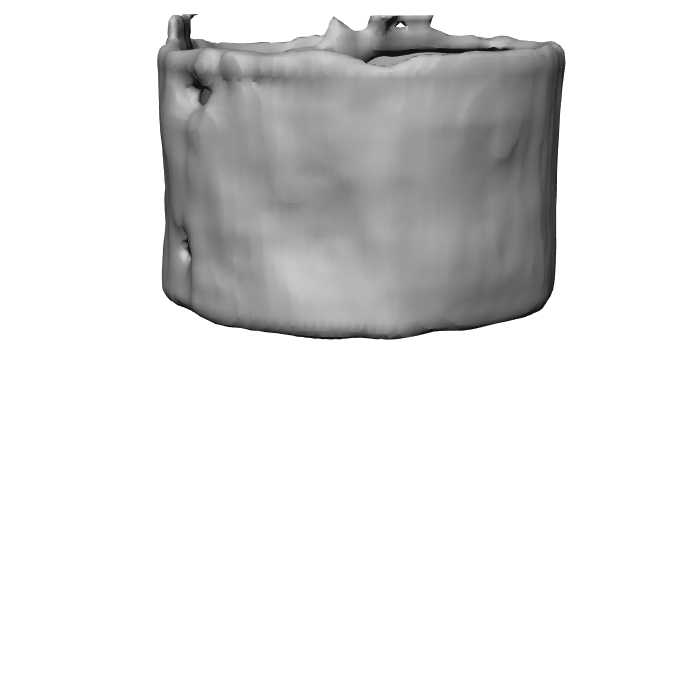}&
\includegraphics[width=\mywidth]{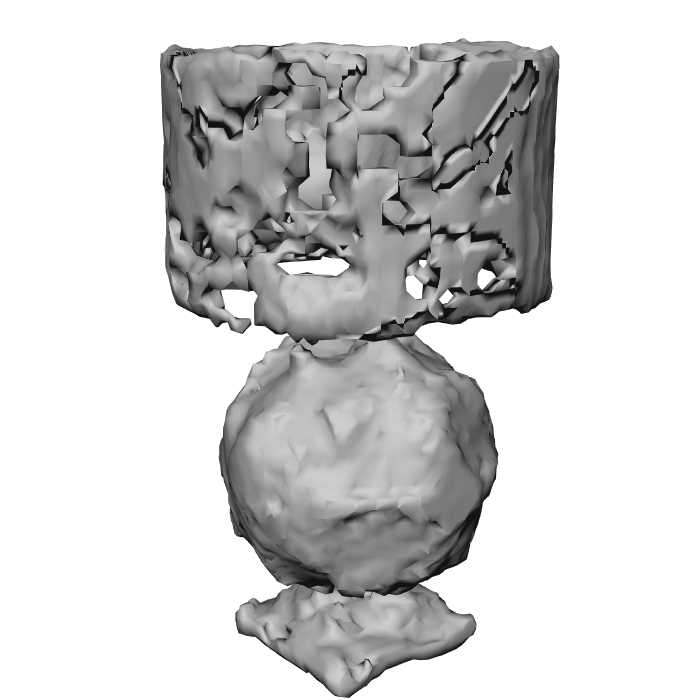}&
\includegraphics[width=\mywidth]{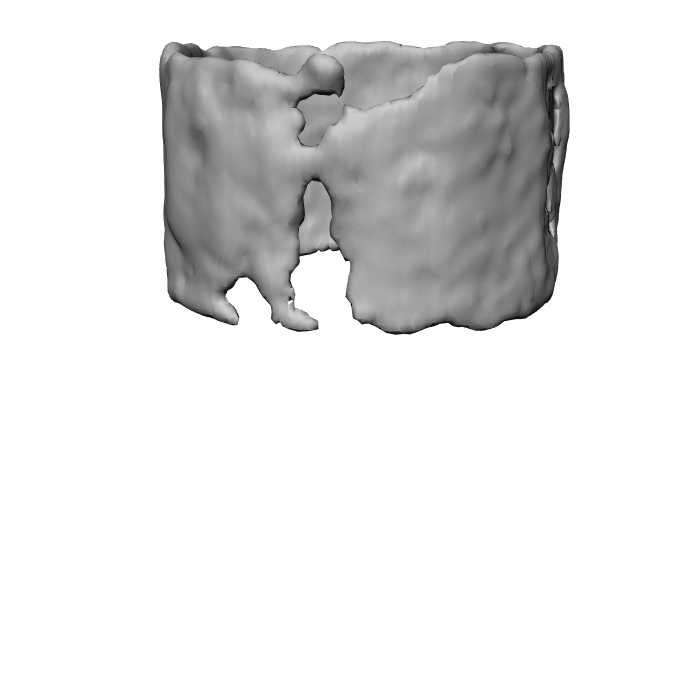}&
\includegraphics[width=\mywidth]{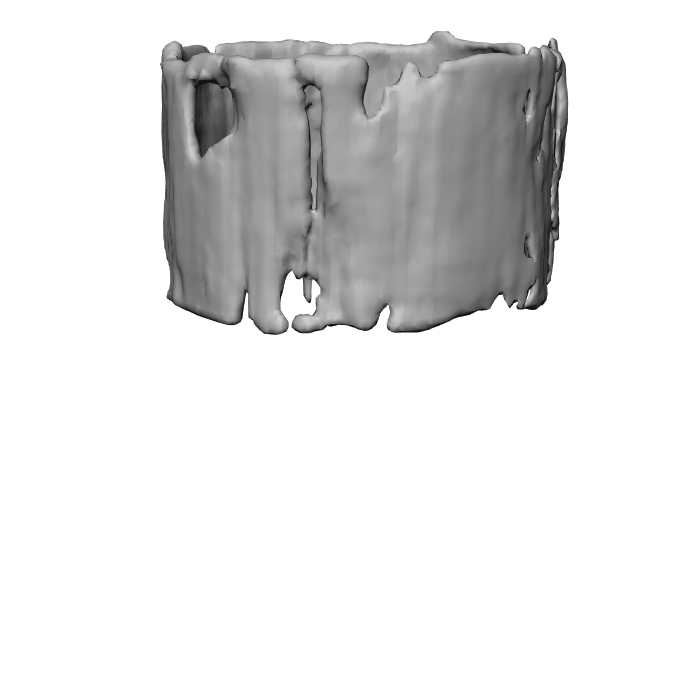}&
\multirow{2}{*}[3em]{\includegraphics[width=\mywidth]{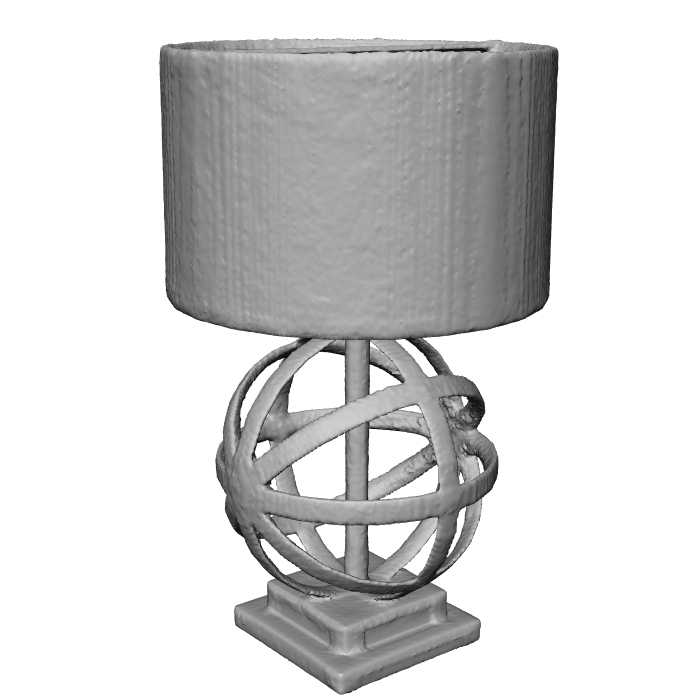}} \\
\rotatebox{90}{\hspace{5mm}Augmented}&
\includegraphics[width=\mywidth]{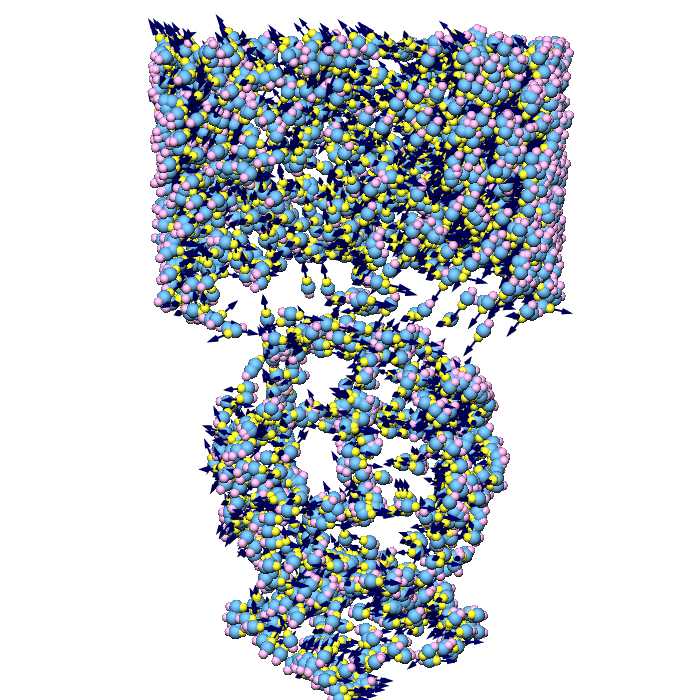}&
\includegraphics[width=\mywidth]{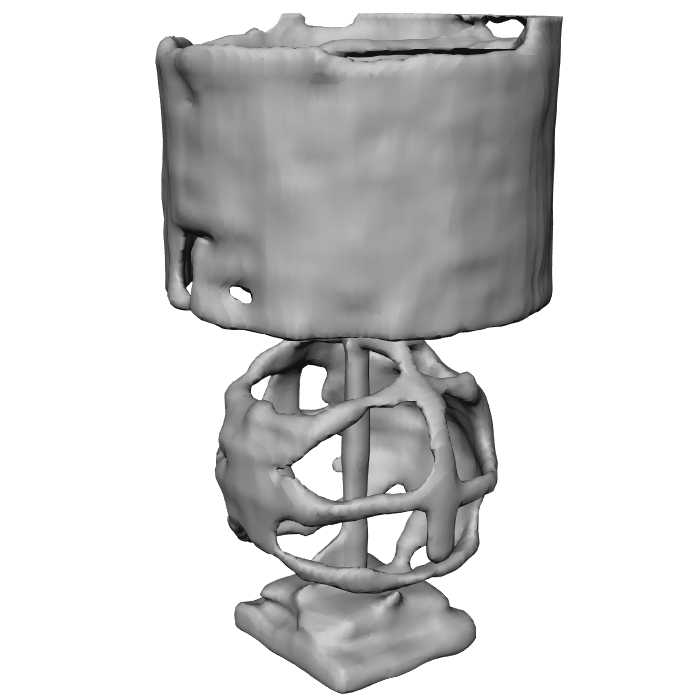}&
\includegraphics[width=\mywidth]{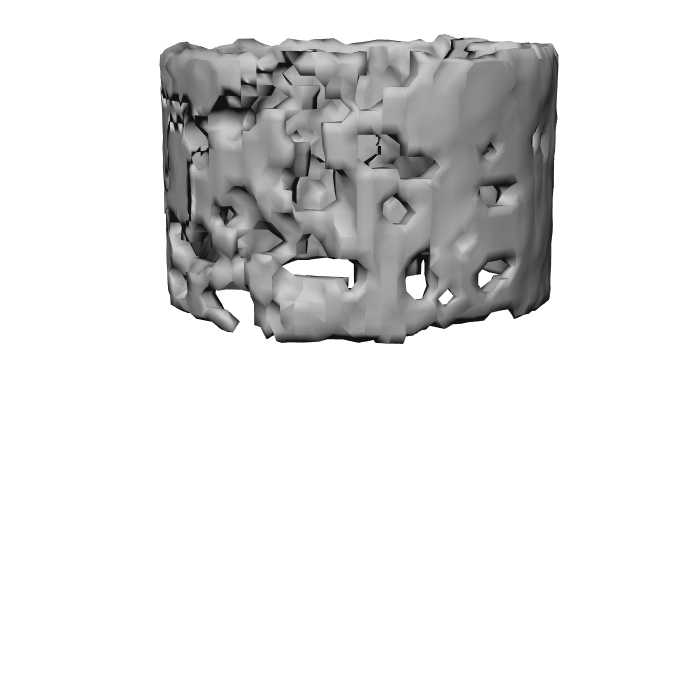}&
\includegraphics[width=\mywidth]{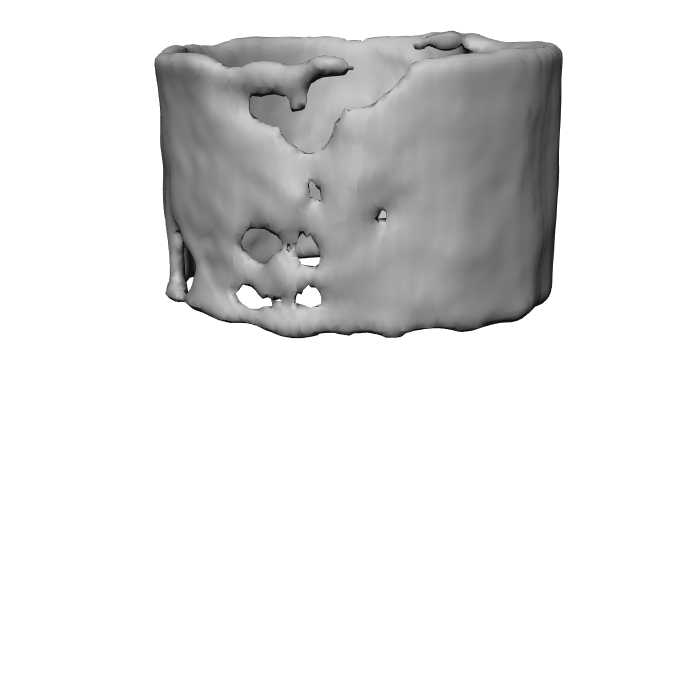}&
\includegraphics[width=\mywidth]{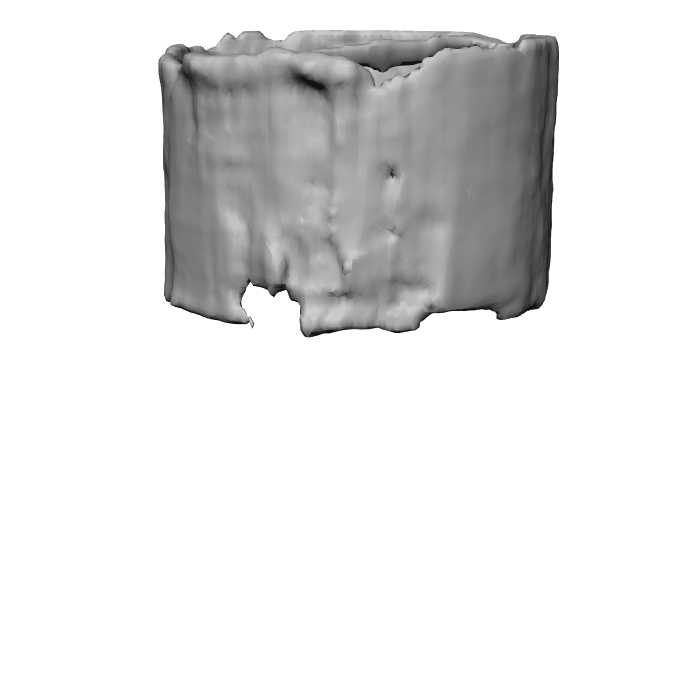}&
\\

\rotatebox{90}{\hspace{10mm}Bare}&
\includegraphics[width=\mywidth]{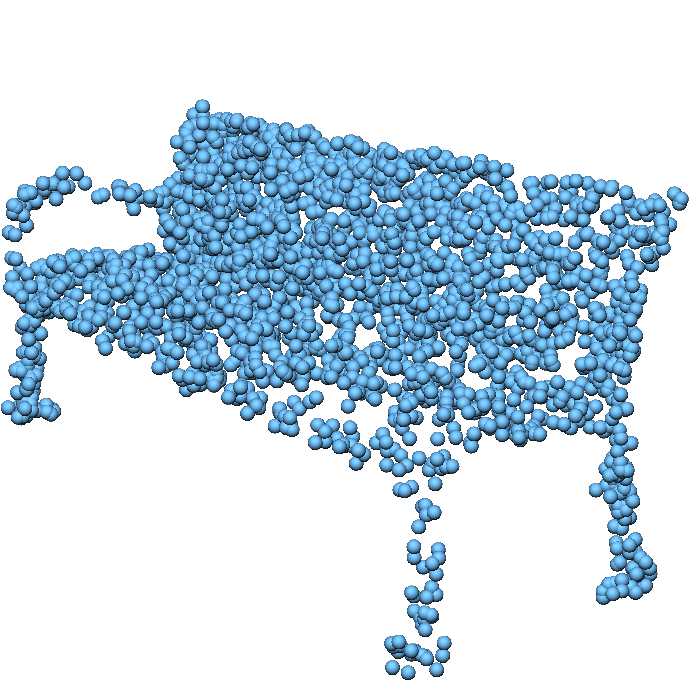}&
\includegraphics[width=\mywidth]{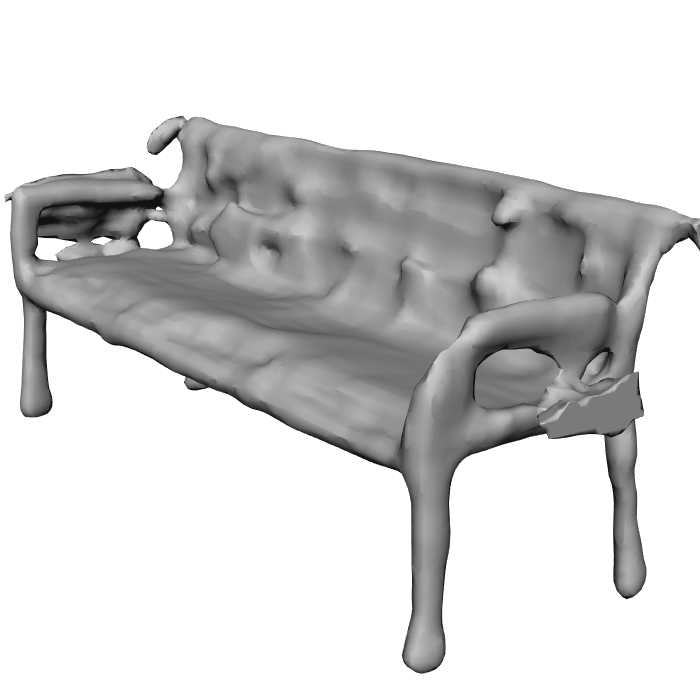}&
\includegraphics[width=\mywidth]{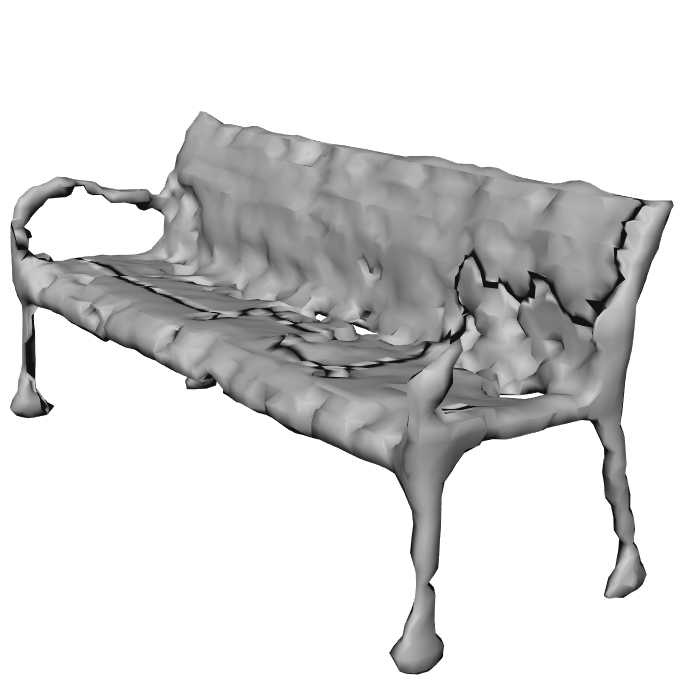}&
\includegraphics[width=\mywidth]{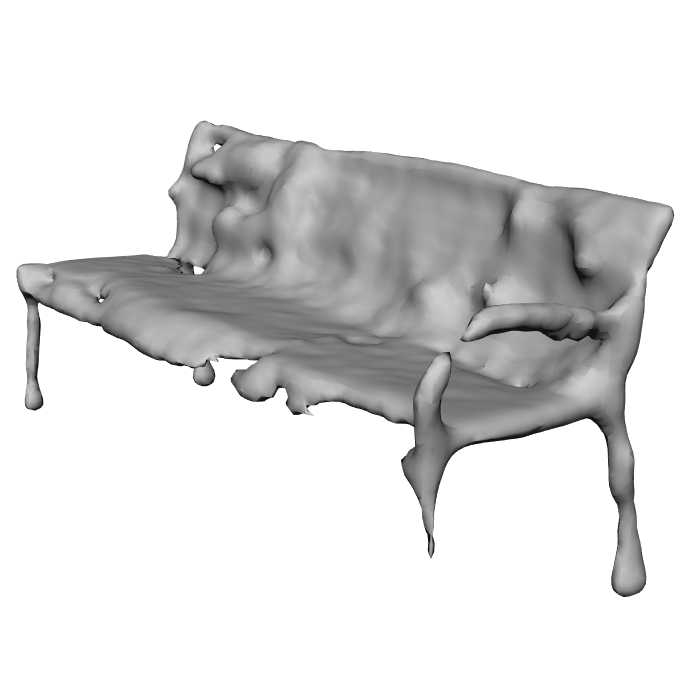}&
\includegraphics[width=\mywidth]{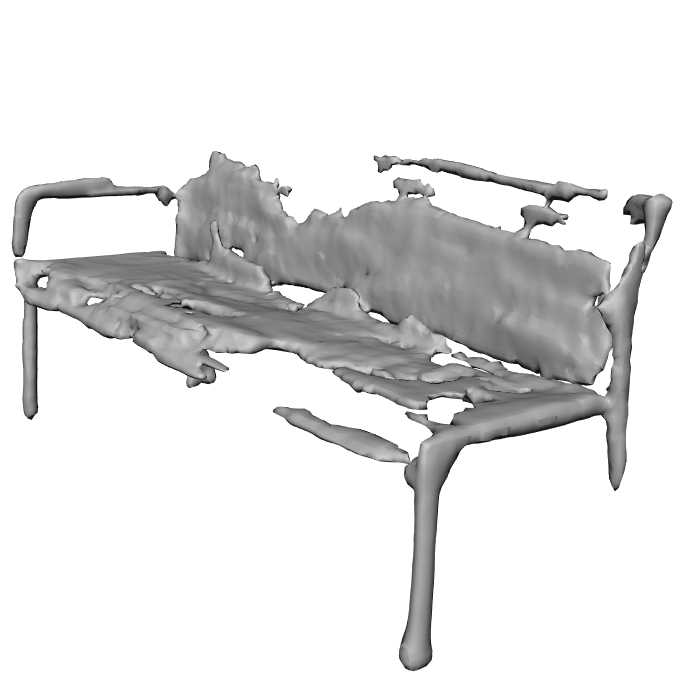}&
\multirow{2}{*}[3em]{\includegraphics[width=\mywidth]{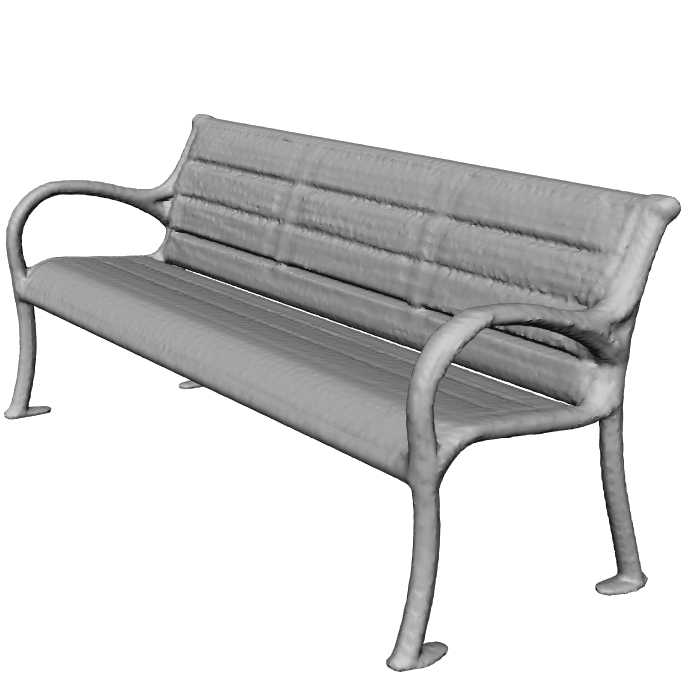}} \\
\rotatebox{90}{\hspace{5mm}Augmented}&
\includegraphics[width=\mywidth]{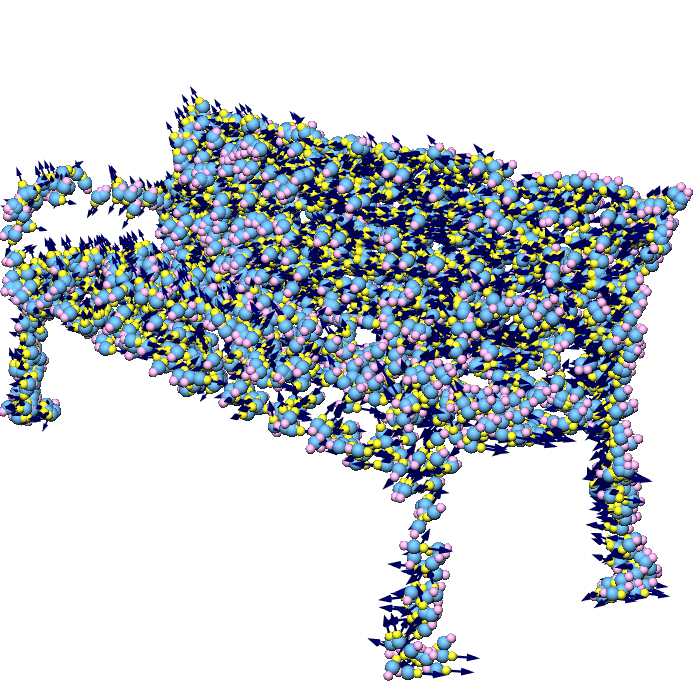}&
\includegraphics[width=\mywidth]{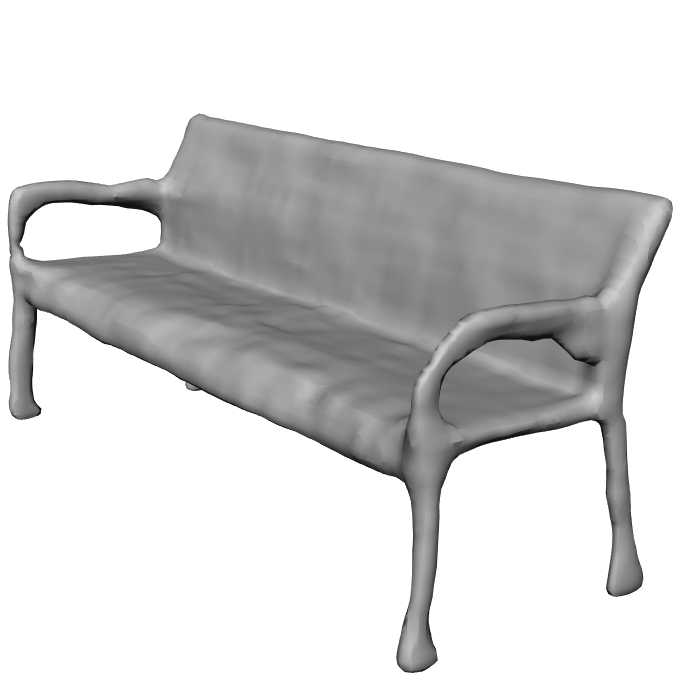}&
\includegraphics[width=\mywidth]{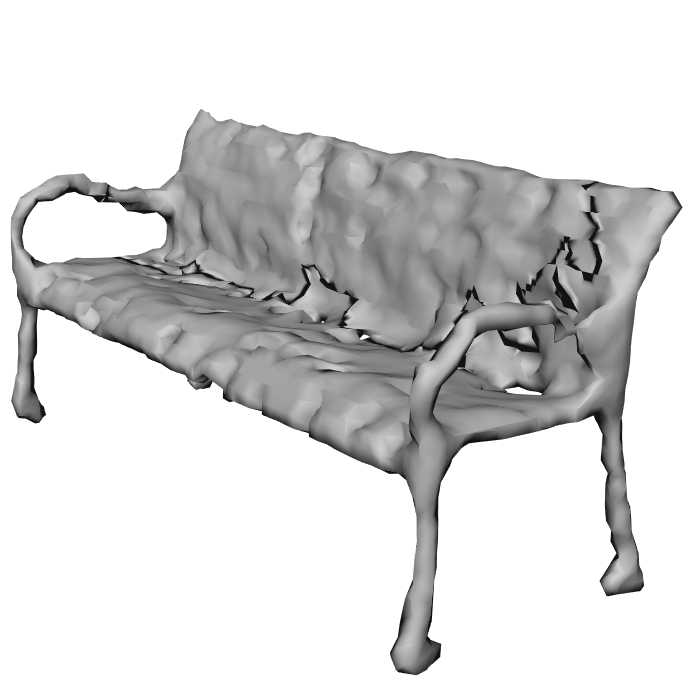}&
\includegraphics[width=\mywidth]{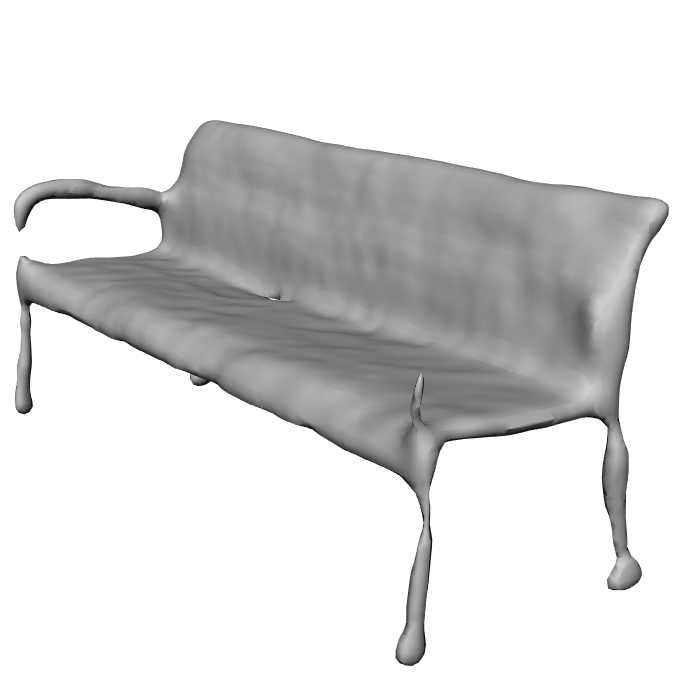}&
\includegraphics[width=\mywidth]{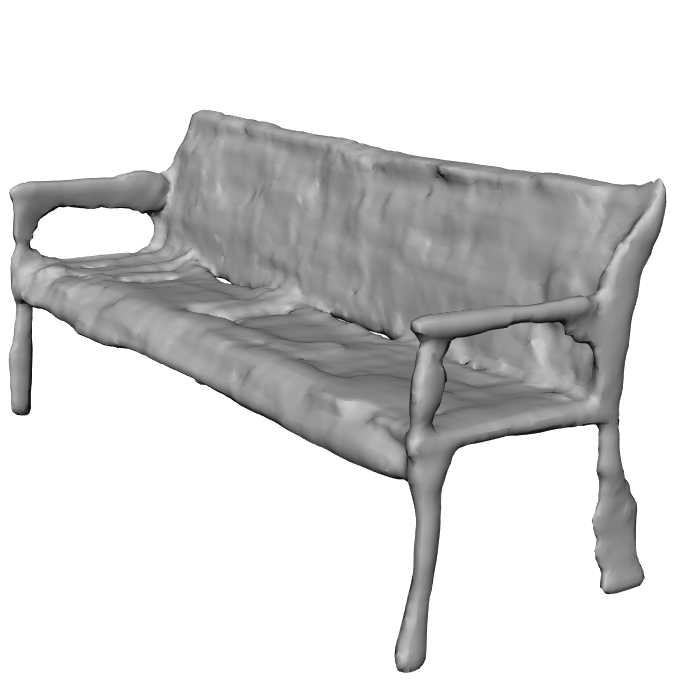}&
\\

\rotatebox{90}{\hspace{10mm}Bare}&
\includegraphics[width=\mywidth]{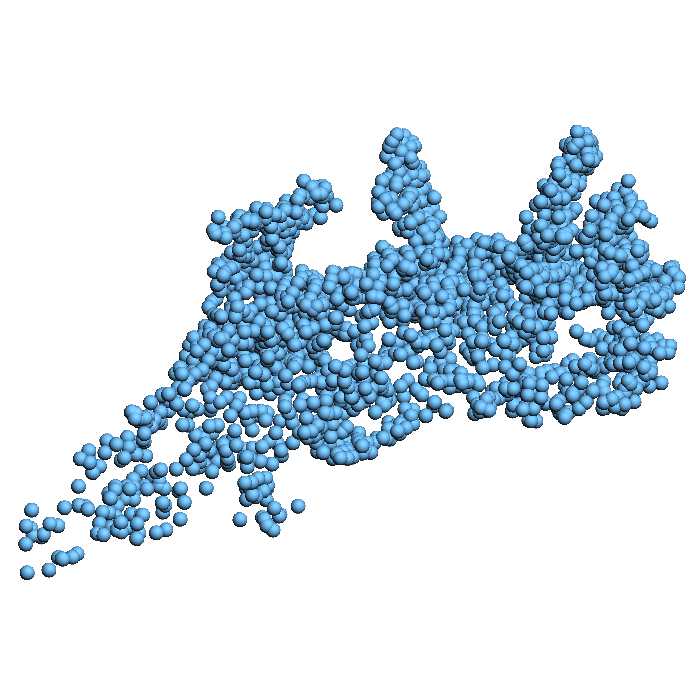}&
\includegraphics[width=\mywidth]{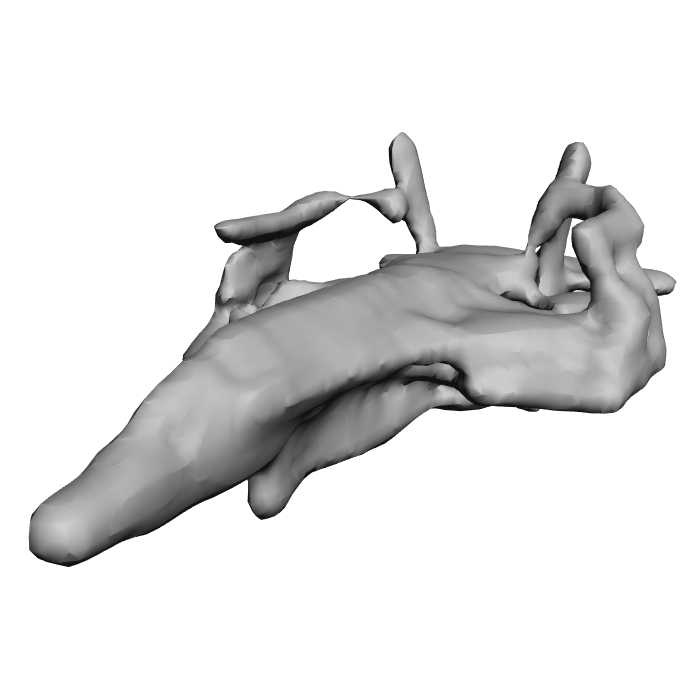}&
\includegraphics[width=\mywidth]{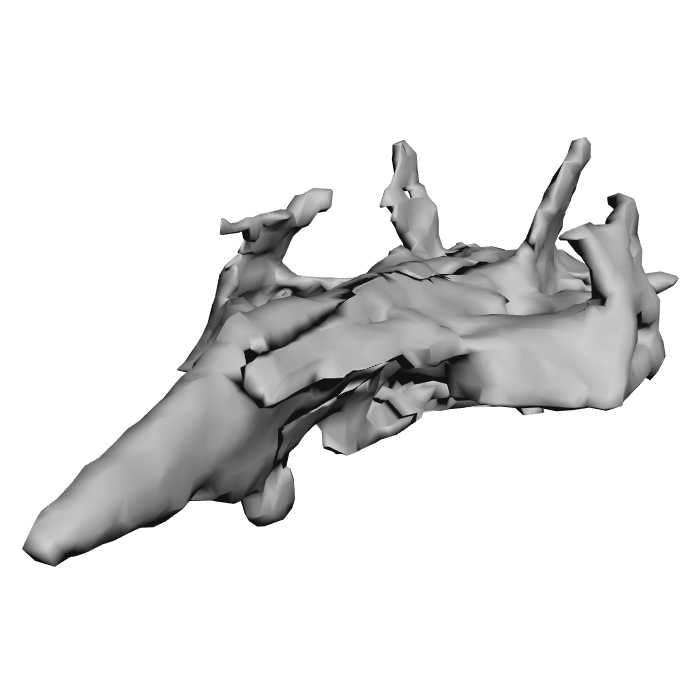}&
\includegraphics[width=\mywidth]{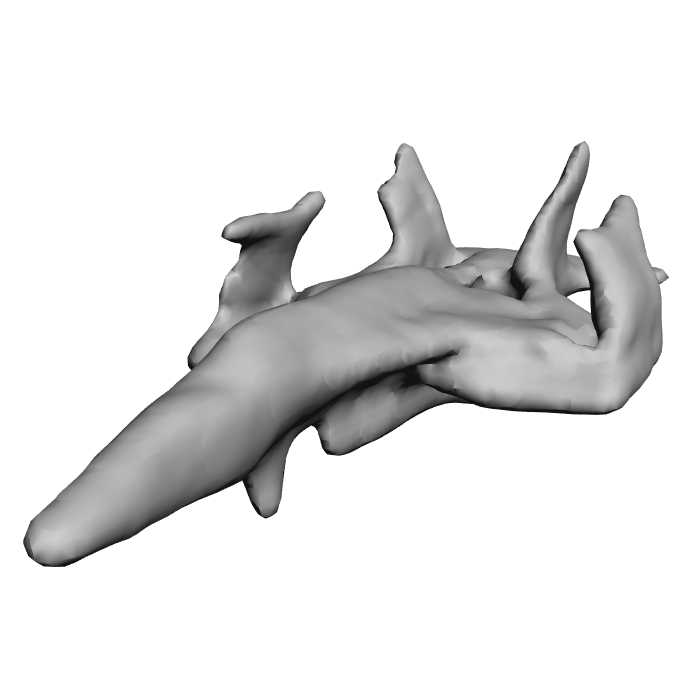}&
\includegraphics[width=\mywidth]{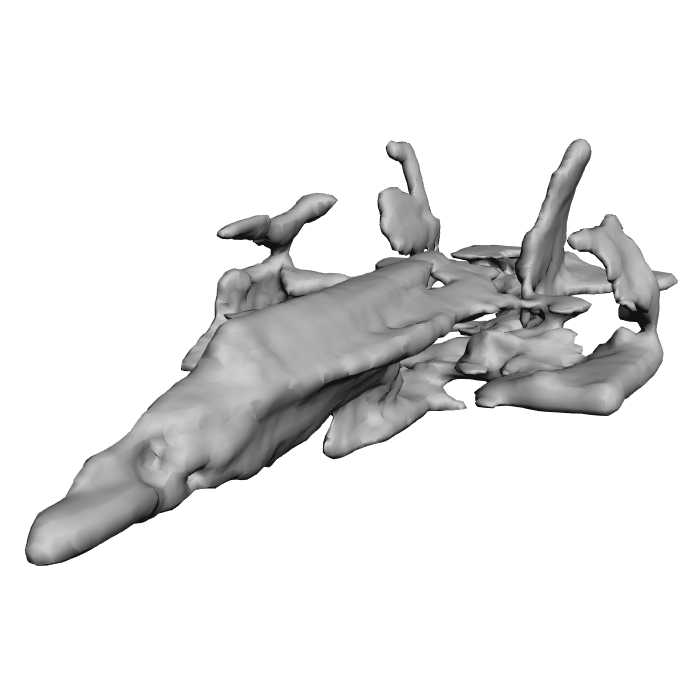}&
\multirow{2}{*}[3em]{\includegraphics[width=\mywidth]{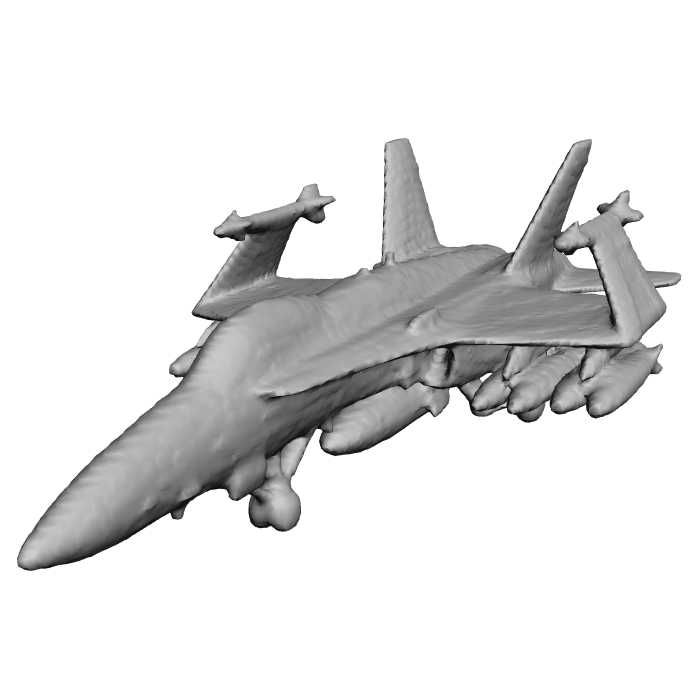}} \\
\rotatebox{90}{\hspace{5mm}Augmented}&
\includegraphics[width=\mywidth]{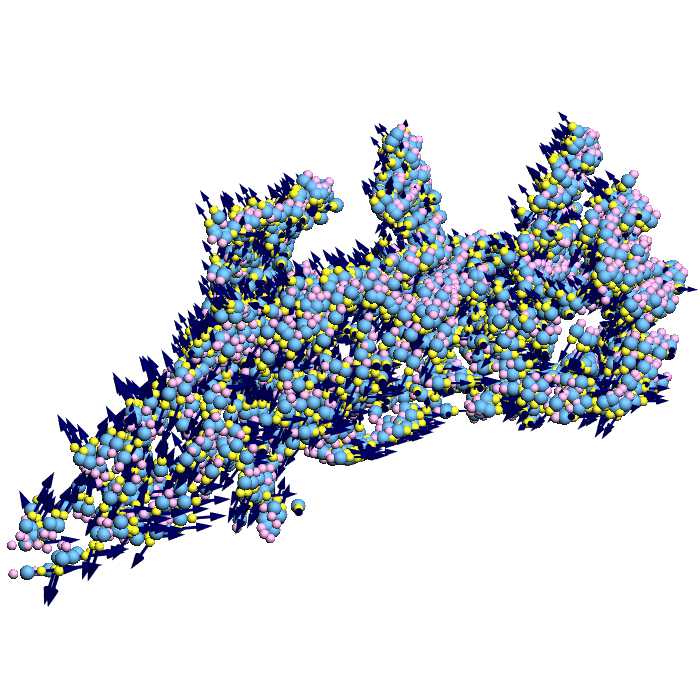}&
\includegraphics[width=\mywidth]{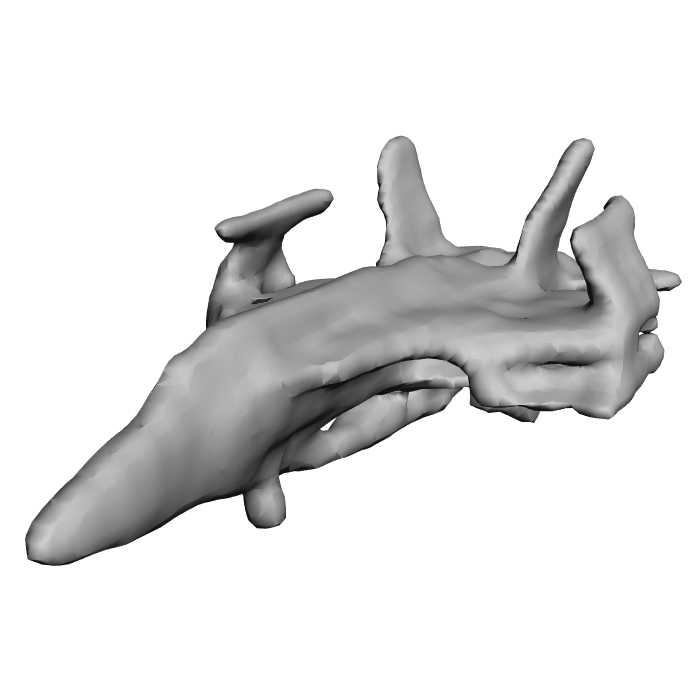}&
\includegraphics[width=\mywidth]{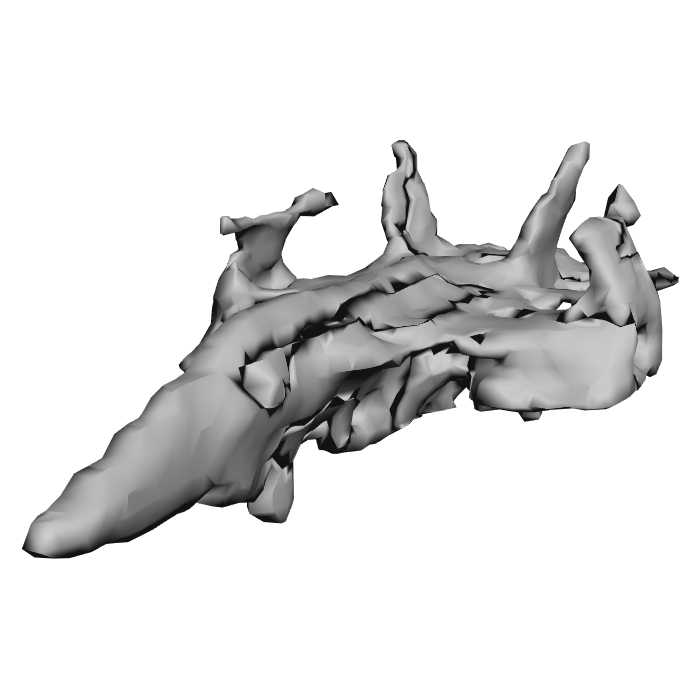}&
\includegraphics[width=\mywidth]{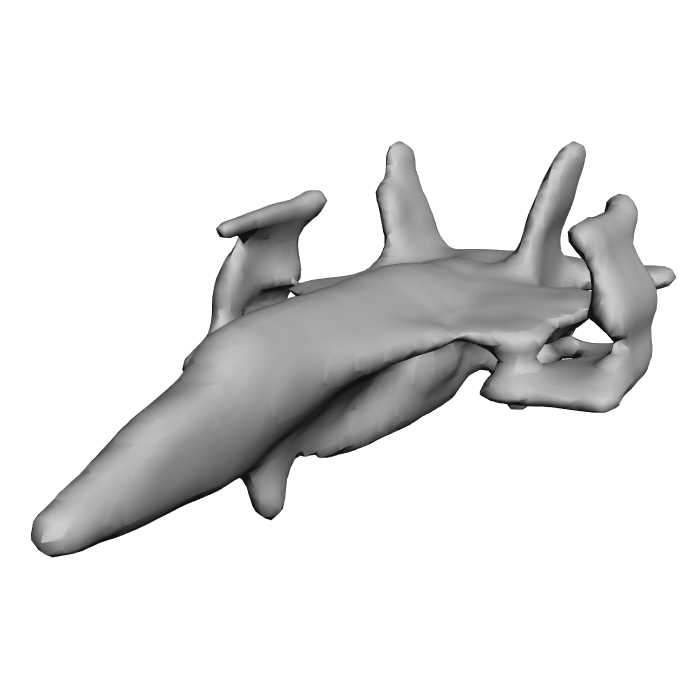}&
\includegraphics[width=\mywidth]{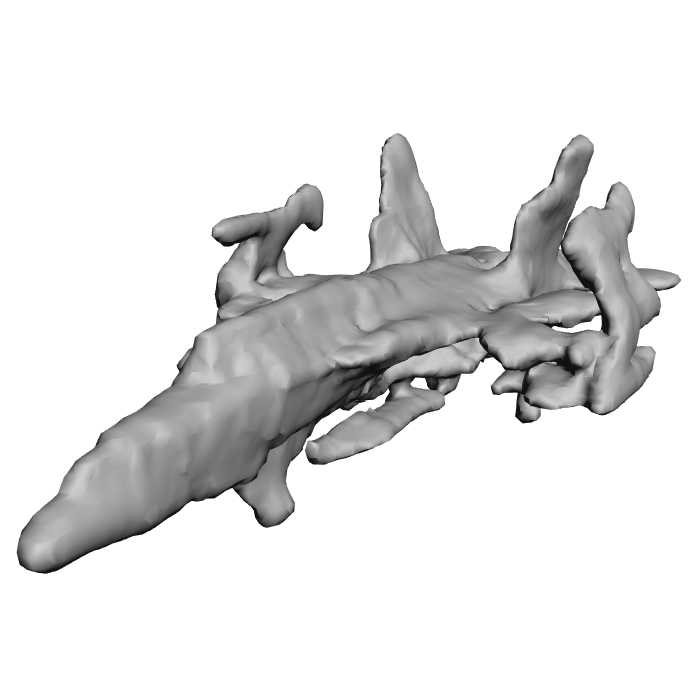}&
\\

\rotatebox{90}{\hspace{10mm}Bare}&
\includegraphics[width=\mywidth]{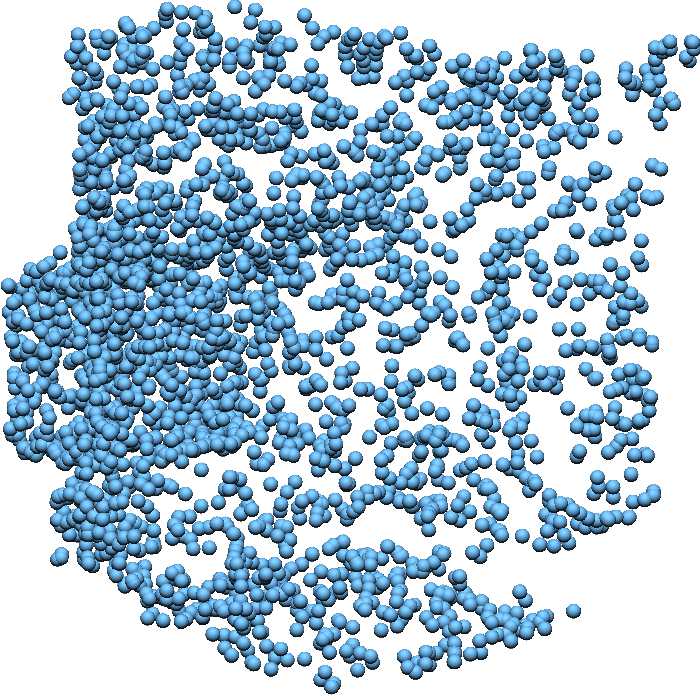}&
\includegraphics[width=\mywidth]{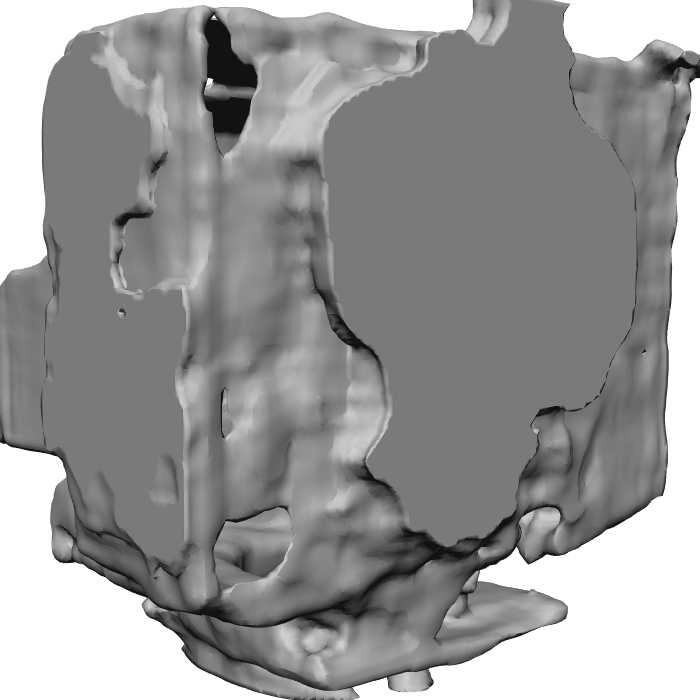}&
\includegraphics[width=\mywidth]{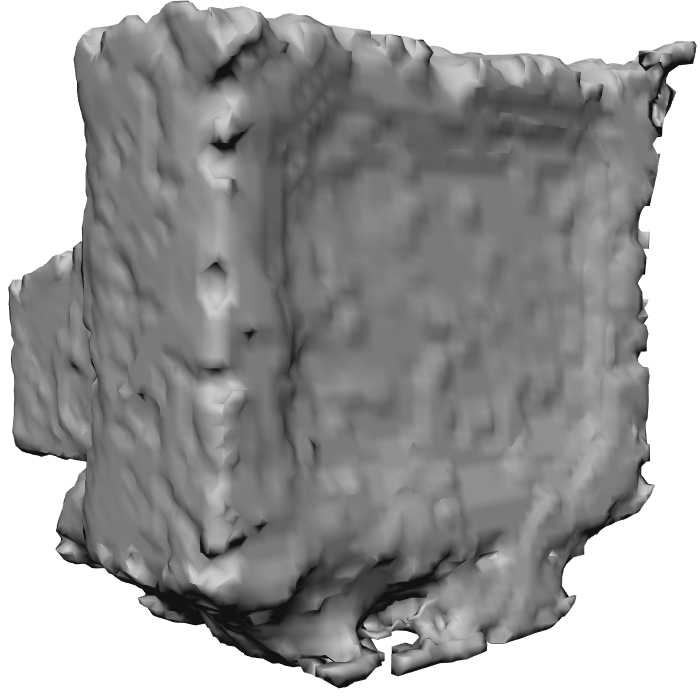}&
\includegraphics[width=\mywidth]{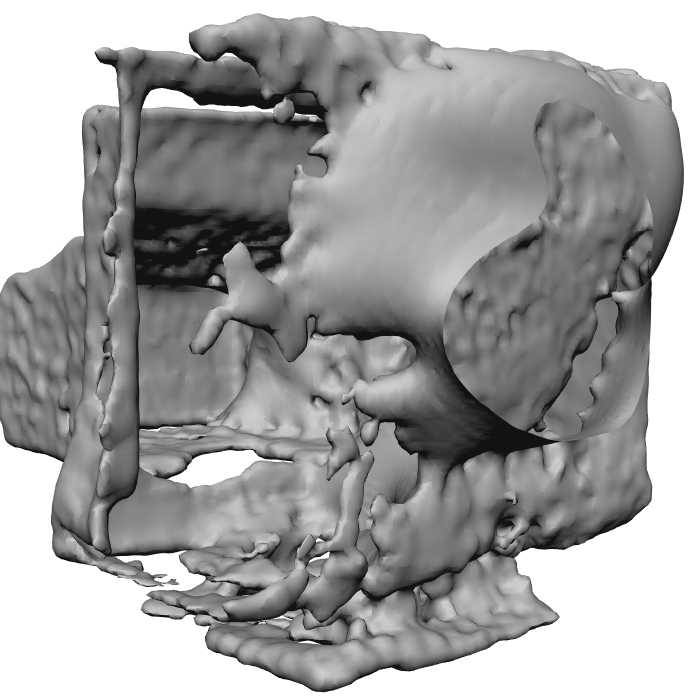}&
\includegraphics[width=\mywidth]{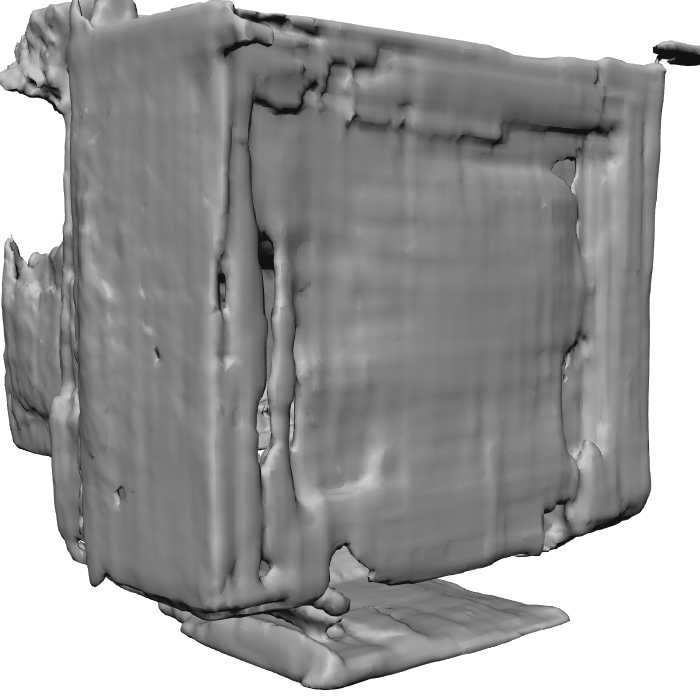}&
\multirow{2}{*}[3em]{\includegraphics[width=\mywidth]{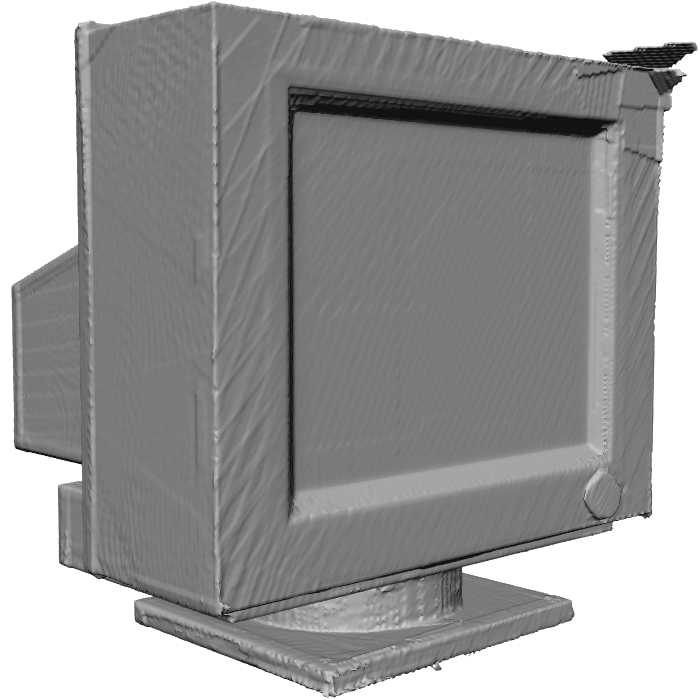}} \\
\rotatebox{90}{\hspace{4mm}Augmented}&
\includegraphics[width=\mywidth]{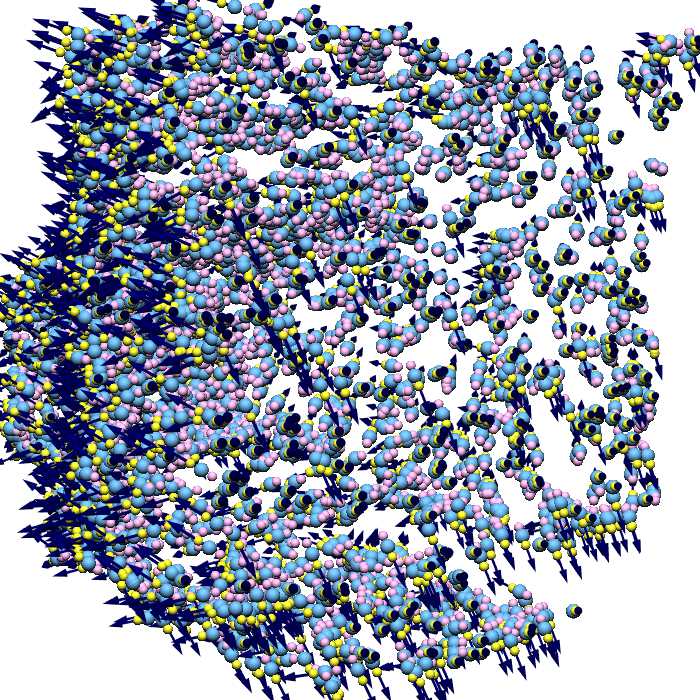}&
\includegraphics[width=\mywidth]{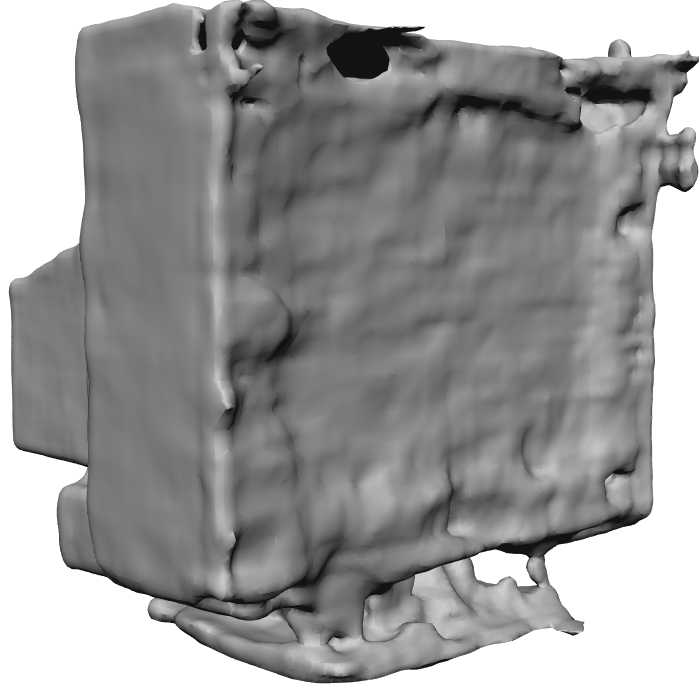}&
\includegraphics[width=\mywidth]{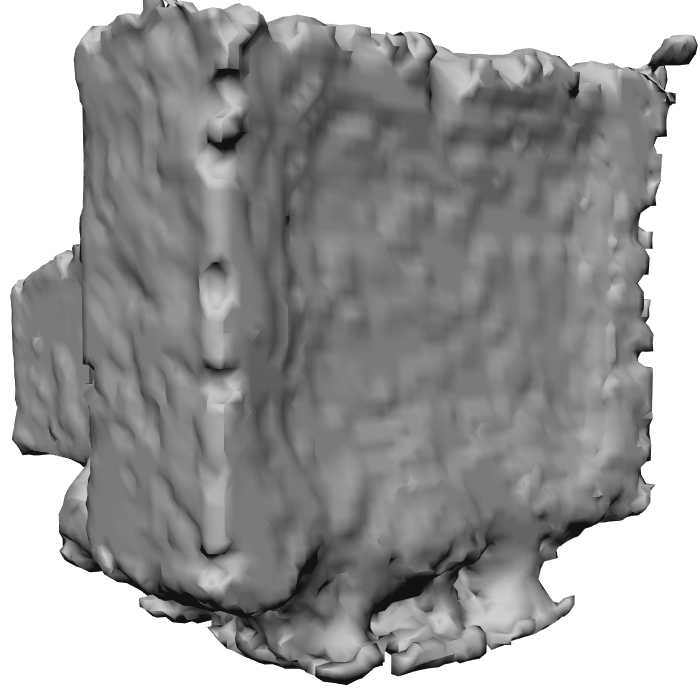}&
\includegraphics[width=\mywidth]{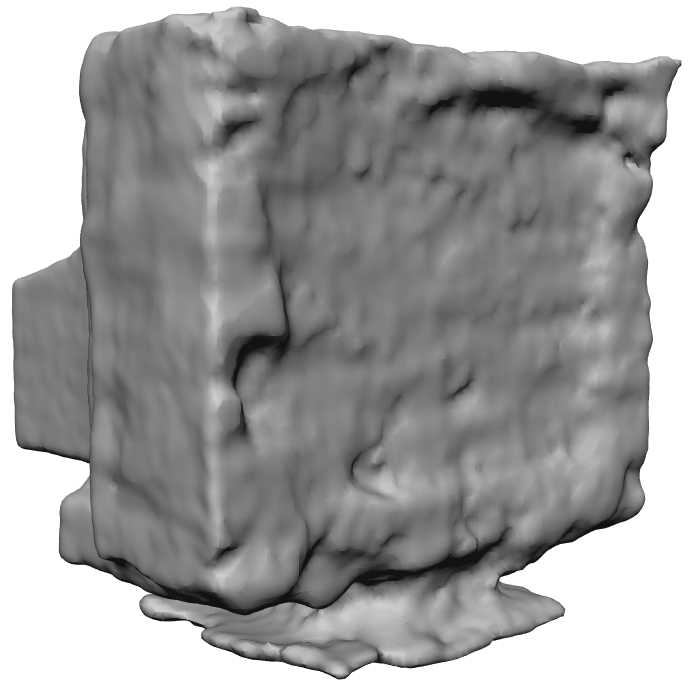}&
\includegraphics[width=\mywidth]{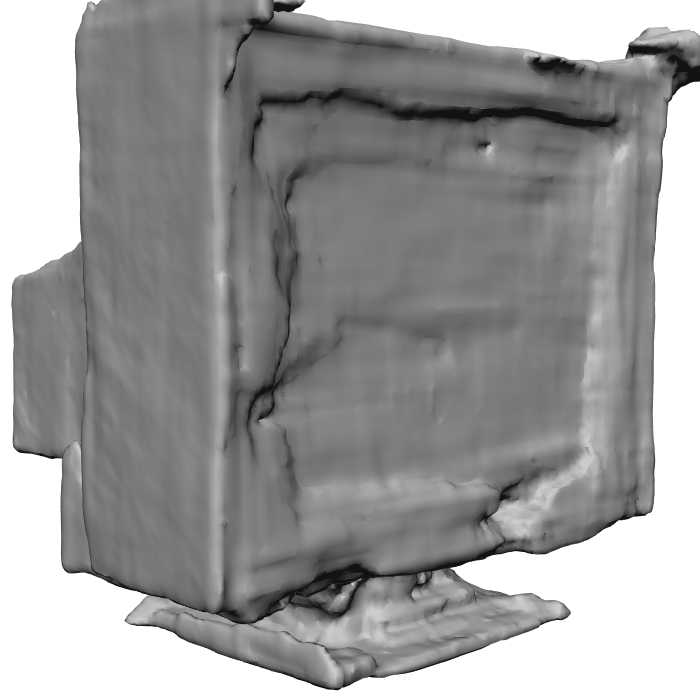}&
\\

\rotatebox{90}{\hspace{10mm}Bare}&
\includegraphics[width=\mywidth]{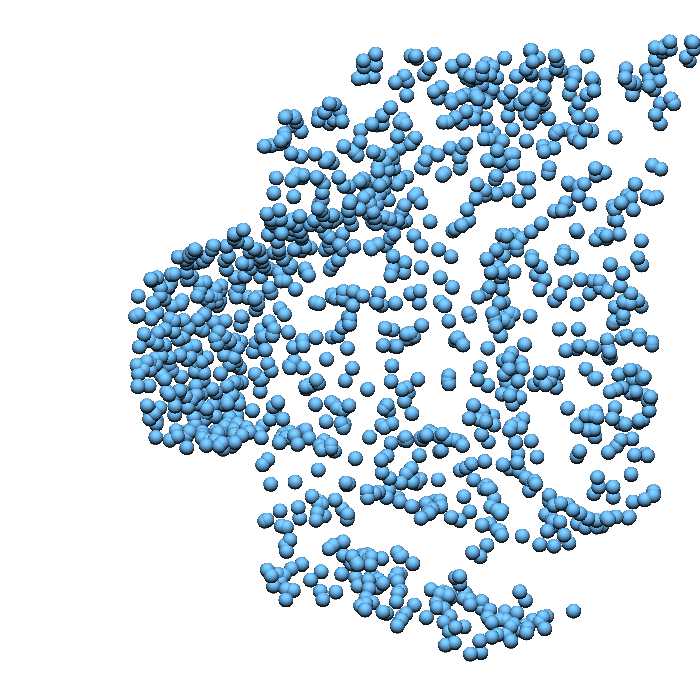}&
\includegraphics[width=\mywidth]{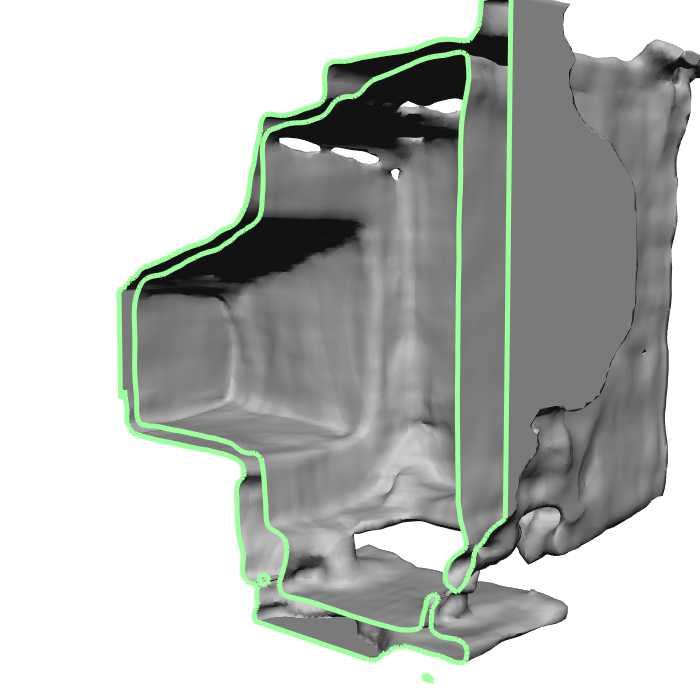}&
\includegraphics[width=\mywidth]{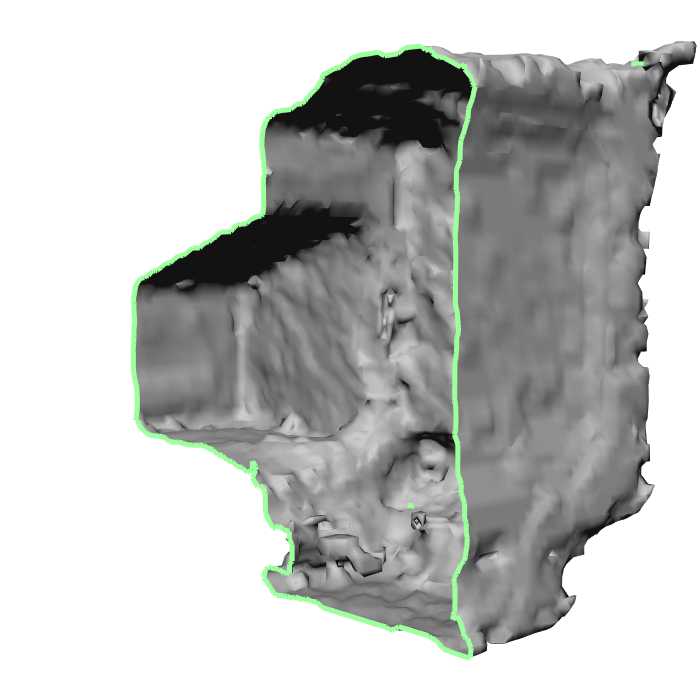}&
\includegraphics[width=\mywidth]{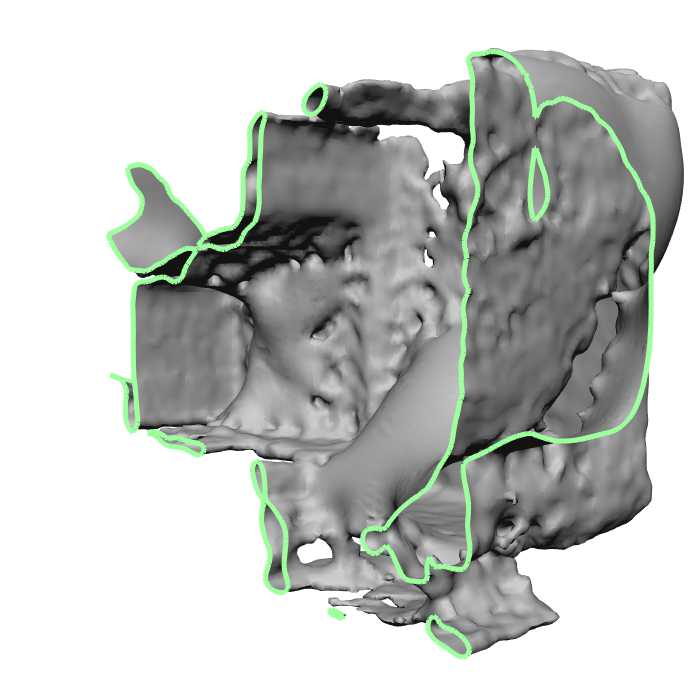}&
\includegraphics[width=\mywidth]{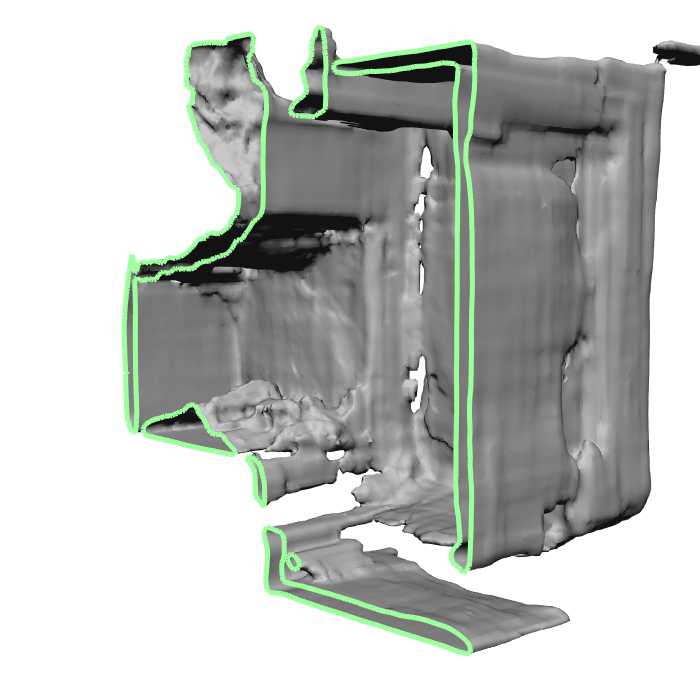}&
\multirow{2}{*}[3em]{\includegraphics[width=\mywidth]{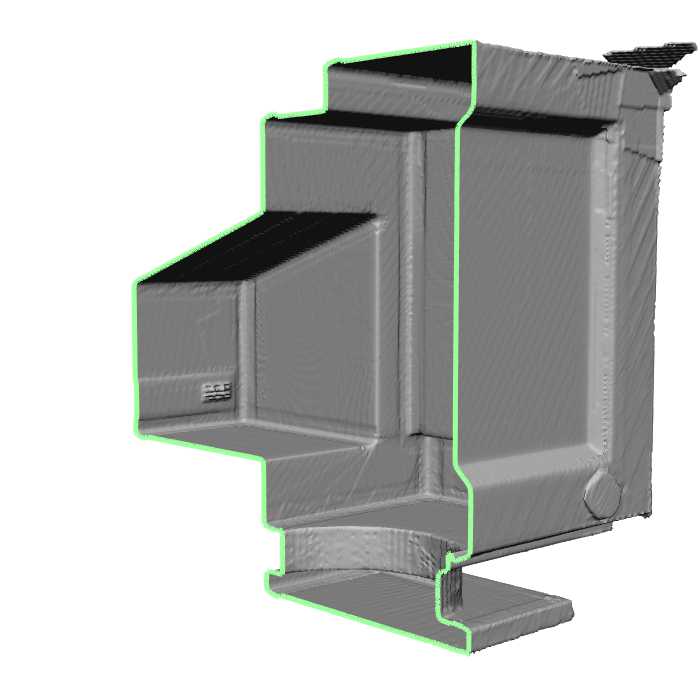}} \\
\rotatebox{90}{\hspace{4mm}Augmented}&
\includegraphics[width=\mywidth]{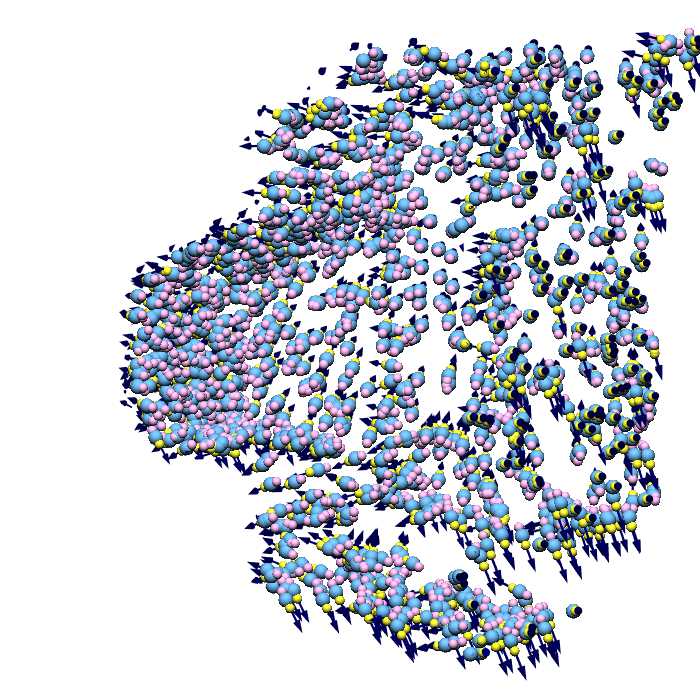}&
\includegraphics[width=\mywidth]{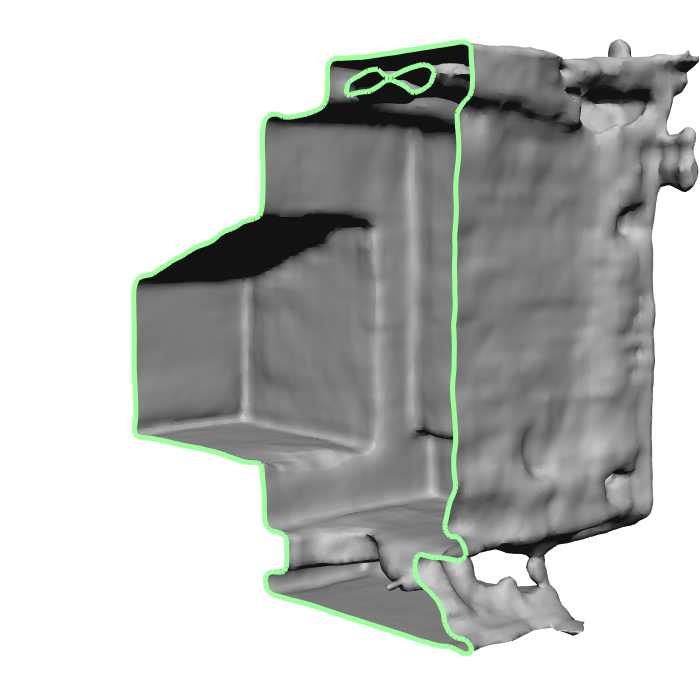}&
\includegraphics[width=\mywidth]{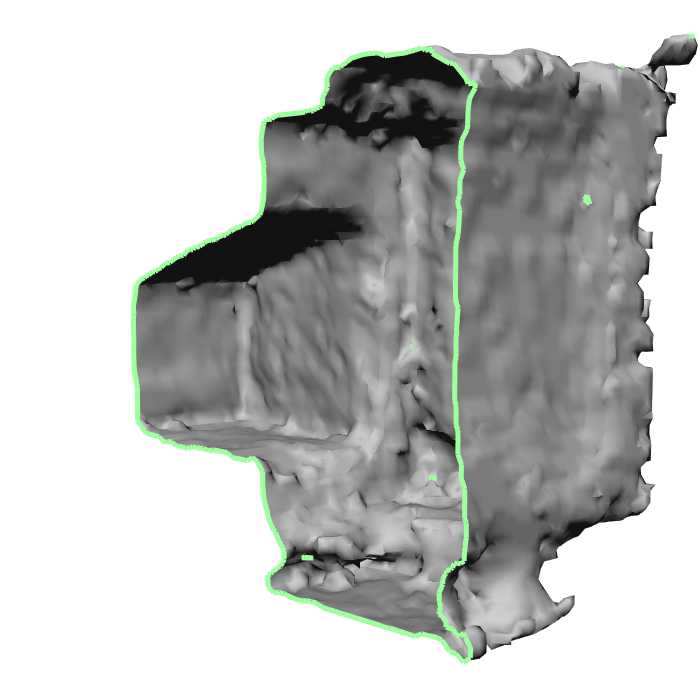}&
\includegraphics[width=\mywidth]{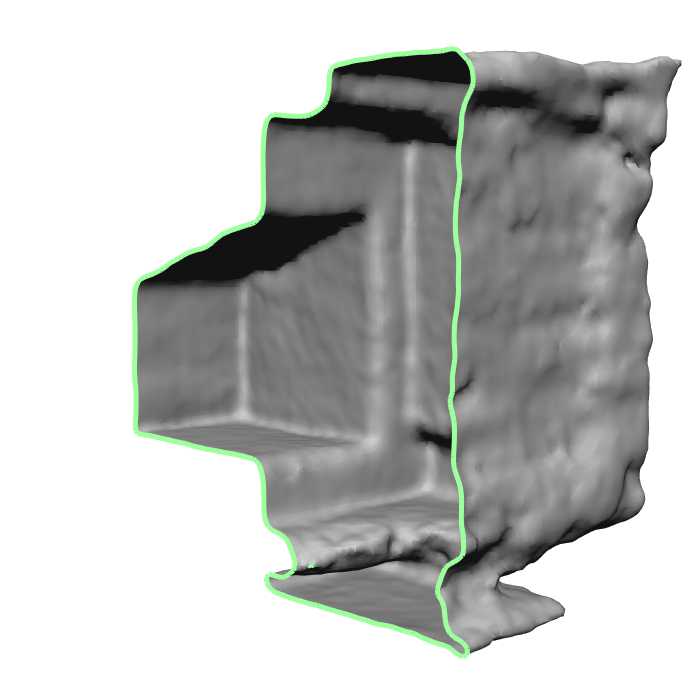}&
\includegraphics[width=\mywidth]{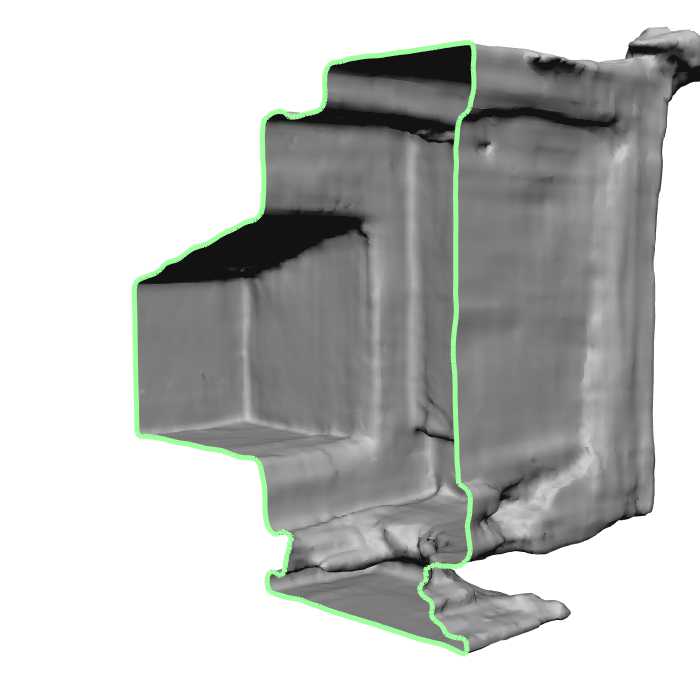}&
\\


&
Input &
ConvONet-3D~\cite{Peng2020}&
Points2Surf~\cite{points2surf}&
Shape\,As\,Points~\cite{Peng2021SAP}&
POCO~\cite{boulch2022poco}&
Ground Truth
\end{tabular}
	\caption{
		\textbf{Object-Level Reconstruction on ShapeNet.} 
		Reconstructed shapes from the ShapeNet test set using four different DSR methods trained on ModelNet10.
		Top rows of each object use the bare point cloud as input, and bottom rows use the point cloud augmented with visibility information. The last two rows show a cut
		of the reconstructions that are shown on the two other rows immediately above.
	}
	\label{fig:shapenet_suppmat}
\end{figure*}
We show the results of object-level reconstruction on ShapeNet in \figref{fig:shapenet_suppmat} from methods trained on ModelNet10. All methods benefit from added visibility information. In particular, ConvONet produces very accurate and complete surfaces of the unseen shape classes. 

A common problem for most baseline methods is the false reconstruction of hollow shapes with enclosed outside space. Using visibility information can address this issue and the artifacts then do not occur.

\subsection{ModelNet10}
\begin{figure*}[]
	\centering\small
	\newcommand{\mywidth}{0.115\textwidth}
\begin{tabular}{cc|cccc|c}

\rotatebox{90}{\hspace{10mm}Bare}&
\includegraphics[width=\mywidth]{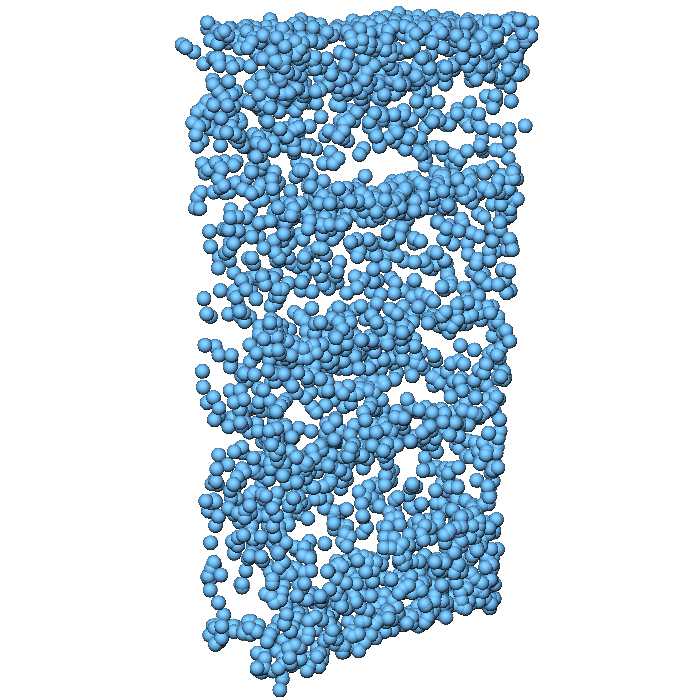}&
\includegraphics[width=\mywidth]{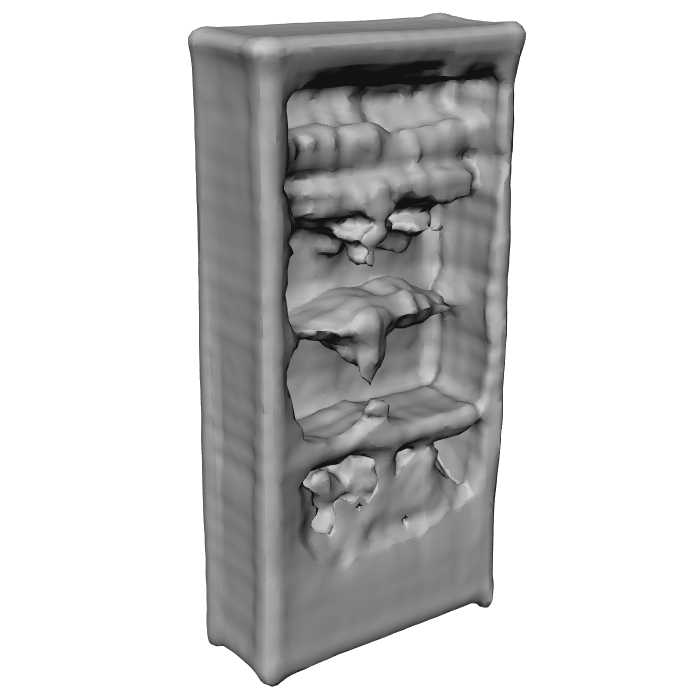}&
\includegraphics[width=\mywidth]{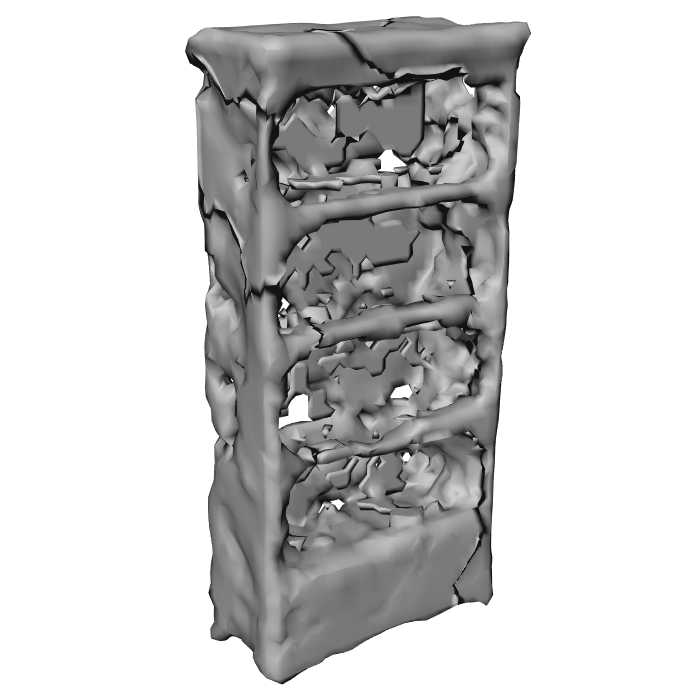}&
\includegraphics[width=\mywidth]{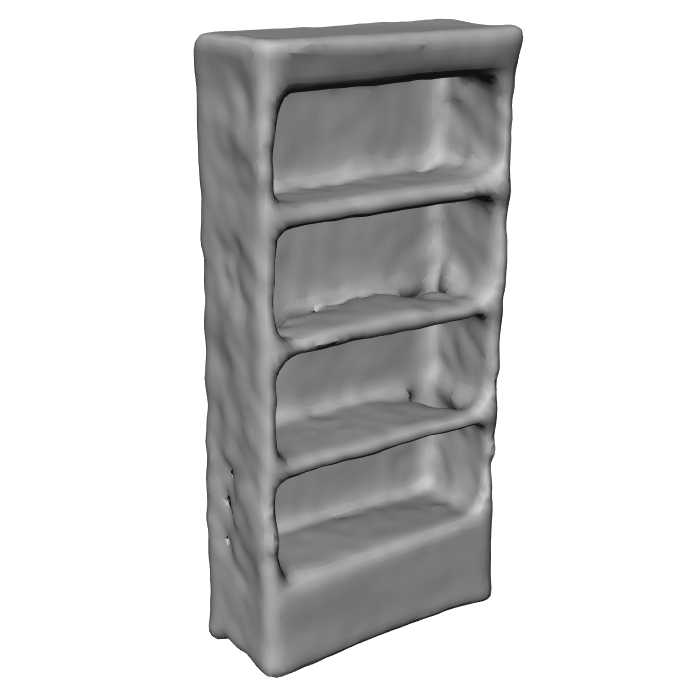}&
\includegraphics[width=\mywidth]{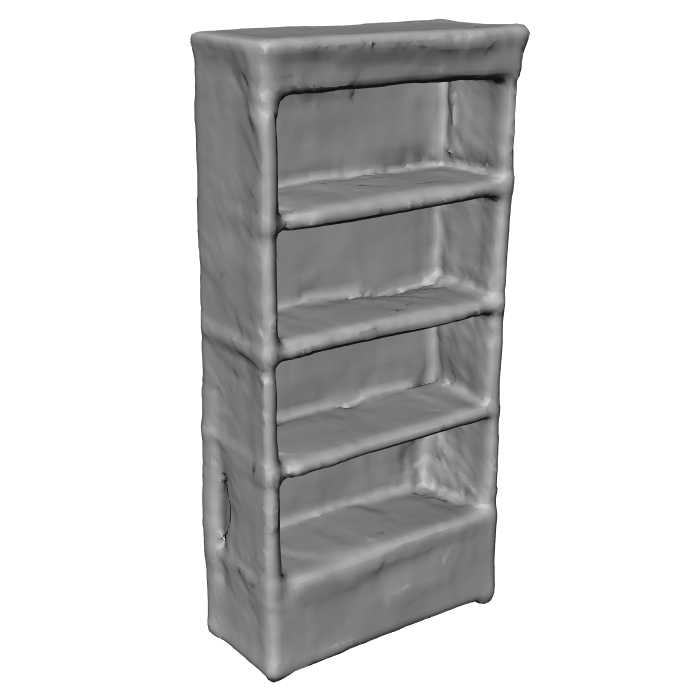}&
\multirow{2}{*}[3em]{\includegraphics[width=\mywidth]{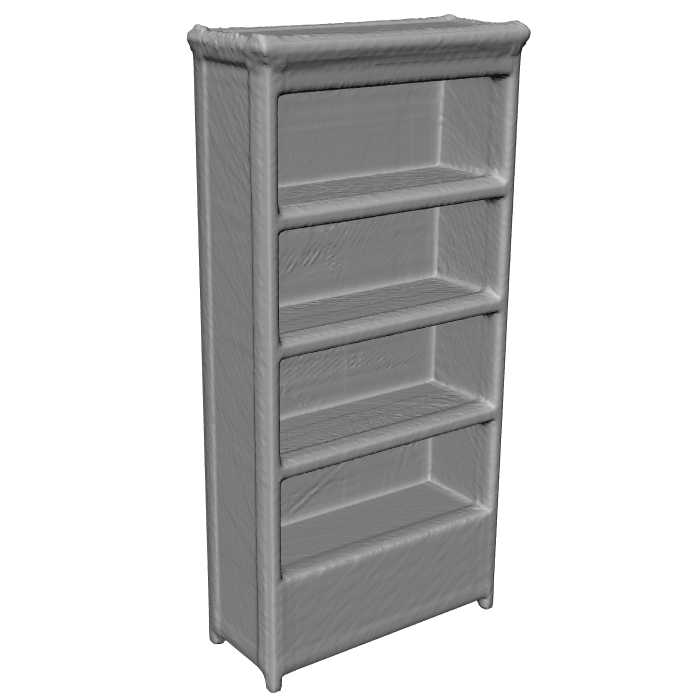}}
\\
\rotatebox{90}{\hspace{5mm}Augmented}&
\includegraphics[width=\mywidth]{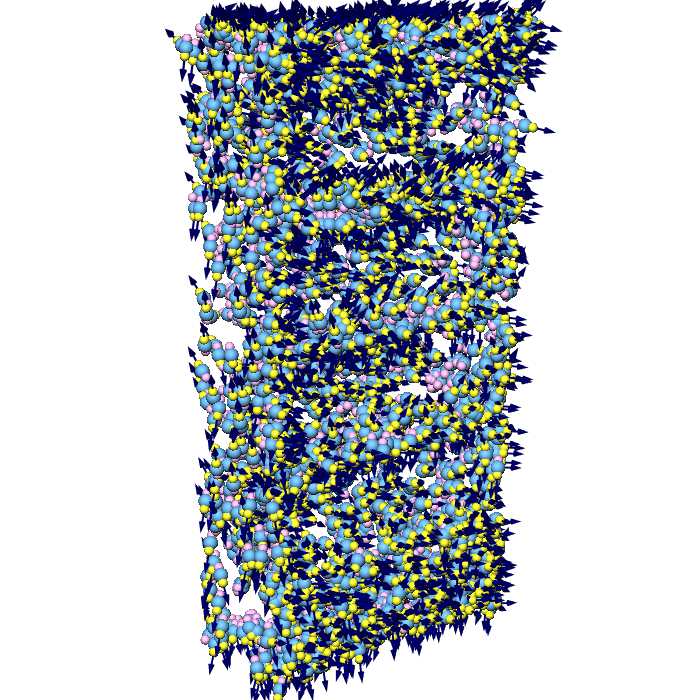}&
\includegraphics[width=\mywidth]{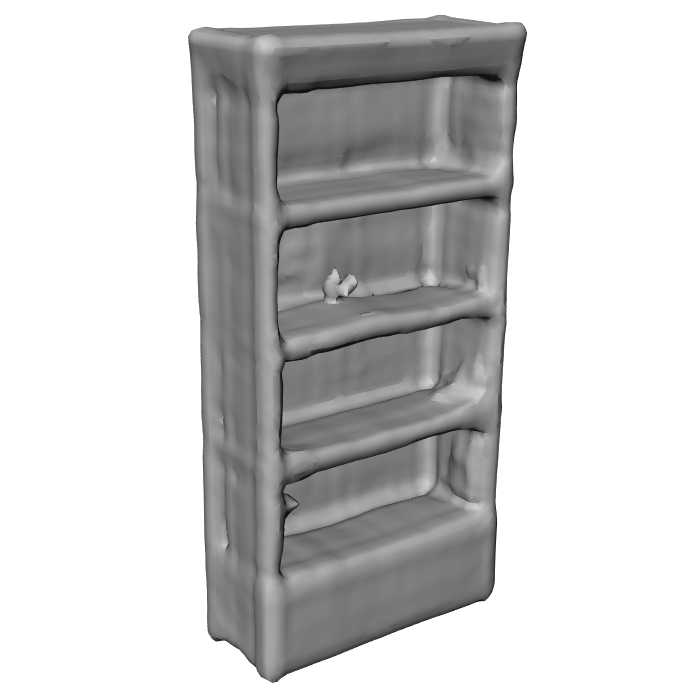}&
\includegraphics[width=\mywidth]{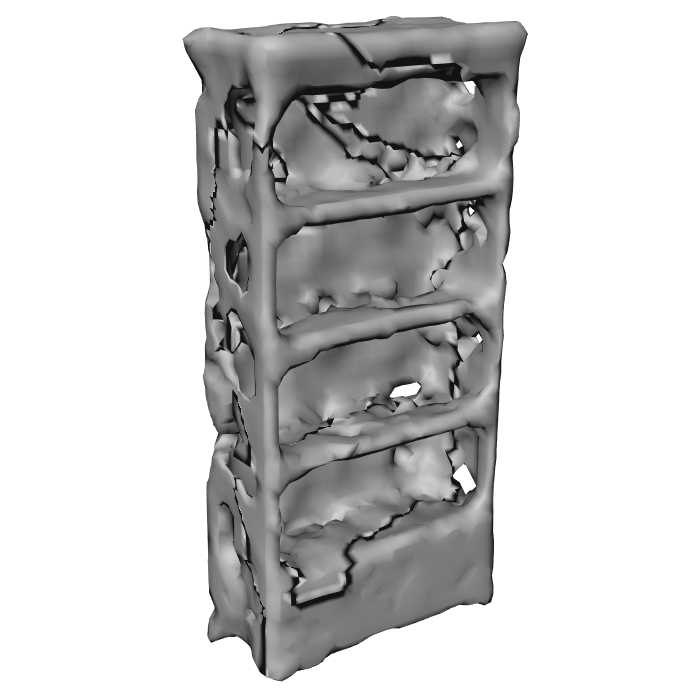}&
\includegraphics[width=\mywidth]{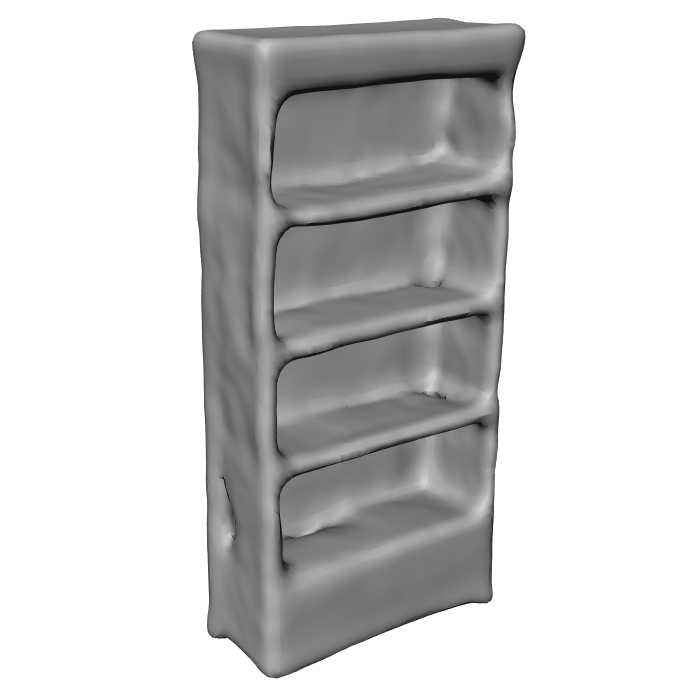}&
\includegraphics[width=\mywidth]{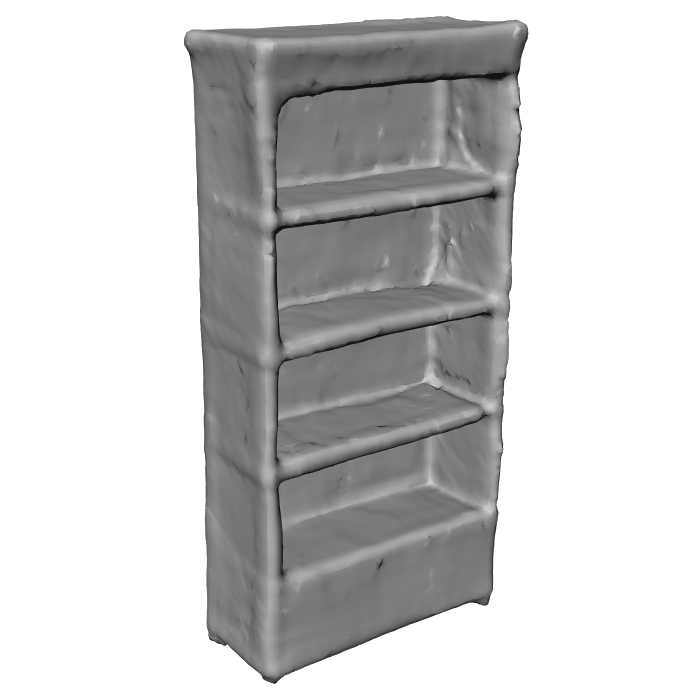}&
\\

\rotatebox{90}{\hspace{3mm}Bare}&
\includegraphics[width=\mywidth,trim=0 140 0 140,clip]{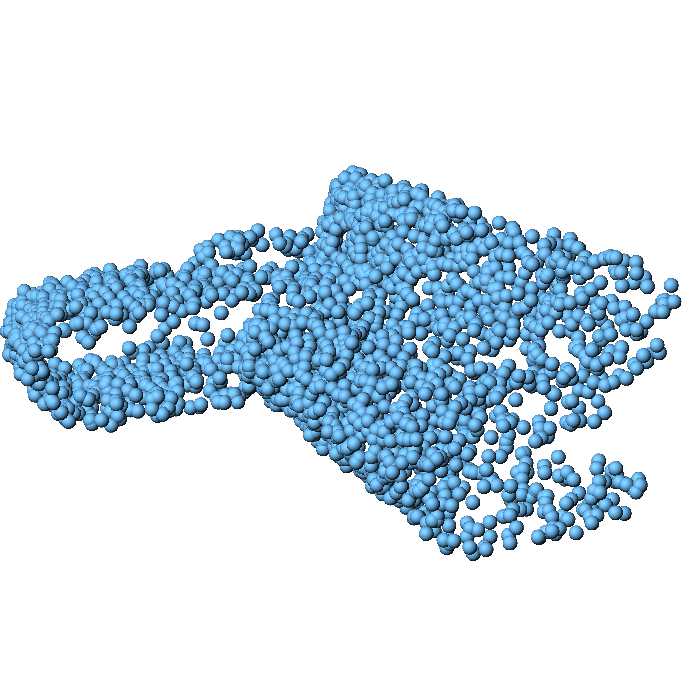}&
\includegraphics[width=\mywidth,trim=0 140 0 140,clip]{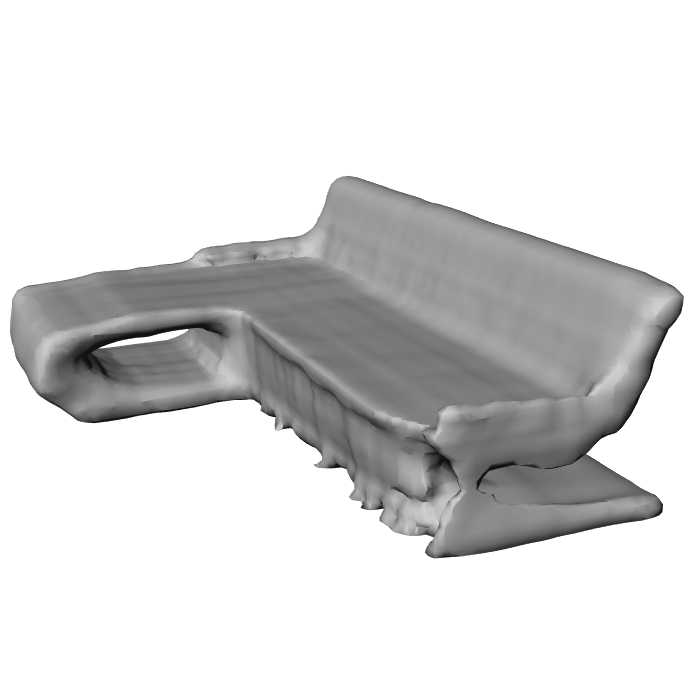}&
\includegraphics[width=\mywidth,trim=0 140 0 140,clip]{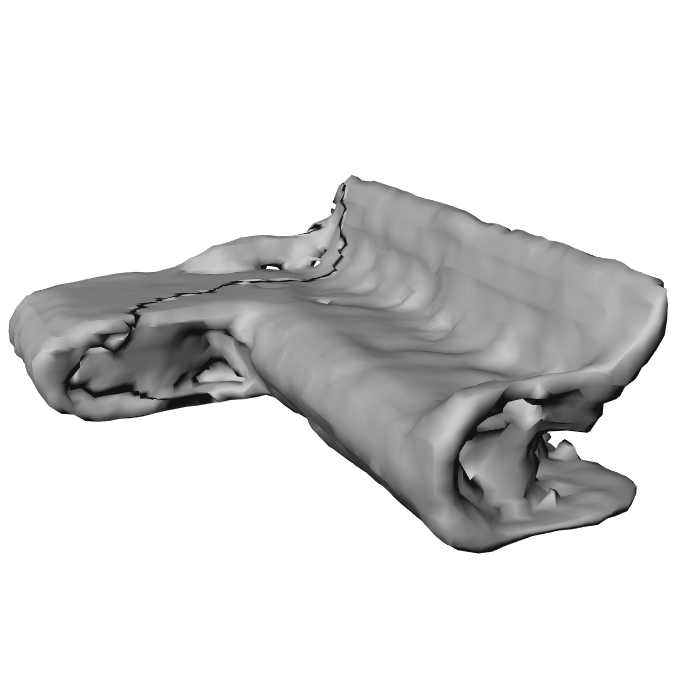}&
\includegraphics[width=\mywidth,trim=0 140 0 140,clip]{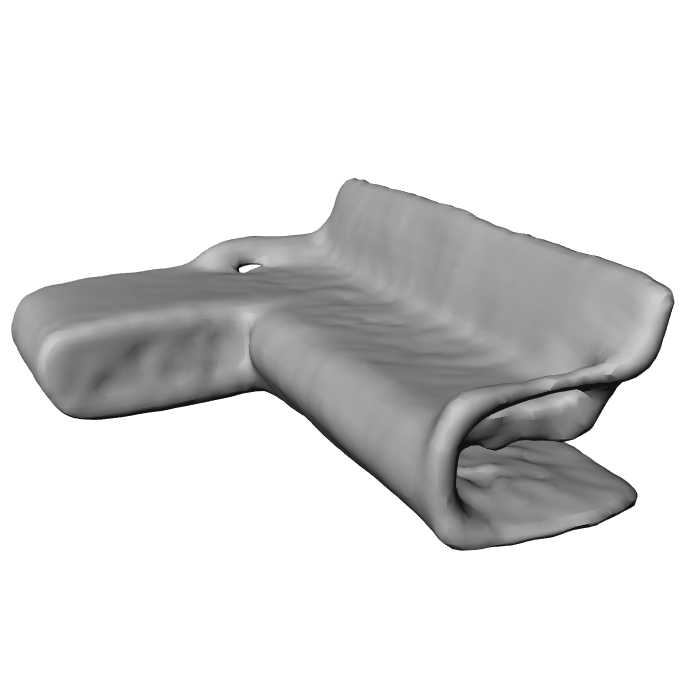}&
\includegraphics[width=\mywidth,trim=0 140 0 140,clip]{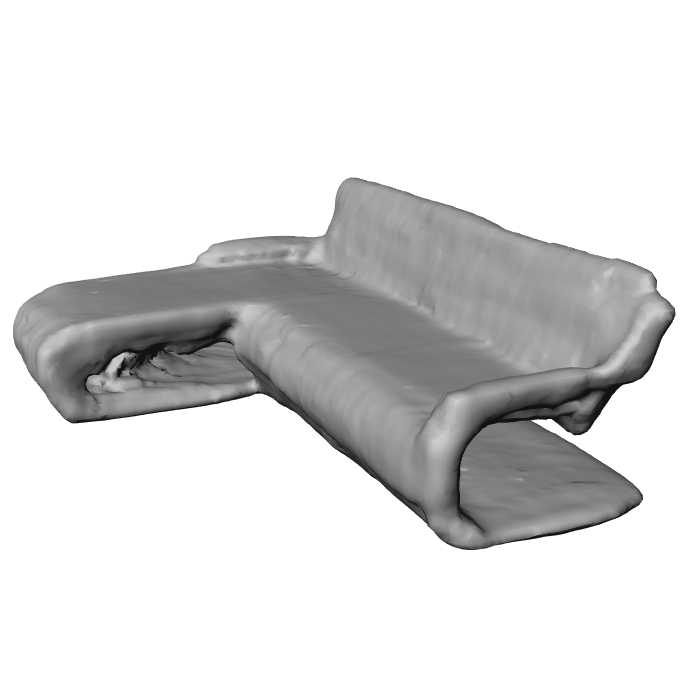}&
\multirow{2}{*}[3em]{\includegraphics[width=\mywidth,trim=0 140 0 140,clip]{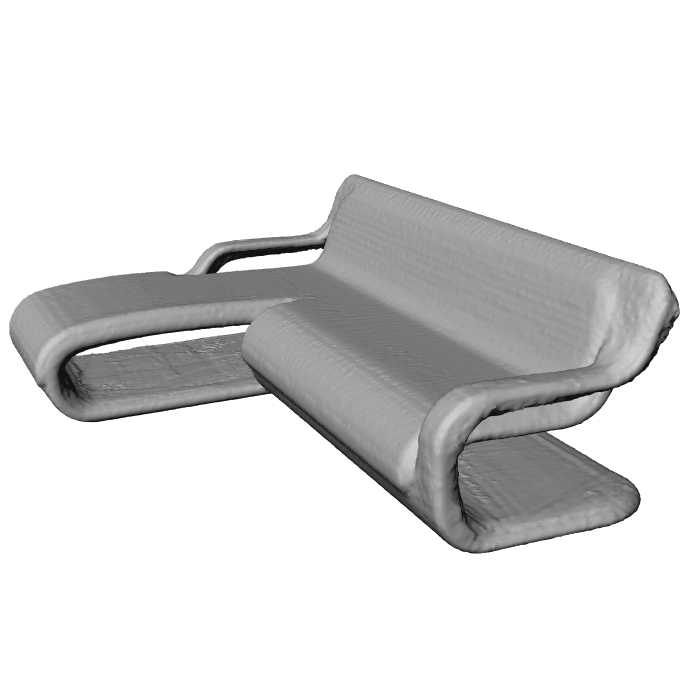}} \\
\rotatebox{90}{\hspace{0mm}Augmented}&
\includegraphics[width=\mywidth,trim=0 140 0 140,clip]{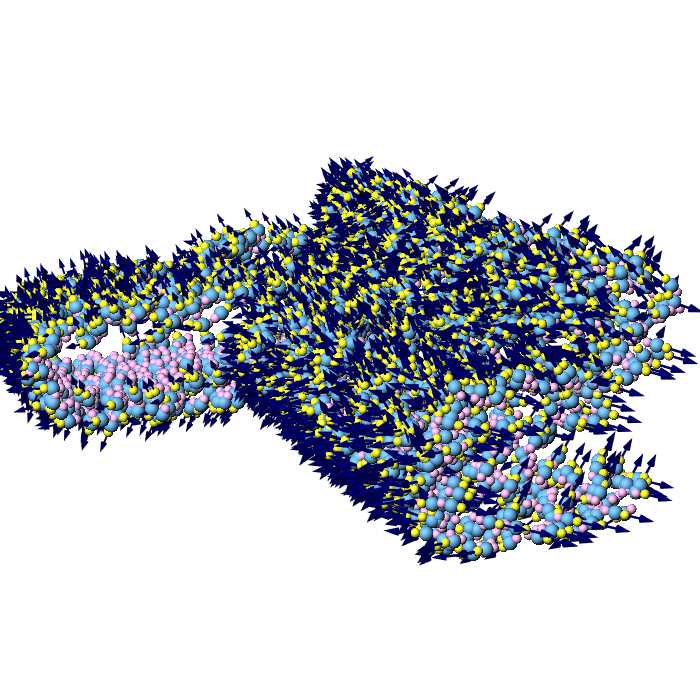}&
\includegraphics[width=\mywidth,trim=0 140 0 140,clip]{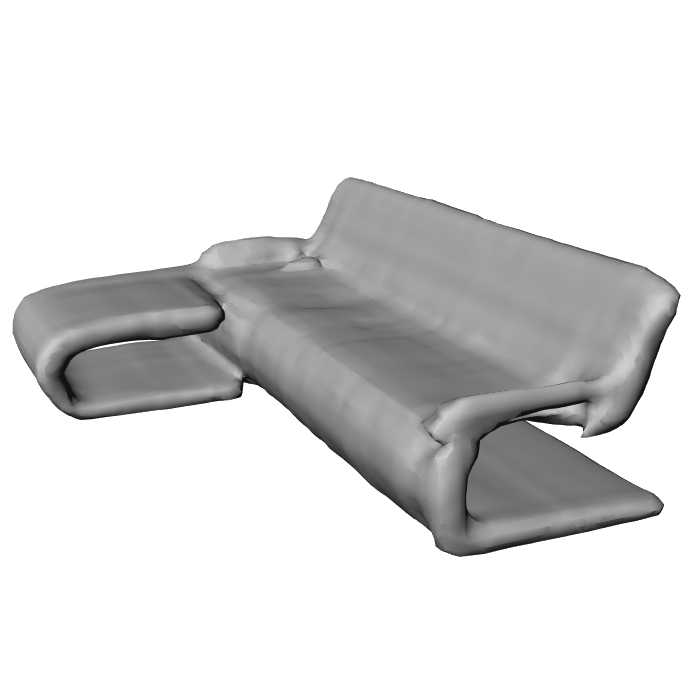}&
\includegraphics[width=\mywidth,trim=0 140 0 140,clip]{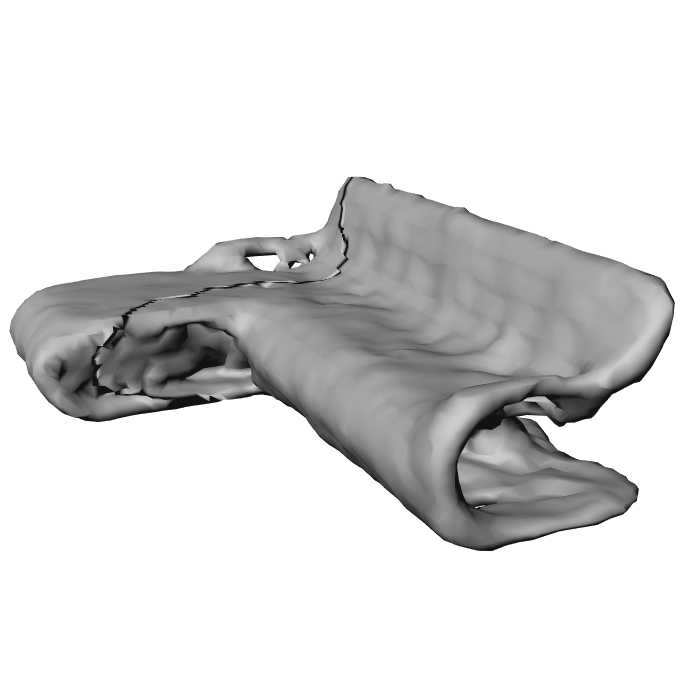}&
\includegraphics[width=\mywidth,trim=0 140 0 140,clip]{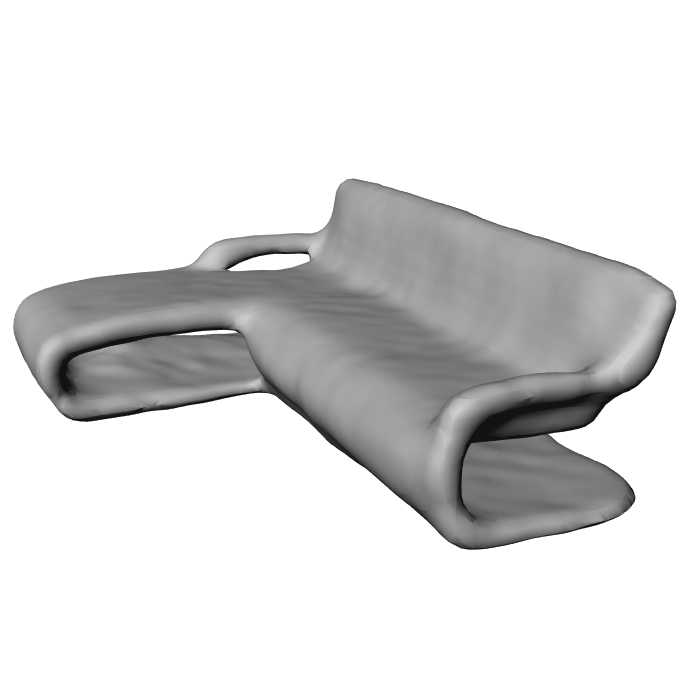}&
\includegraphics[width=\mywidth,trim=0 140 0 140,clip]{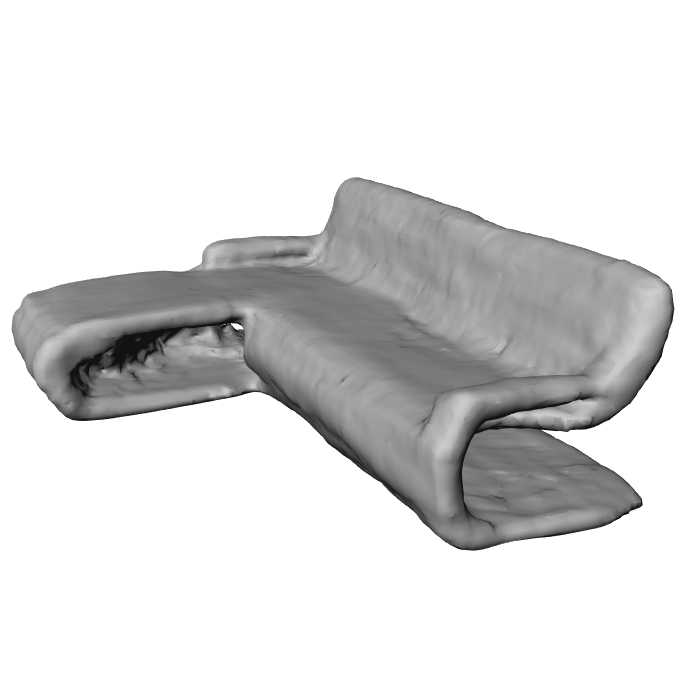}&
\\

\rotatebox{90}{\hspace{6mm}Bare}&
\includegraphics[width=\mywidth]{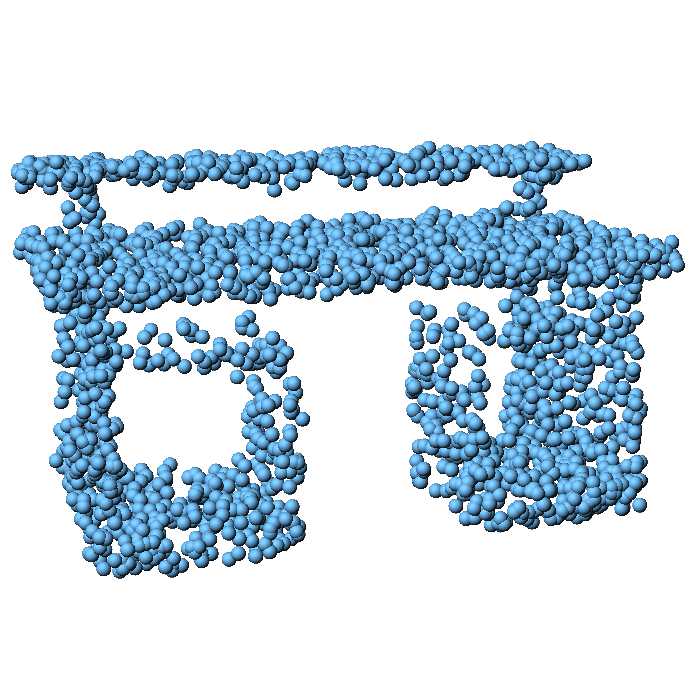}&
\includegraphics[width=\mywidth]{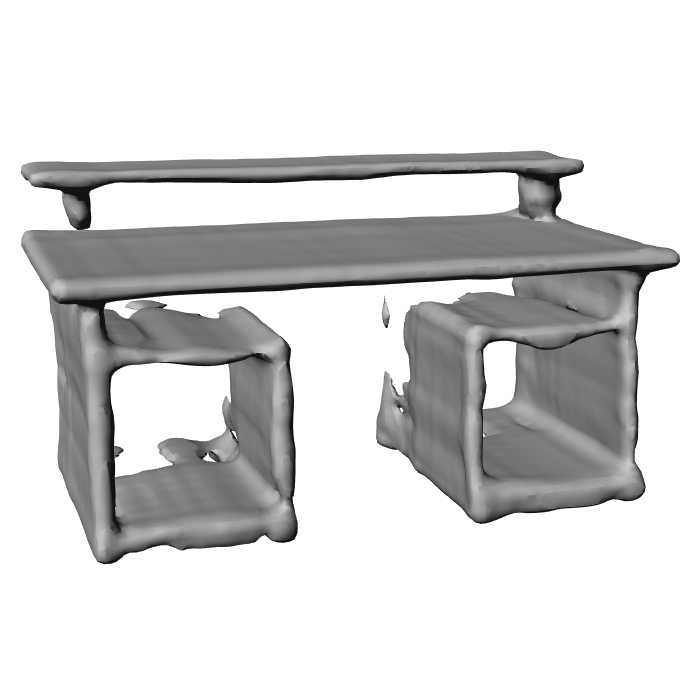}&
\includegraphics[width=\mywidth]{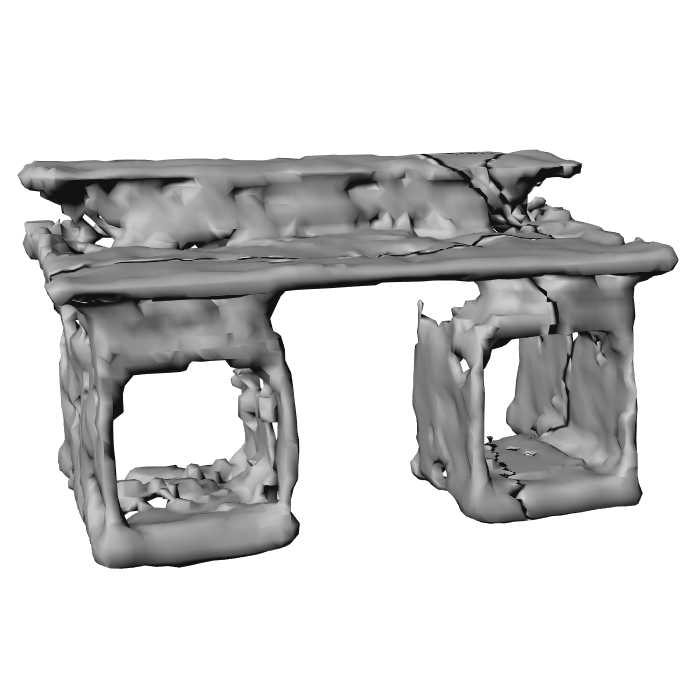}&
\includegraphics[width=\mywidth]{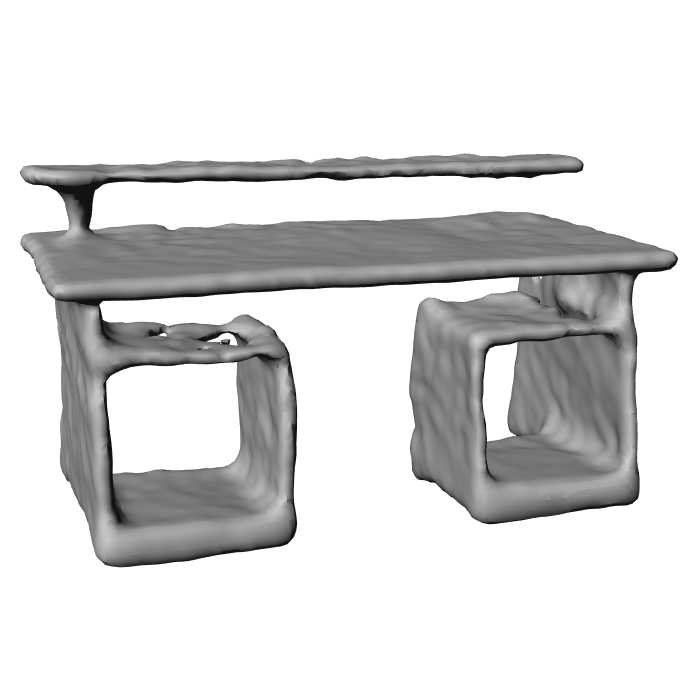}&
\includegraphics[width=\mywidth]{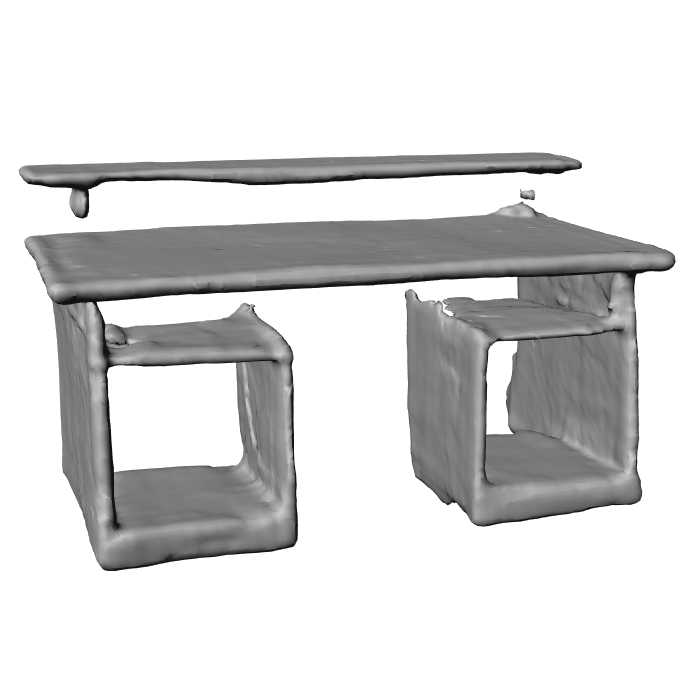}&
\multirow{2}{*}[3em]{\includegraphics[width=\mywidth]{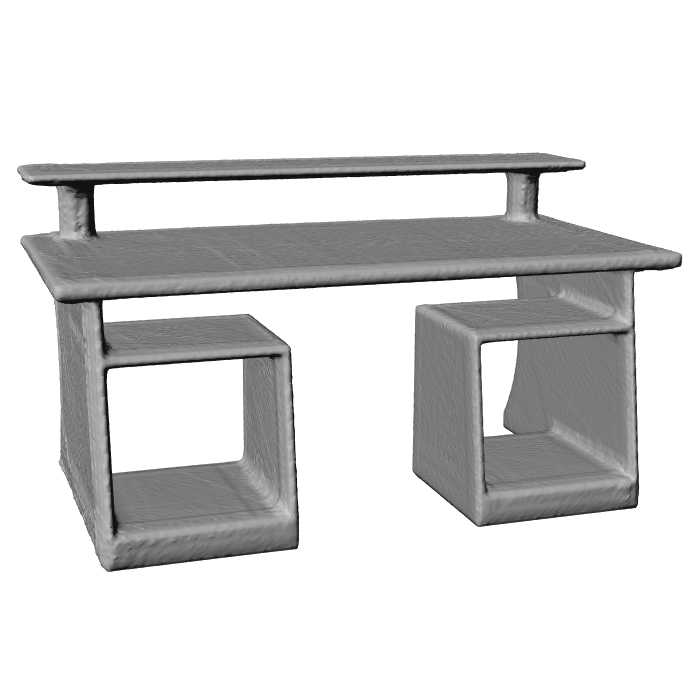}} \\
\rotatebox{90}{\hspace{3mm}Augmented}&
\includegraphics[width=\mywidth]{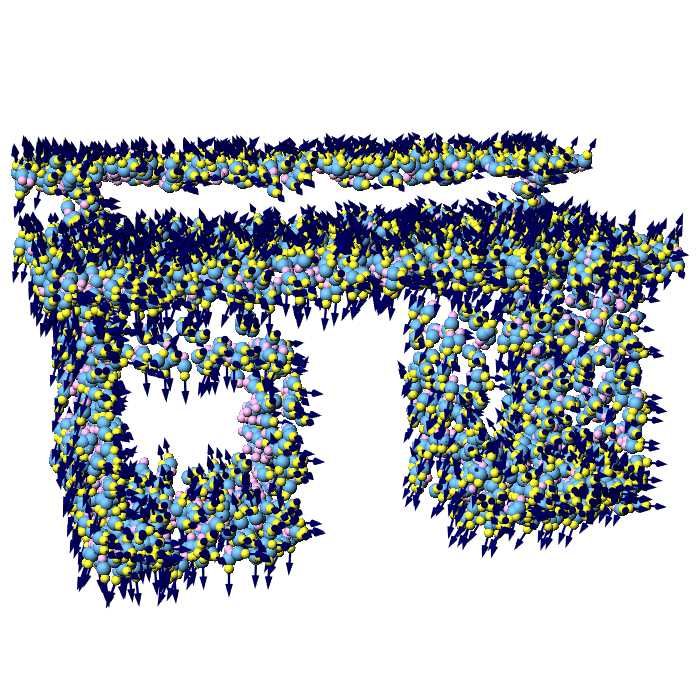}&
\includegraphics[width=\mywidth]{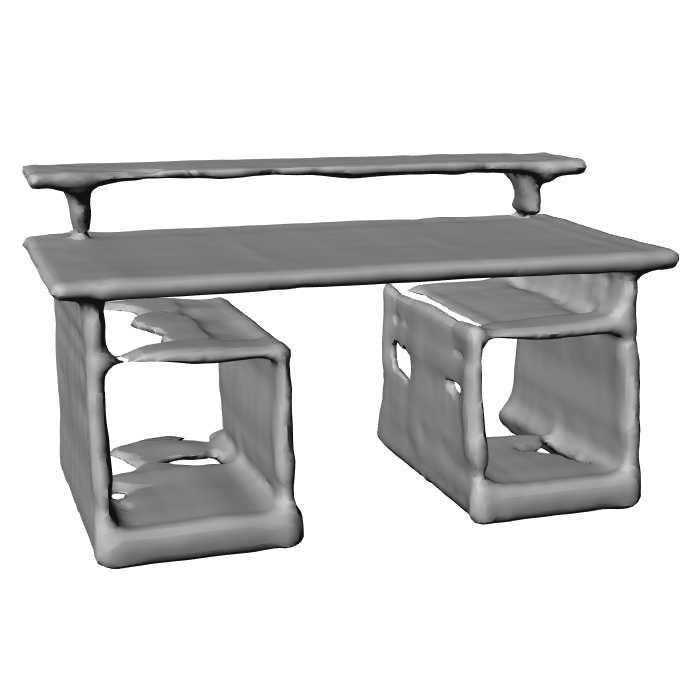}&
\includegraphics[width=\mywidth]{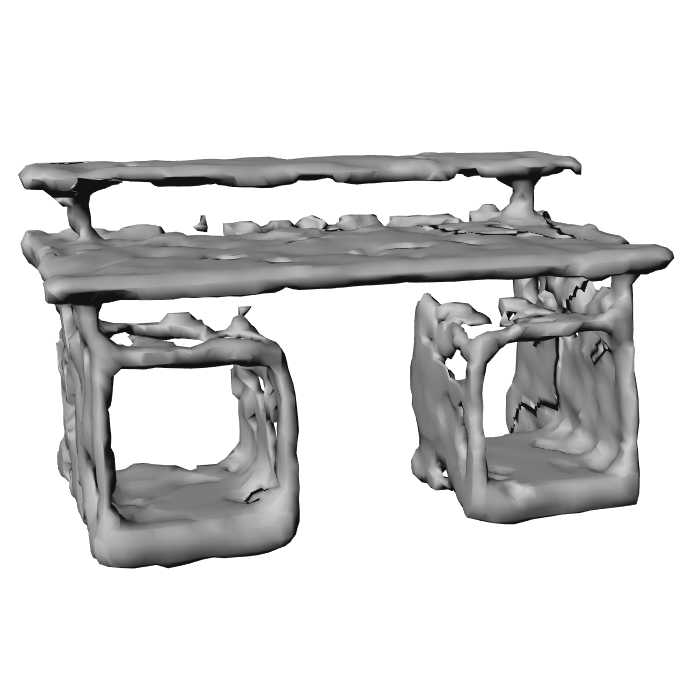}&
\includegraphics[width=\mywidth]{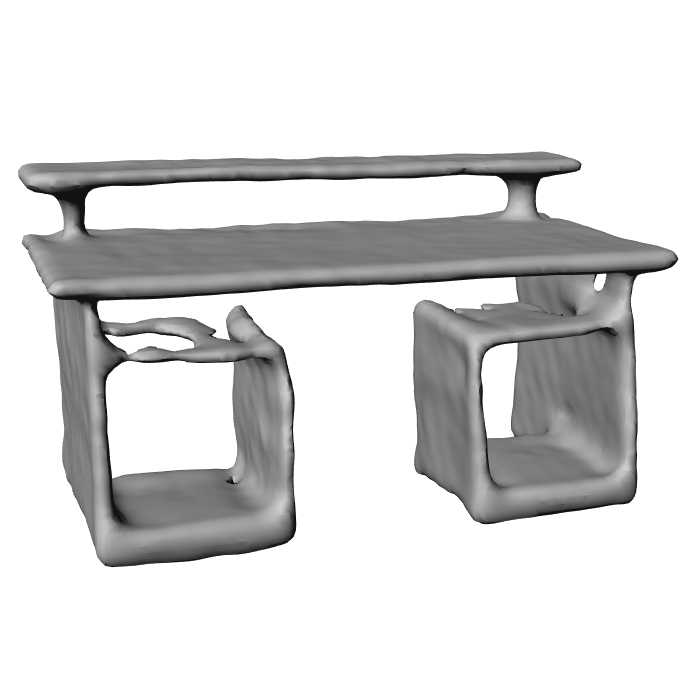}&
\includegraphics[width=\mywidth]{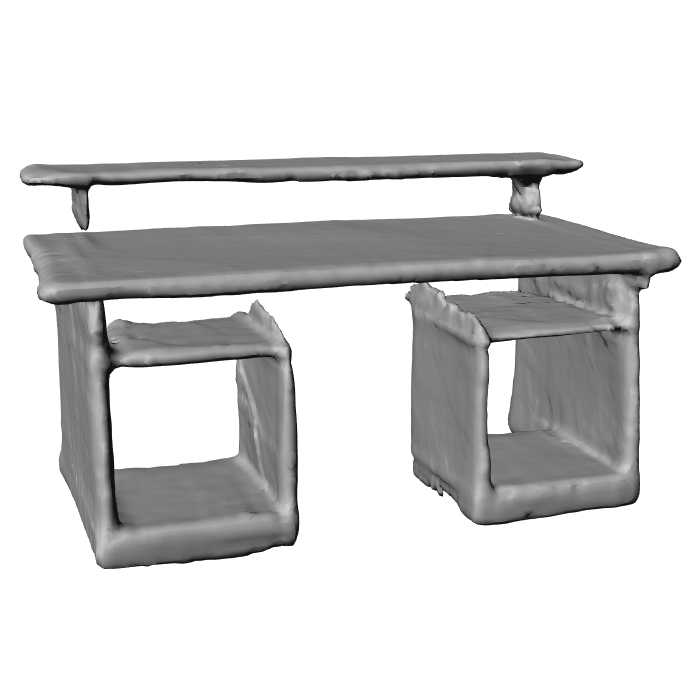}&
\\

\rotatebox{90}{\hspace{6mm}Bare}&
\includegraphics[width=\mywidth]{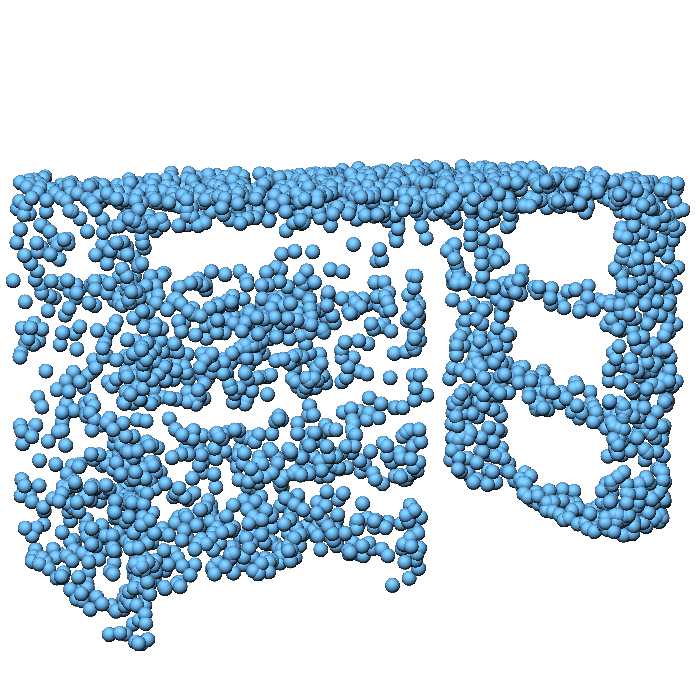}&
\includegraphics[width=\mywidth]{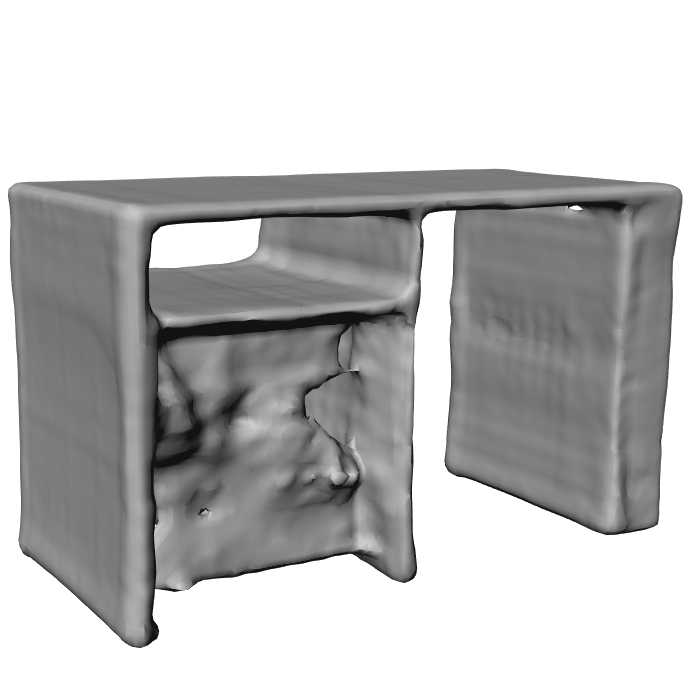}&
\includegraphics[width=\mywidth]{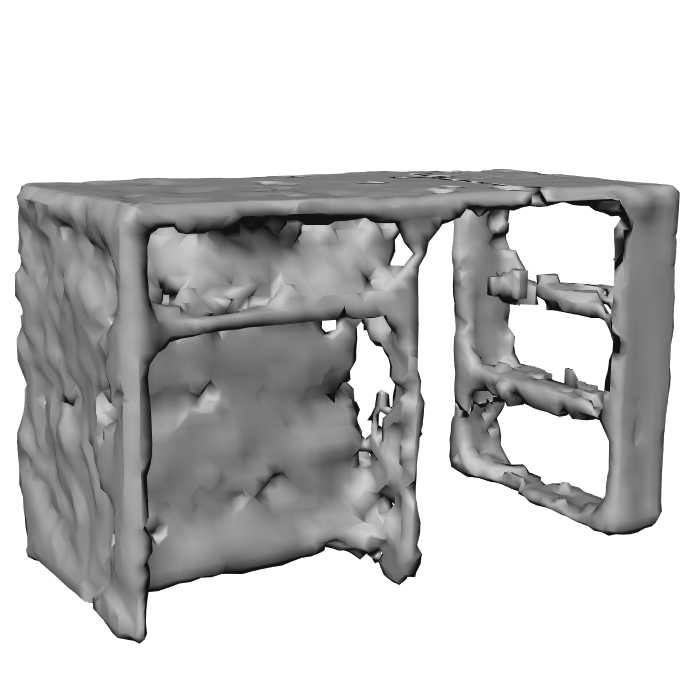}&
\includegraphics[width=\mywidth]{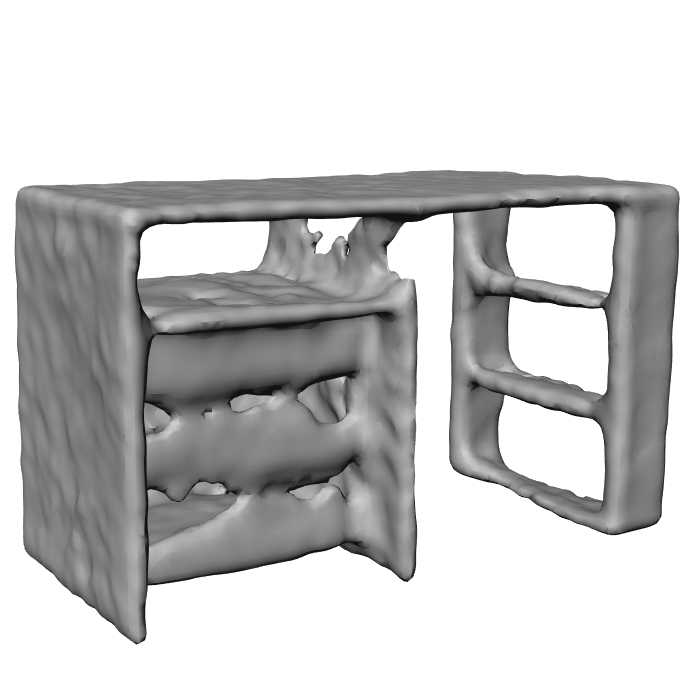}&
\includegraphics[width=\mywidth]{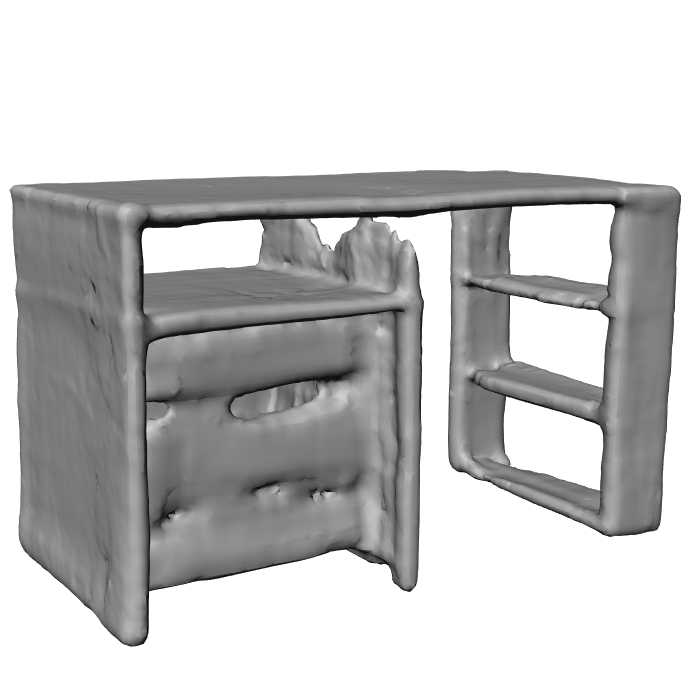}&
\multirow{2}{*}[3em]{\includegraphics[width=\mywidth]{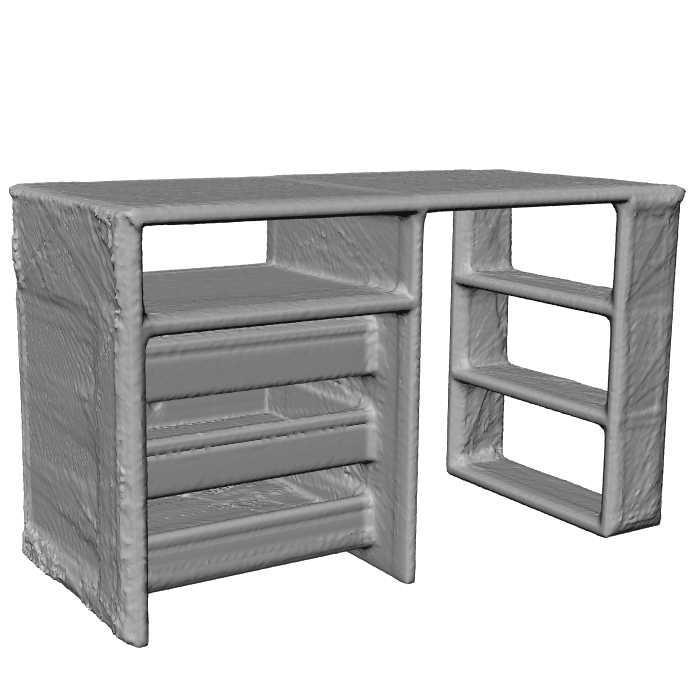}} 
\\
\rotatebox{90}{\hspace{2mm}Augmented}&
\includegraphics[width=\mywidth]{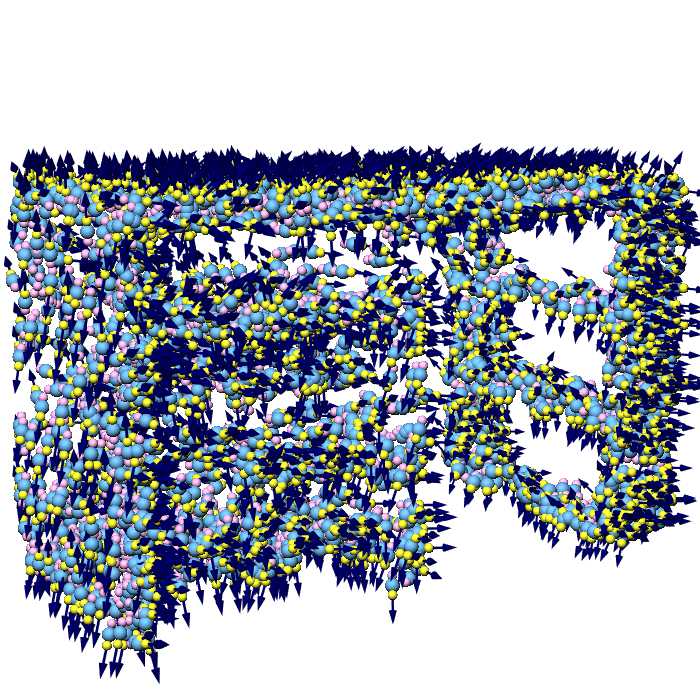}&
\includegraphics[width=\mywidth]{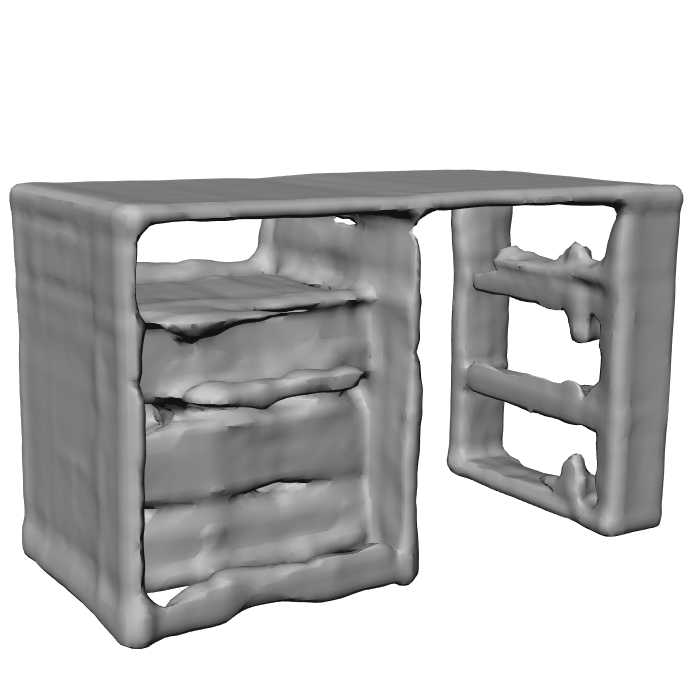}&
\includegraphics[width=\mywidth]{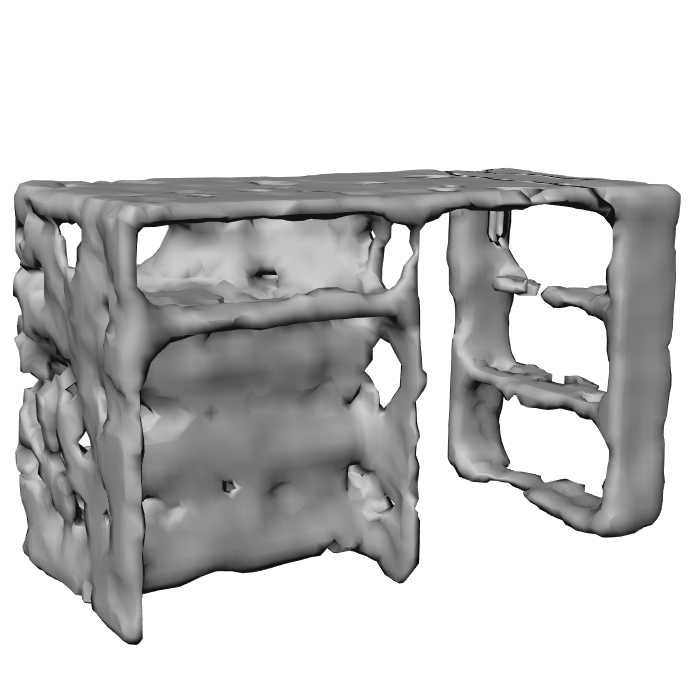}&
\includegraphics[width=\mywidth]{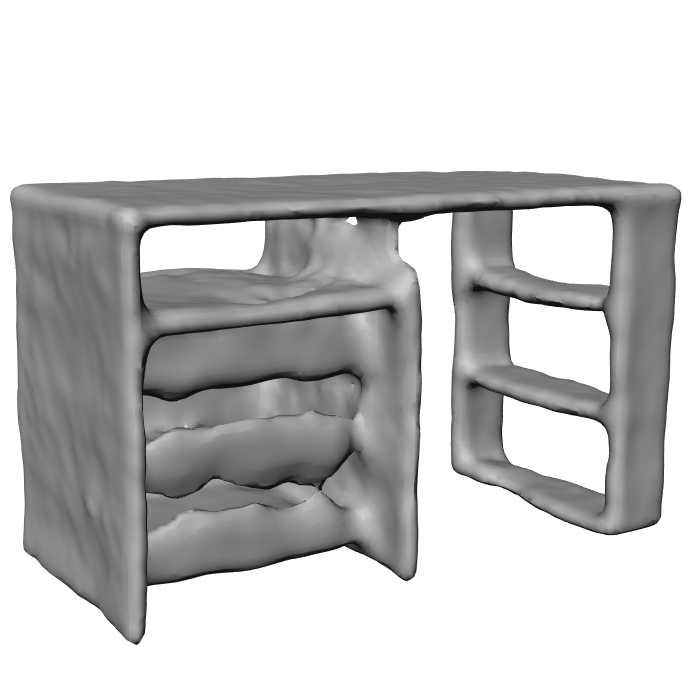}&
\includegraphics[width=\mywidth]{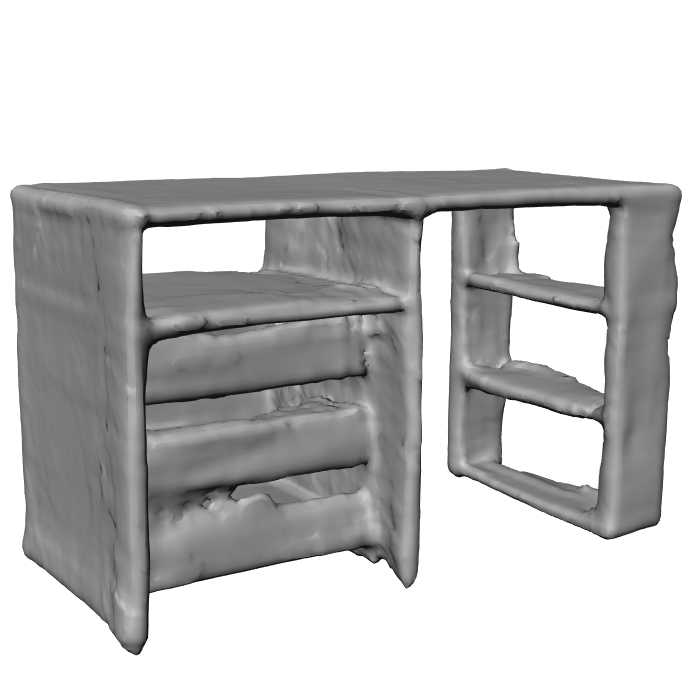}&
\\

\rotatebox{90}{\hspace{10mm}Bare}&
\includegraphics[width=\mywidth]{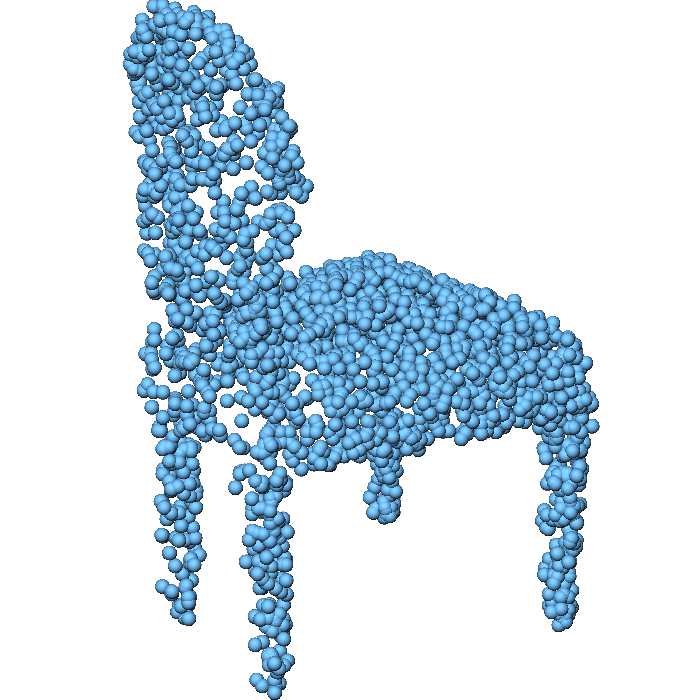}&
\includegraphics[width=\mywidth]{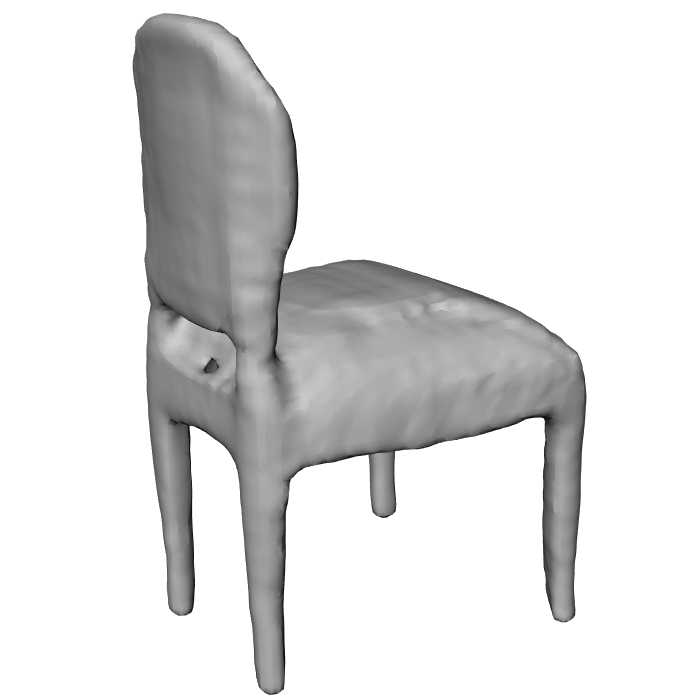}&
\includegraphics[width=\mywidth]{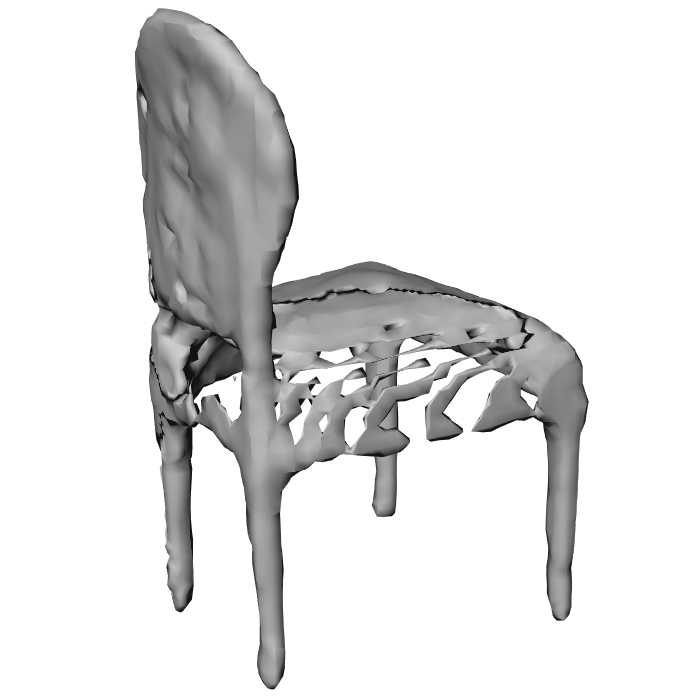}&
\includegraphics[width=\mywidth]{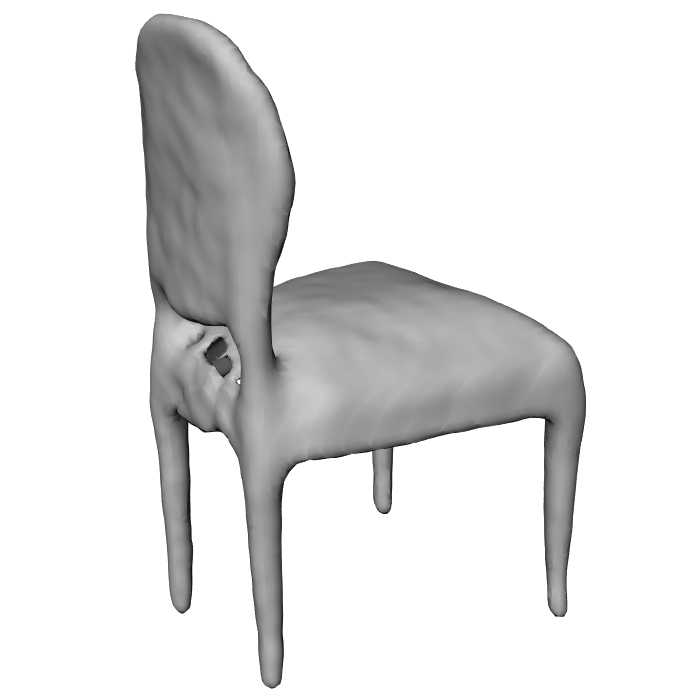}&
\includegraphics[width=\mywidth]{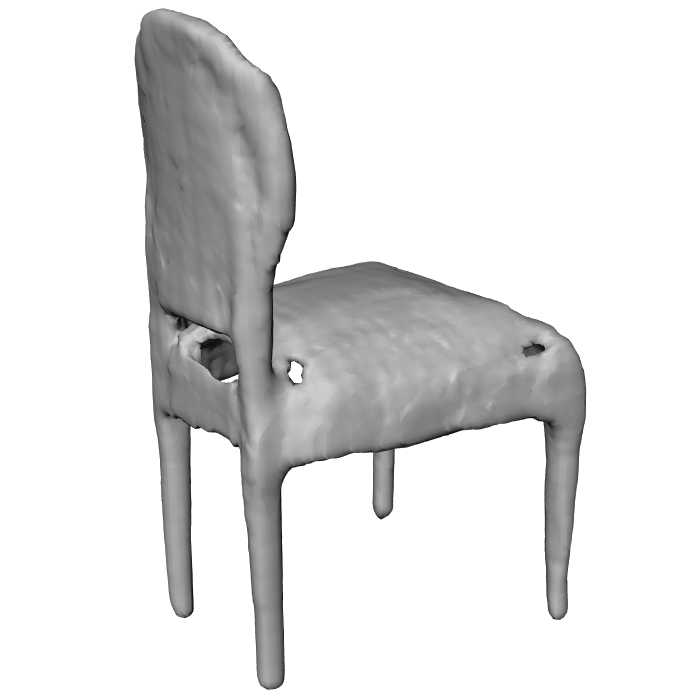}&
\multirow{2}{*}[3em]{\includegraphics[width=\mywidth]{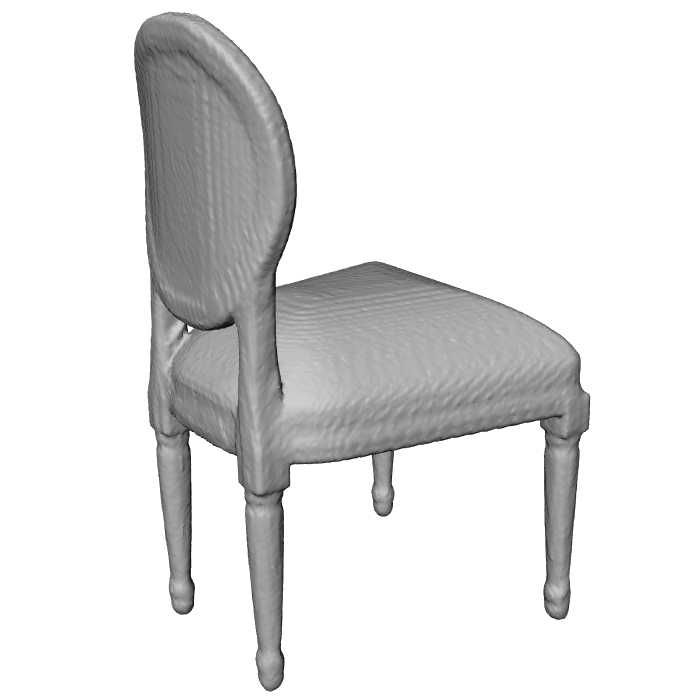}} 
\\
\rotatebox{90}{\hspace{5mm}Augmented}&
\includegraphics[width=\mywidth]{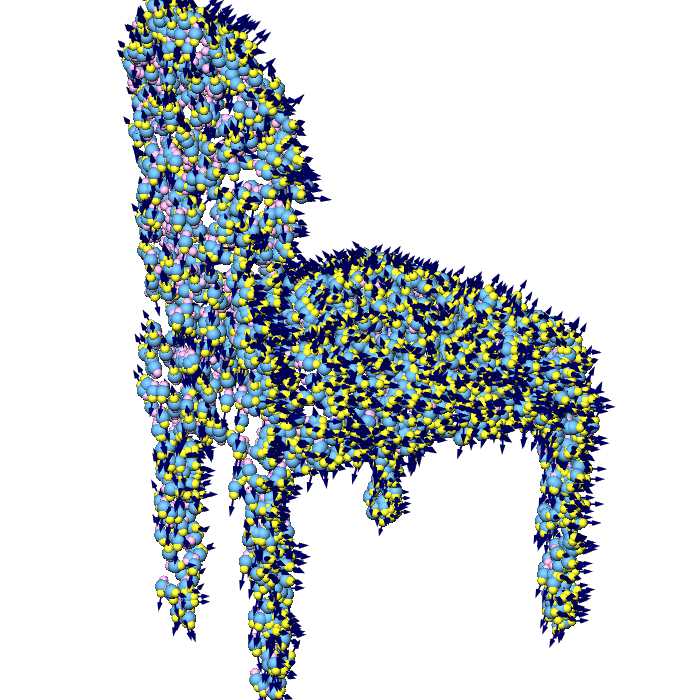}&
\includegraphics[width=\mywidth]{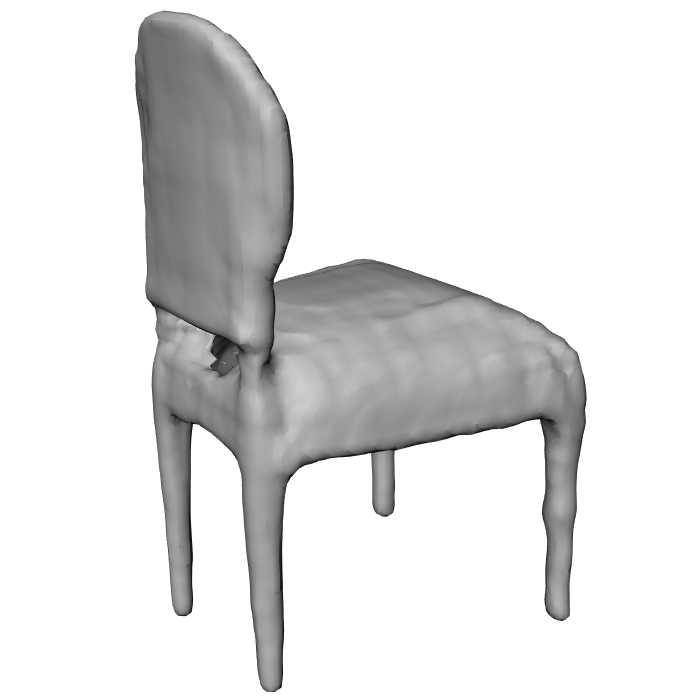}&
\includegraphics[width=\mywidth]{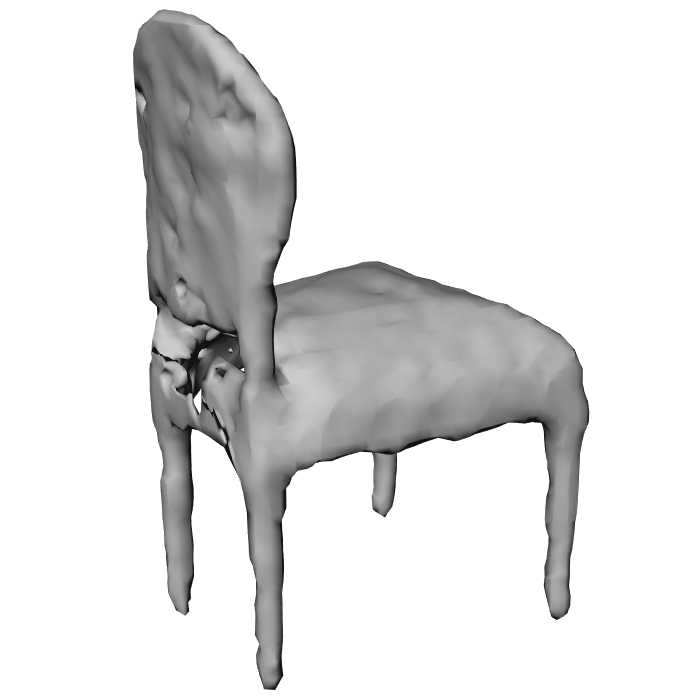}&
\includegraphics[width=\mywidth]{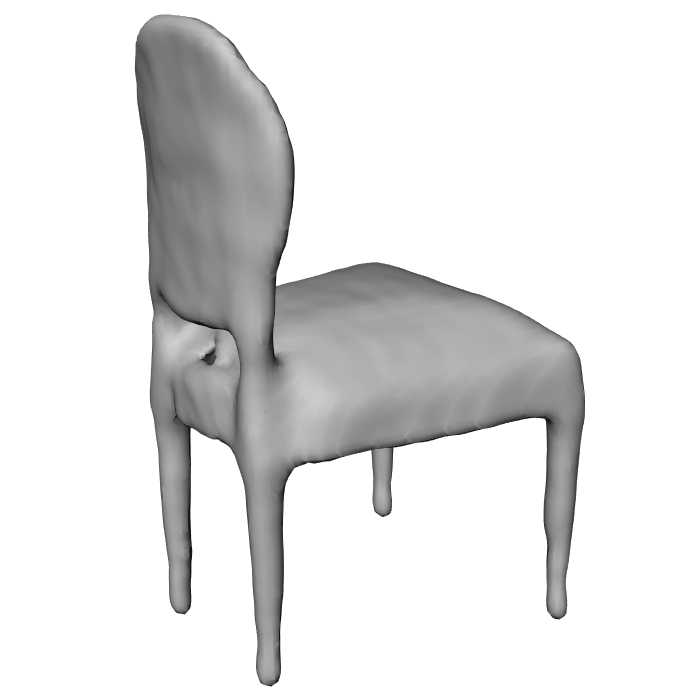}&
\includegraphics[width=\mywidth]{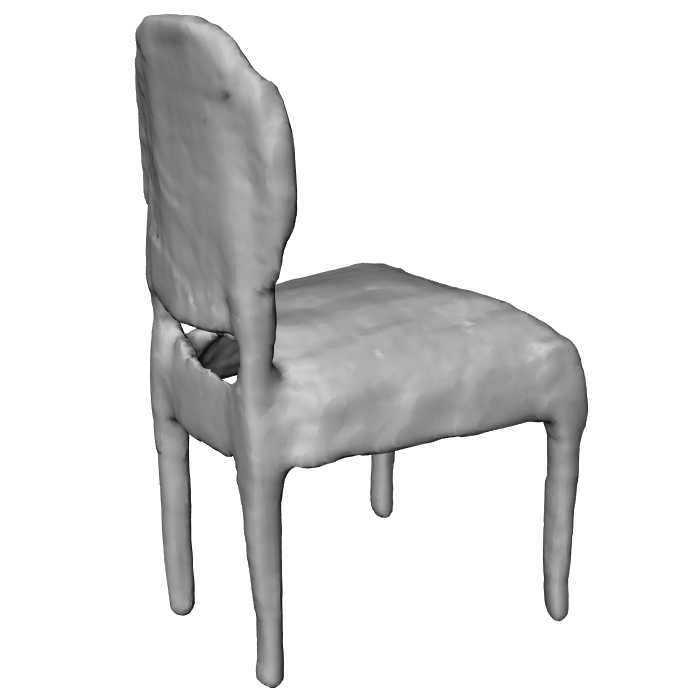}&
\\

\rotatebox{90}{\hspace{8mm}Bare}&
\includegraphics[width=\mywidth]{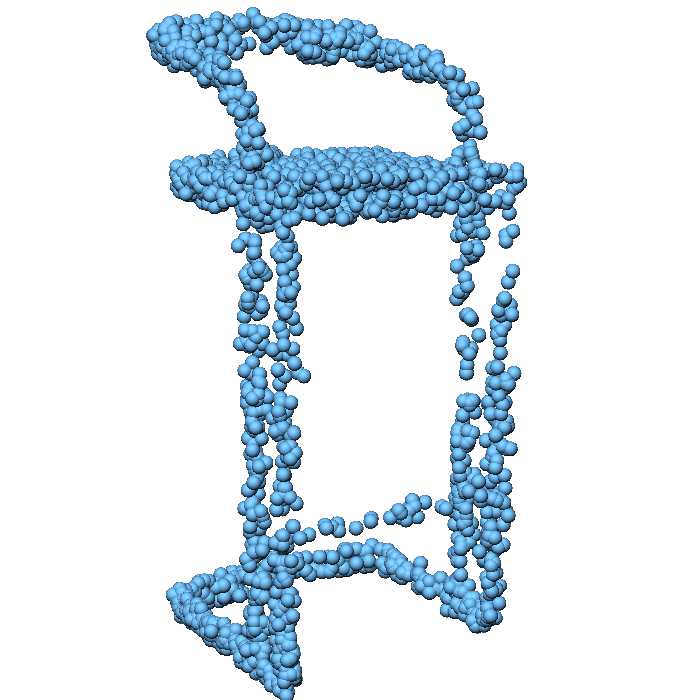}&
\includegraphics[width=\mywidth]{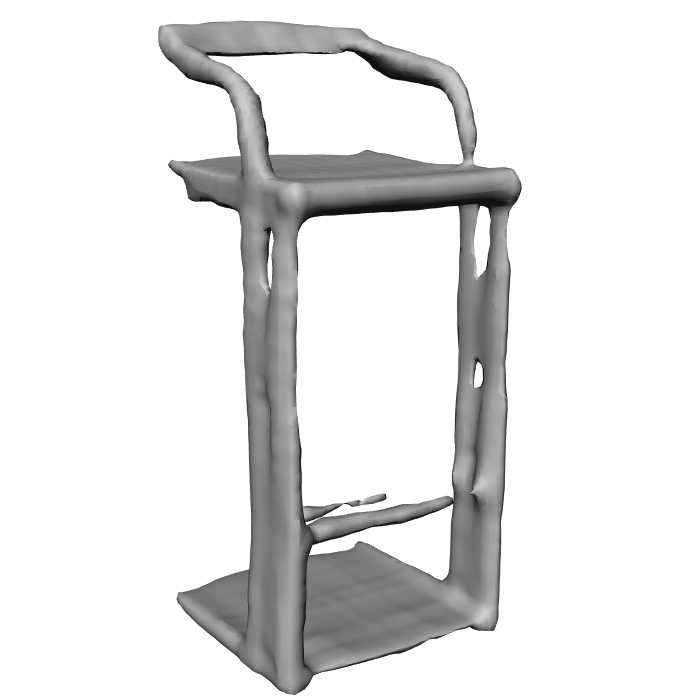}&
\includegraphics[width=\mywidth]{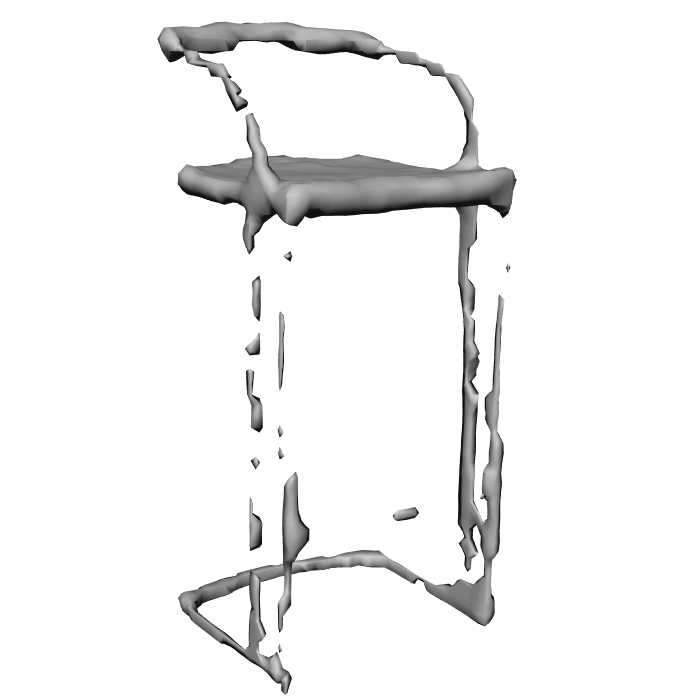}&
\includegraphics[width=\mywidth]{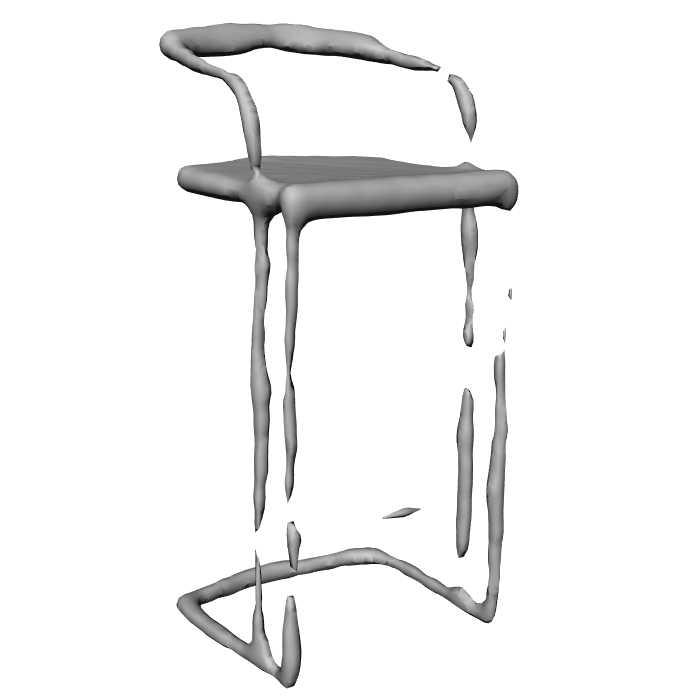}&
\includegraphics[width=\mywidth]{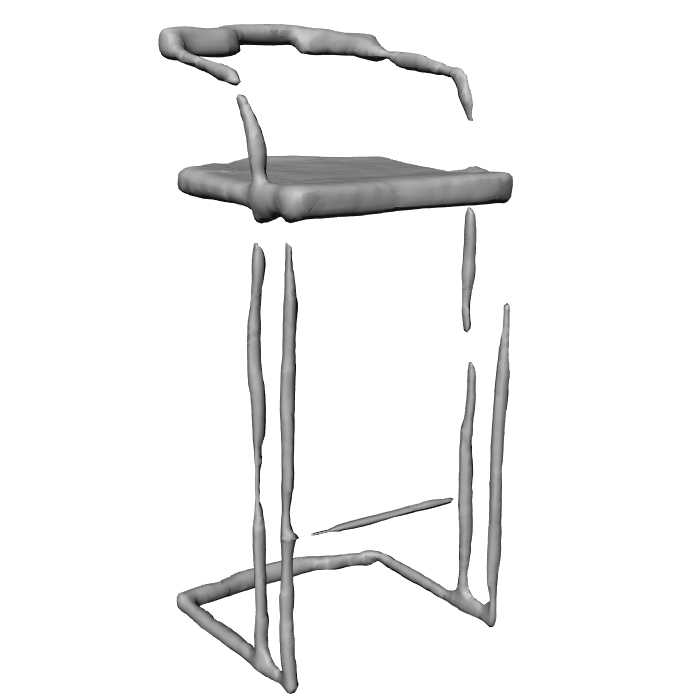}&
\multirow{2}{*}[3em]{\includegraphics[width=\mywidth]{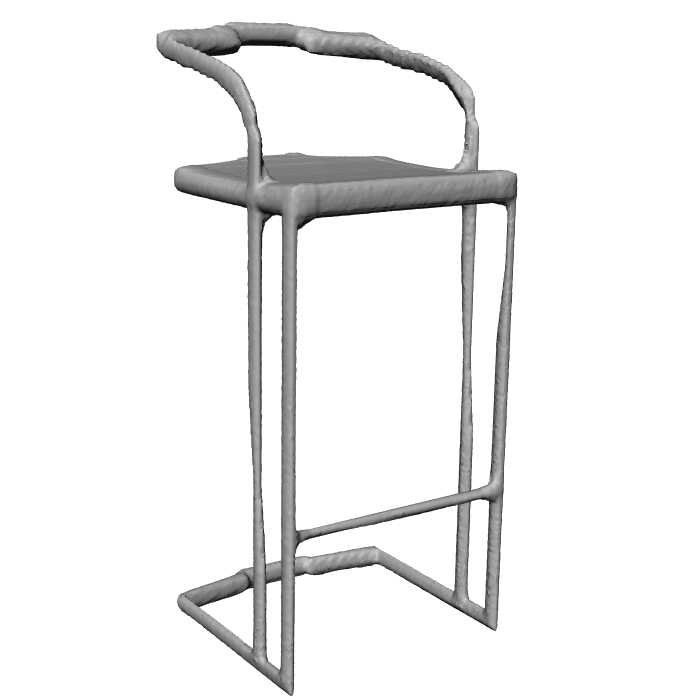}} 
\\
\rotatebox{90}{\hspace{3mm}Augmented}&
\includegraphics[width=\mywidth]{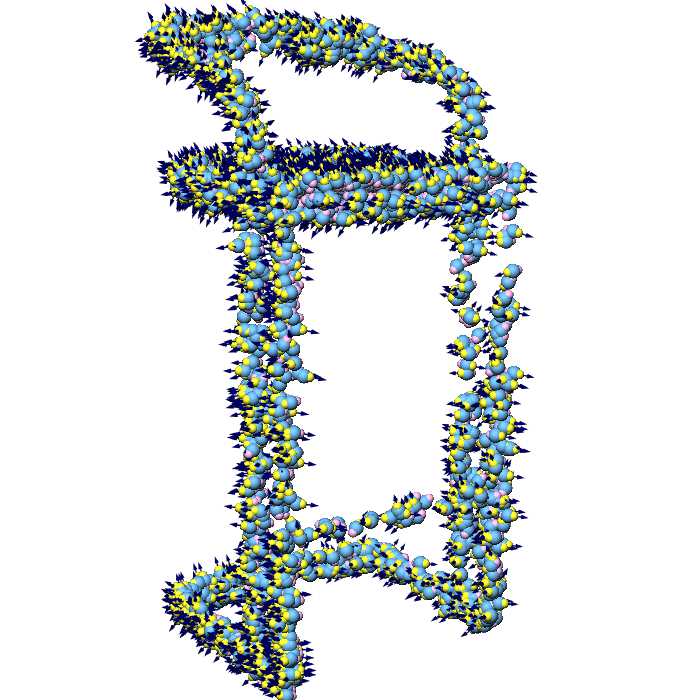}&
\includegraphics[width=\mywidth]{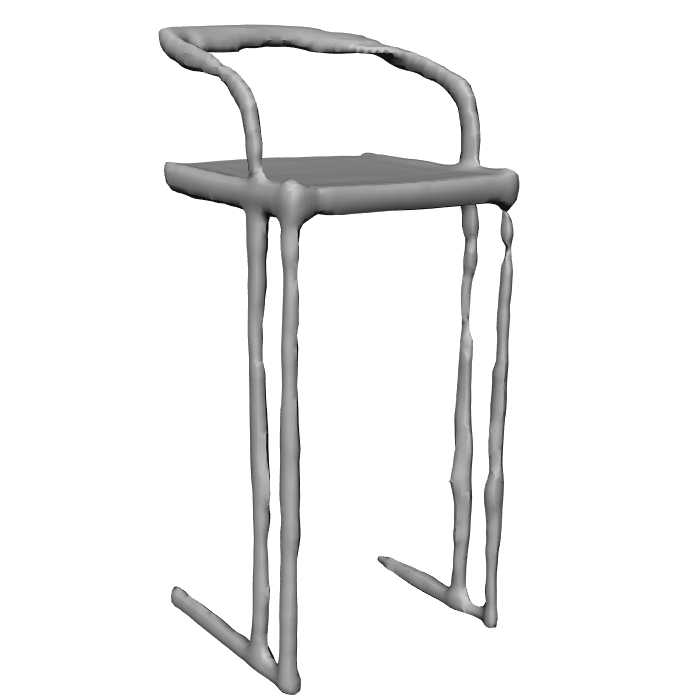}&
\includegraphics[width=\mywidth]{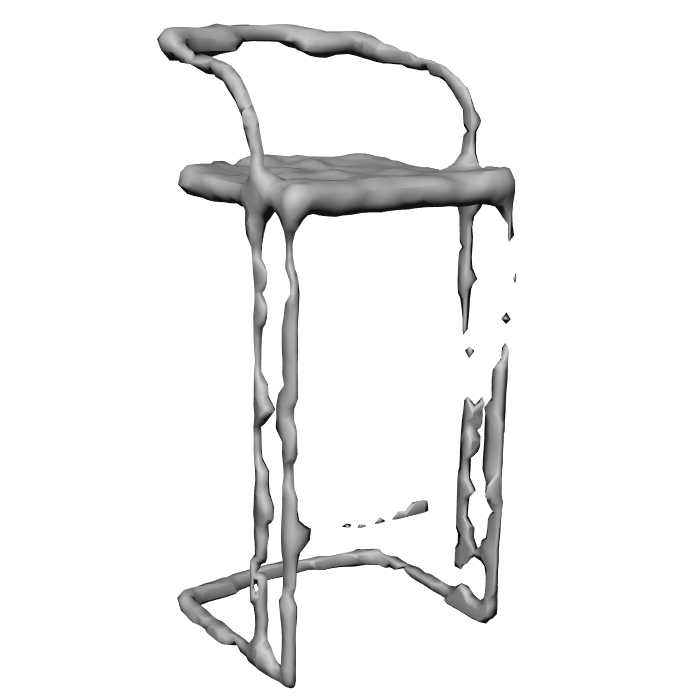}&
\includegraphics[width=\mywidth]{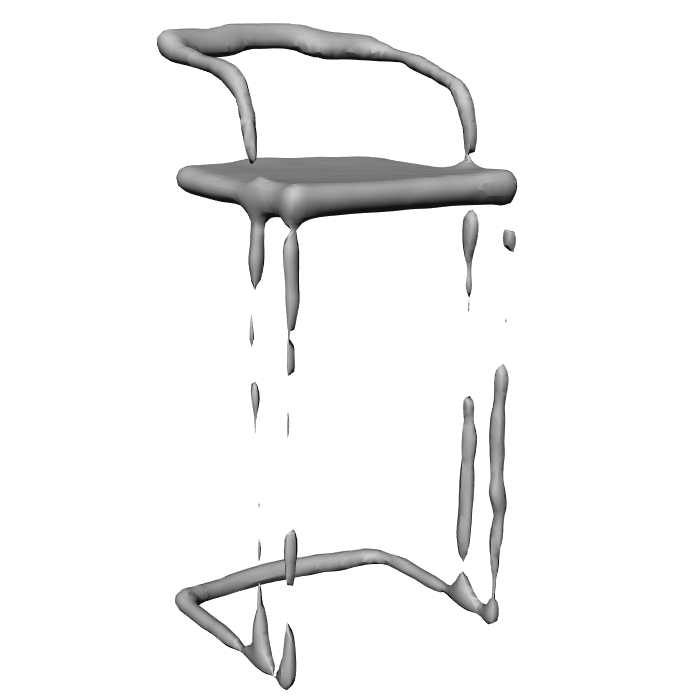}&
\includegraphics[width=\mywidth]{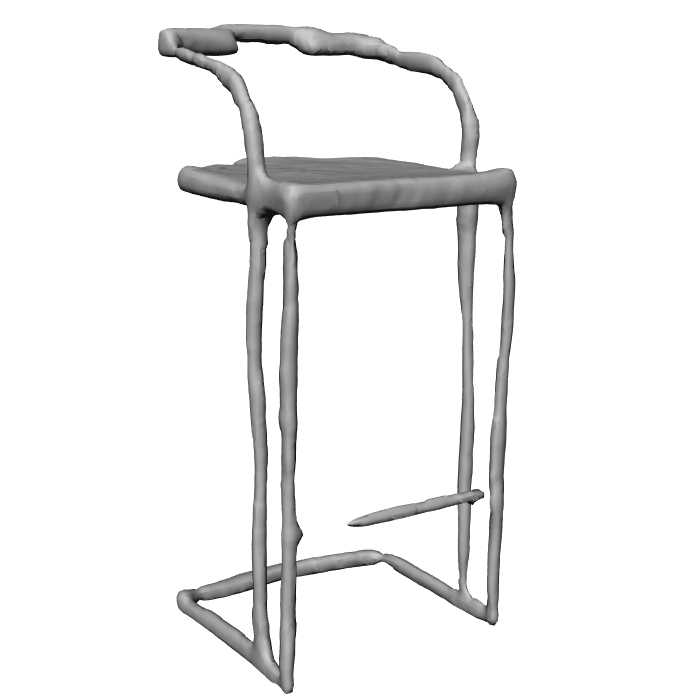}&
\\

&
Input &
ConvONet-2D~\cite{Peng2020}&
Points2Surf~\cite{points2surf}&
Shape\,As\,Points~\cite{Peng2021SAP}&
POCO~\cite{boulch2022poco}&
Ground Truth
\end{tabular}
	\caption{
		\textbf{Object-Level Reconstruction On ModelNet10.} 
		Reconstructed shapes from the ModelNet10 test set using four different DSR methods trained on ModelNet10.
		Top rows of each object use the bare point cloud as input, and bottom rows use the point cloud augmented with visibility information.
		}
	\label{fig:modelnet_suppmat}
\end{figure*}
In \figref{fig:modelnet_suppmat}, we represent additional results of object-level reconstruction on ModelNet10. Concave parts of the objects are frequently reconstructed more accurately by methods using point clouds with visibility information. Surfaces also tend to be more complete when visibility information is used.

\subsection{Synthetic Rooms Dataset}
\begin{figure*}[]
	\centering
	\newcommand{\mywidth}{0.22\textwidth}
\begin{tabular}{ccccccc}

\rotatebox{90}{\hspace{20mm}Bare}&
\includegraphics[width=\mywidth]{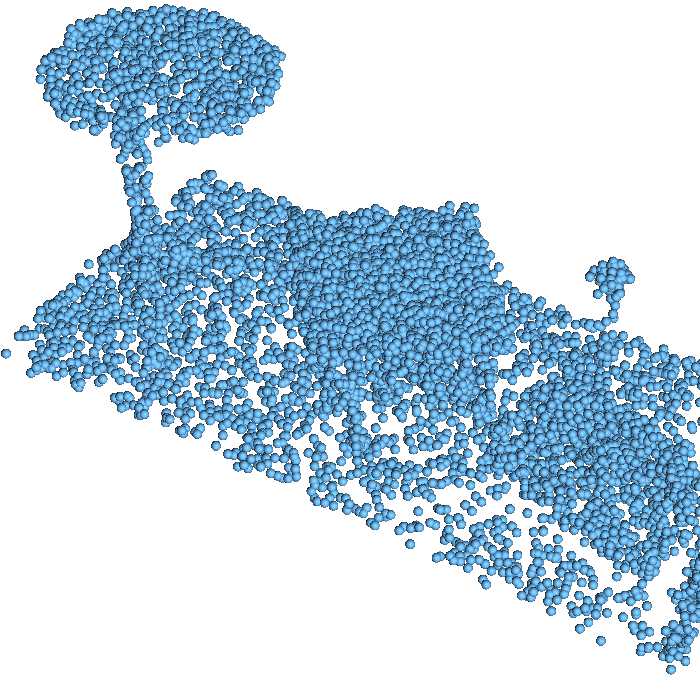}&
\includegraphics[width=\mywidth]{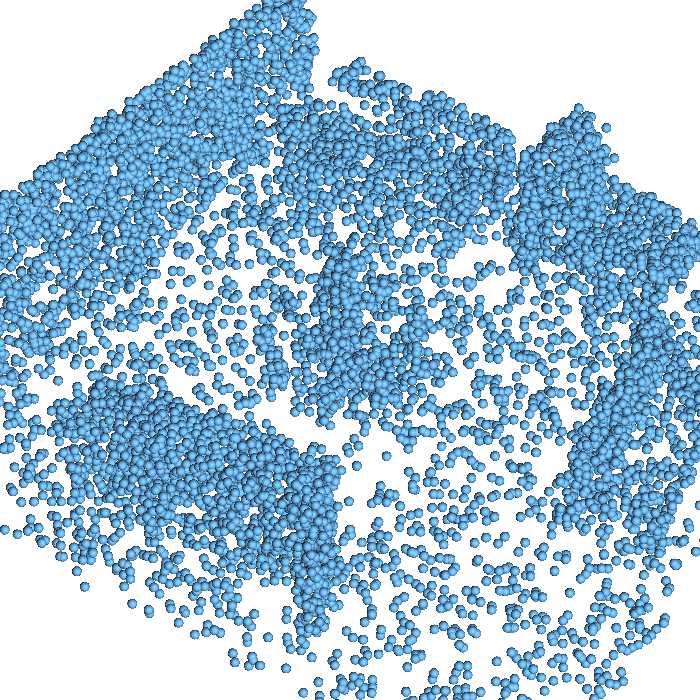}&
\includegraphics[width=\mywidth]{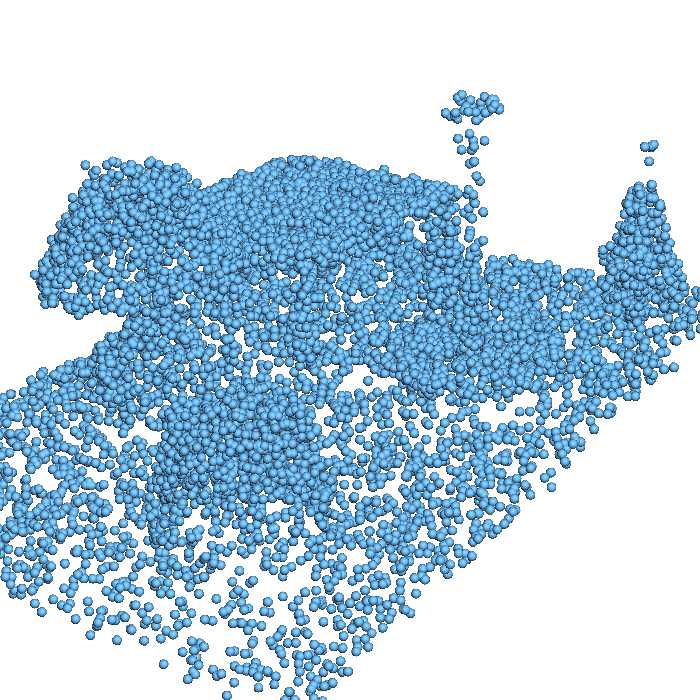}&
\includegraphics[width=\mywidth]{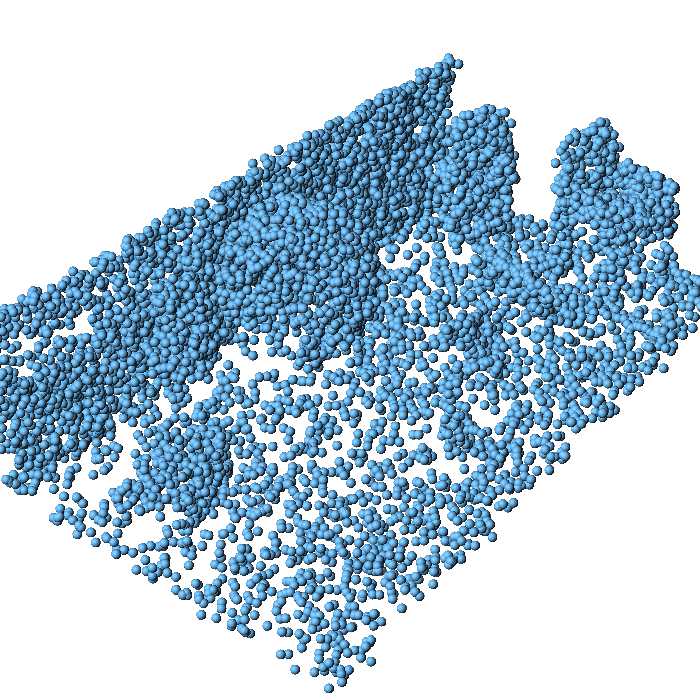}&
\\

\rotatebox{90}{\hspace{12mm}Augmented}&
\includegraphics[width=\mywidth]{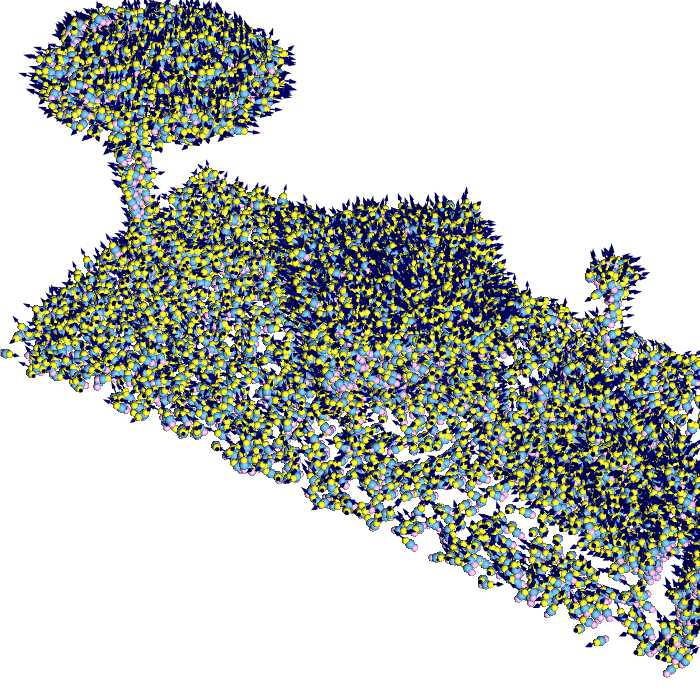}&
\includegraphics[width=\mywidth]{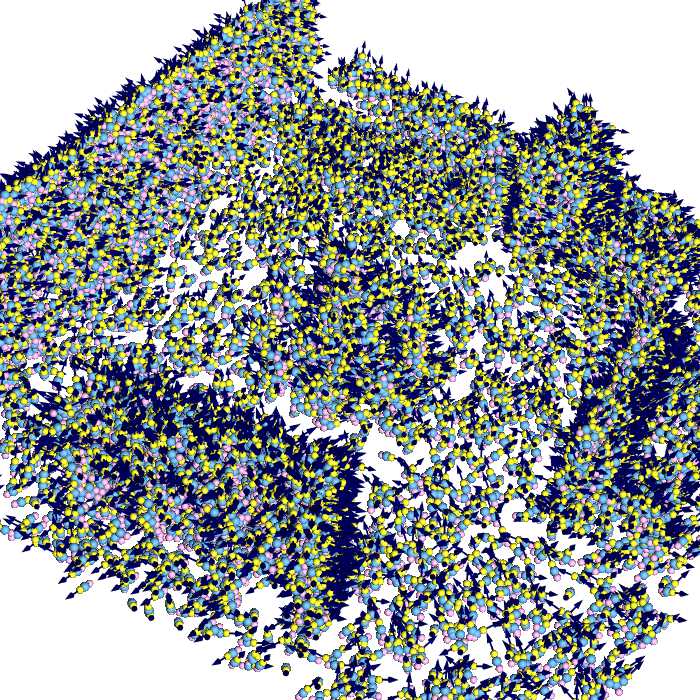}&
\includegraphics[width=\mywidth]{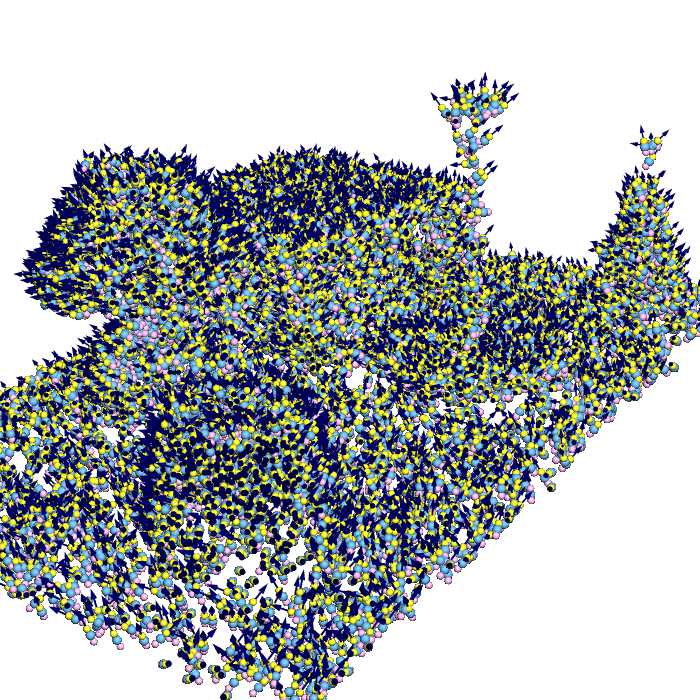}&
\includegraphics[width=\mywidth]{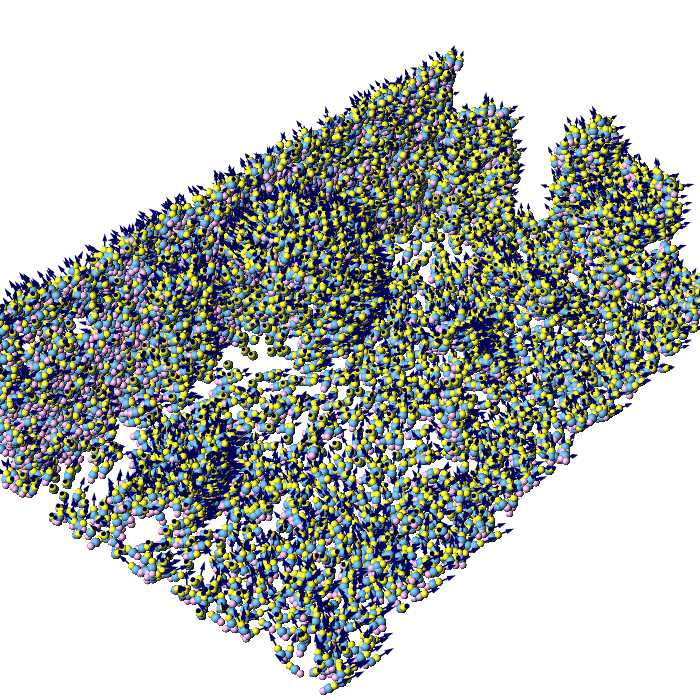}&
\\

\rotatebox{90}{\hspace{9mm}ConvONet~\cite{Peng2020}}&
\includegraphics[width=\mywidth]{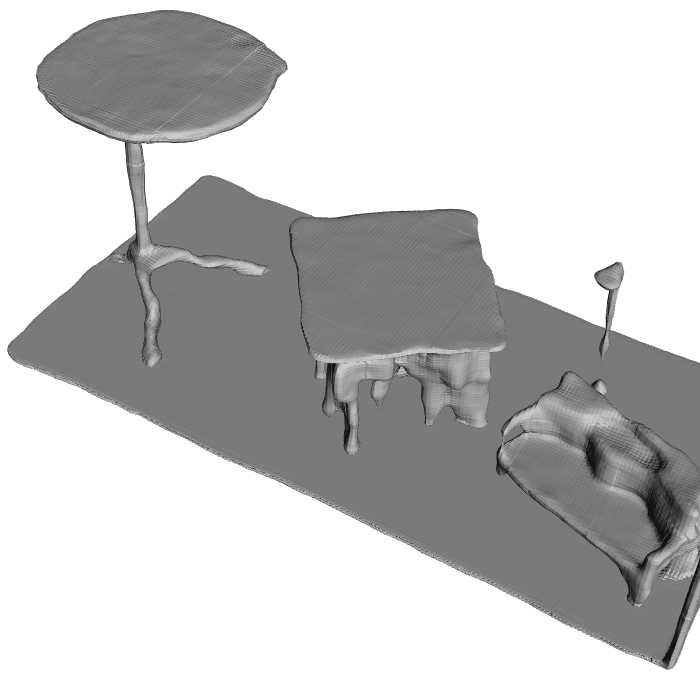}&
\includegraphics[width=\mywidth]{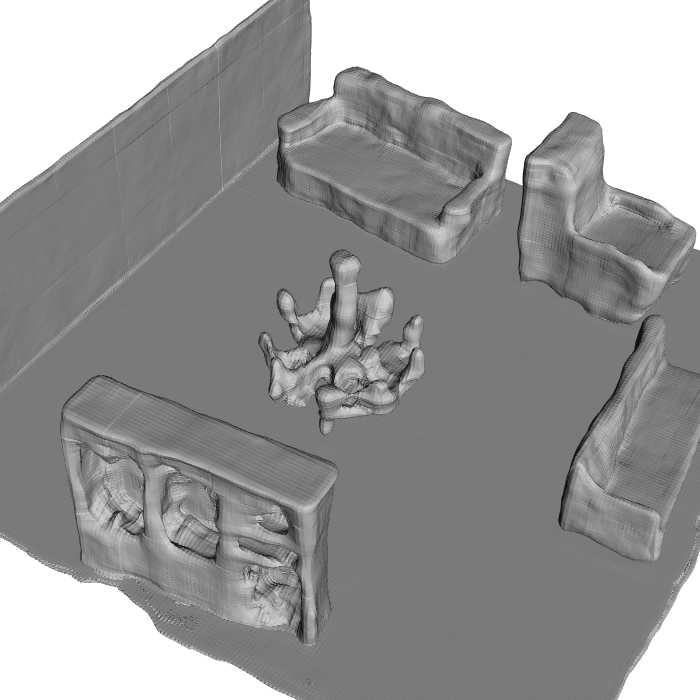}&
\includegraphics[width=\mywidth]{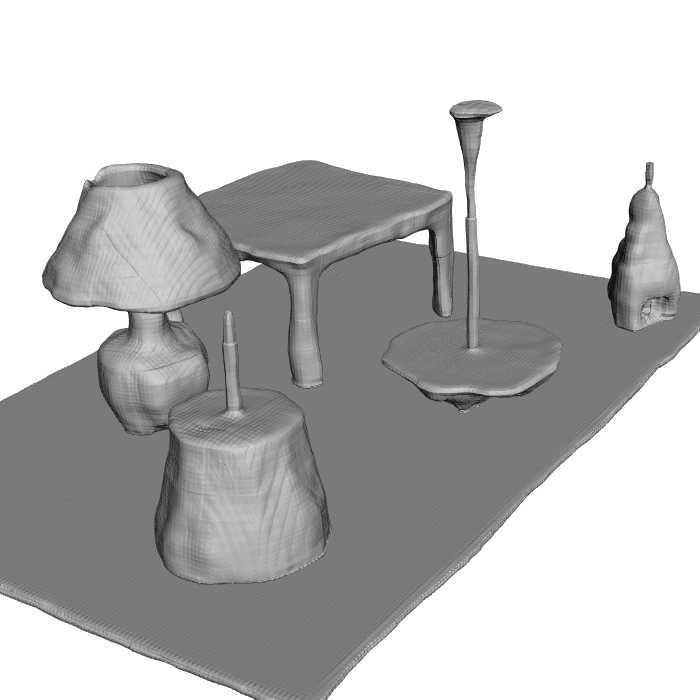}&
\includegraphics[width=\mywidth]{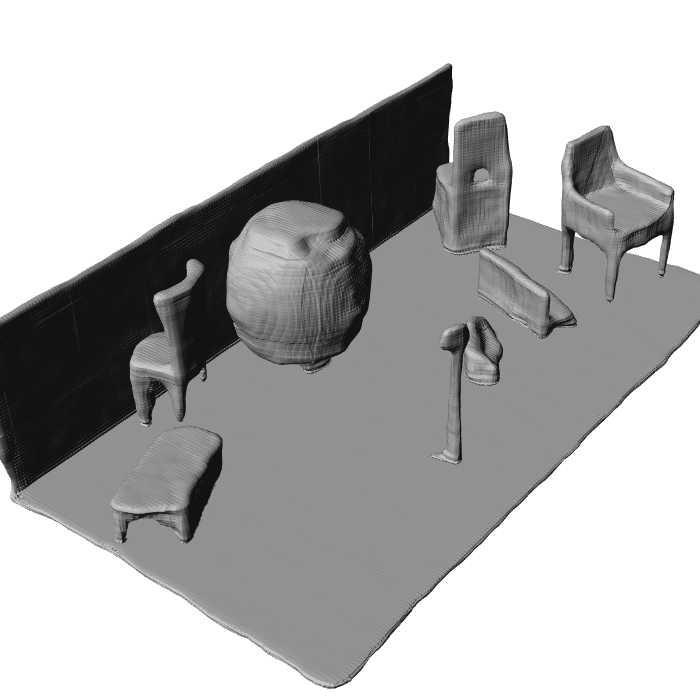}&
\\

\rotatebox{90}{\hspace{4mm}ConvONet~\cite{Peng2020}~(+SV+AP)}&
\includegraphics[width=\mywidth]{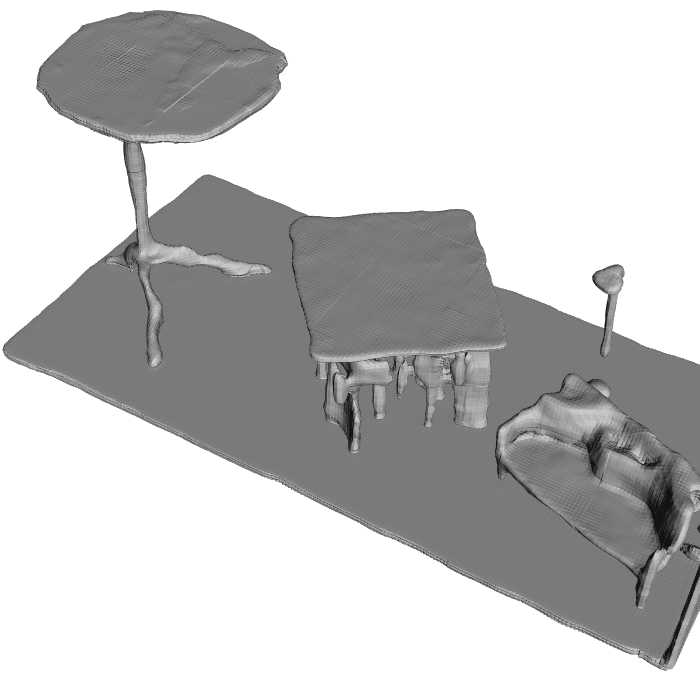}&
\includegraphics[width=\mywidth]{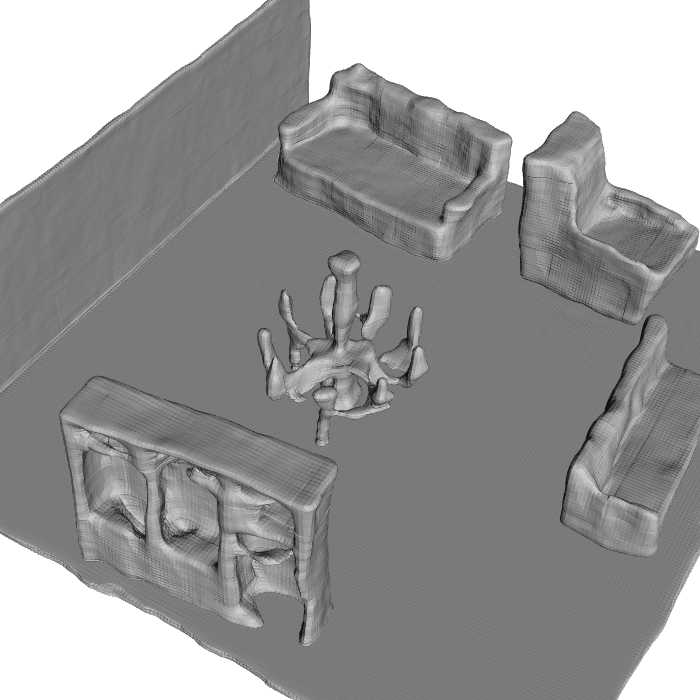}&
\includegraphics[width=\mywidth]{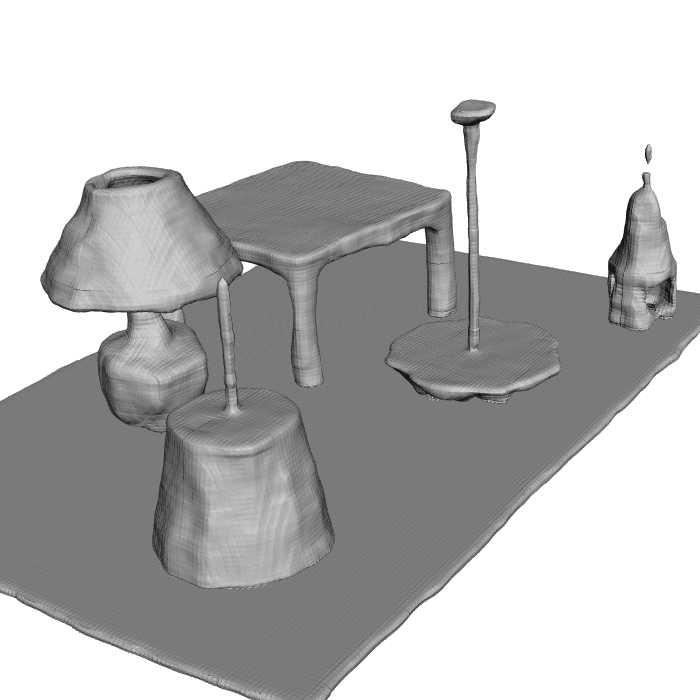}&
\includegraphics[width=\mywidth]{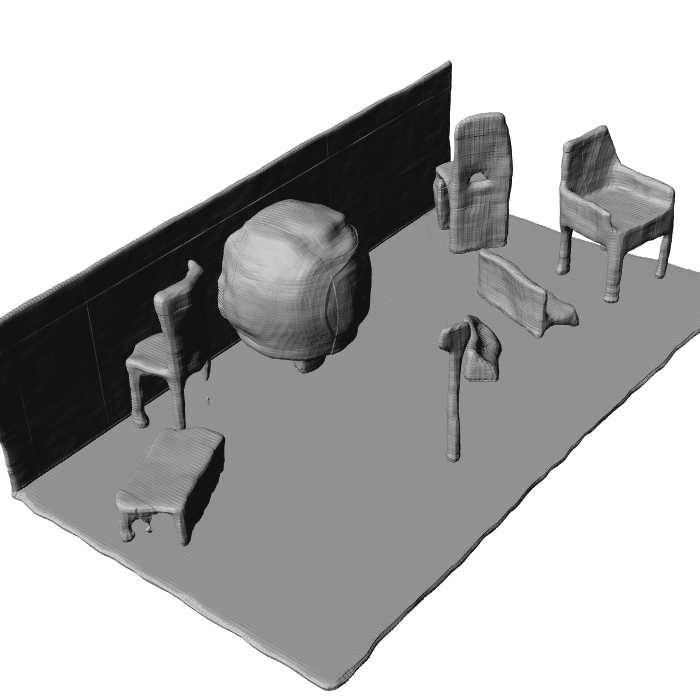}&
\\

\rotatebox{90}{\hspace{9mm}Ground Truth}&
\includegraphics[width=\mywidth]{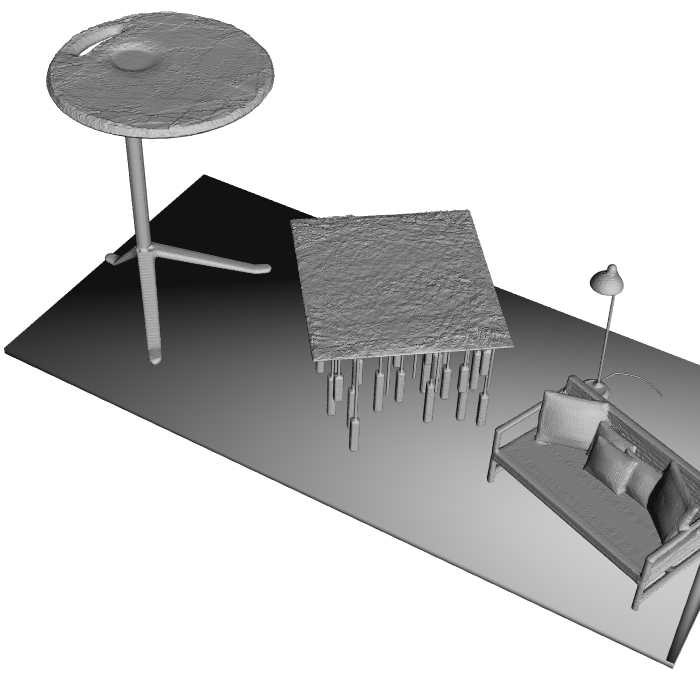}&
\includegraphics[width=\mywidth]{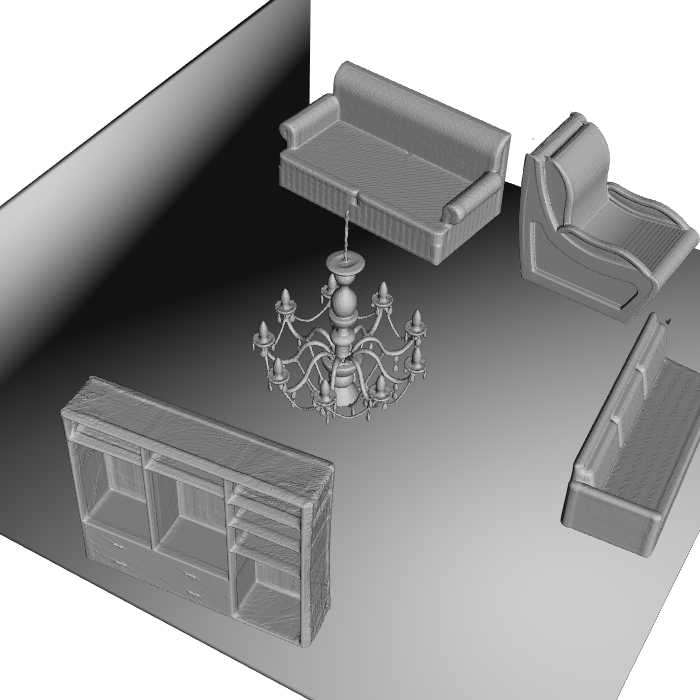}&
\includegraphics[width=\mywidth]{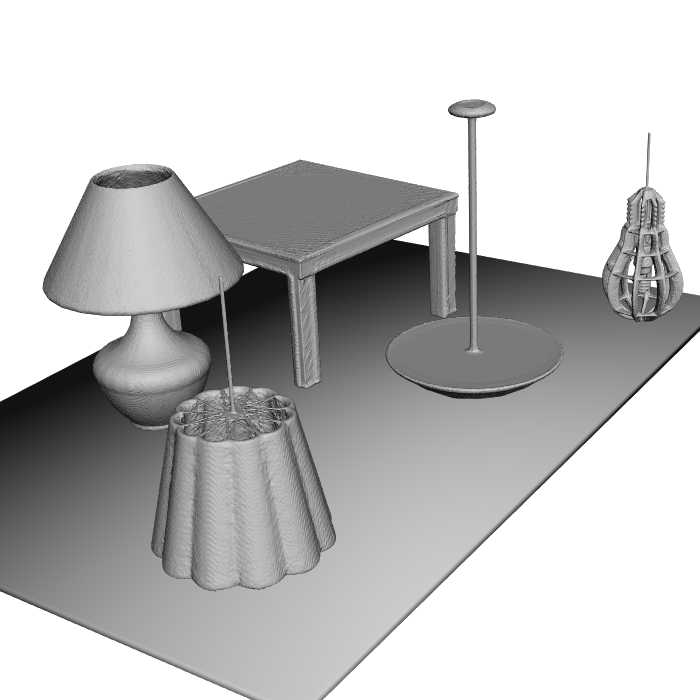}&
\includegraphics[width=\mywidth]{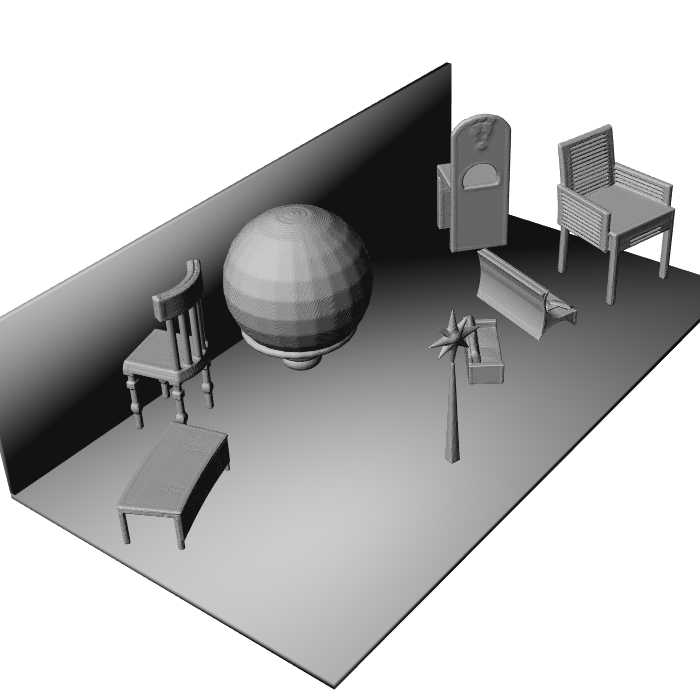}&

\end{tabular}
	\caption{
		\textbf{Scene-Level Reconstruction on Synthetic Rooms.}
	 	Reconstructed scenes of the synthetic room dataset using ConvONet~\cite{Peng2020} in sliding-window mode, with and without visibility information. 
	}
	\label{fig:synthetic_room}
\end{figure*}





In \figref{fig:synthetic_room}, we show the reconstruction results of ConvONet on Synthetic Rooms, with and without visibility information. In contrast to \cite{Peng2020}, we evaluate in sliding window mode, which explains the difference with figures in \cite{Peng2020}. Improvements here are visually not as obvious as with other datasets.

}{}



%



\end{document}